\documentclass{article}
\usepackage{enumitem}
\usepackage{subcaption}
\usepackage{multirow}
\usepackage{caption}
\usepackage{multicol}
\usepackage{algorithm, algorithmicx, algpseudocode}
\usepackage{hyperref}
\usepackage{amsmath,amssymb}
\usepackage{graphicx}

\begin{document}
	
\title{Statistical Modeling of Univariate Multimodal Data}

\author{Paraskevi Chasani and Aristidis Likas$$ \thanks{Corresponding author}\\ \normalsize Department of Computer Science and Engineering \\ \normalsize University of Ioannina \\ 
\normalsize GR 45110, Ioannina, Greece \\
\normalsize e-mail: \{pchasani, arly\}@cs.uoi.gr}

\date{}
\maketitle
\vspace{-0.5cm}
\par\noindent\rule{\textwidth}{0.4pt}

\begin{abstract}
Unimodality constitutes a key property indicating grouping behavior of the data around a single mode of its density. We propose a method that partitions univariate data into unimodal subsets through recursive splitting around valley points of the data density. For valley point detection, we introduce properties of critical points on the convex hull of the empirical cumulative density function (ecdf) plot that provide indications on the existence of density valleys. Next, we apply a unimodal data modeling approach that provides a statistical model for each obtained unimodal subset in the form of a Uniform Mixture Model (UMM). Consequently, a hierarchical statistical model of the initial dataset is obtained in the form of a mixture of UMMs, named as the Unimodal Mixture Model (UDMM). The proposed method is non-parametric, hyperparameter-free, automatically estimates the number of unimodal subsets and provides accurate statistical models as indicated by experimental results on clustering and density estimation tasks.
\end{abstract}

\hspace{-1.3em}\emph{Keywords}: Unimodality, Mixture modeling, Data partitioning, Valley detection, Unimodal mixture model
\par\noindent\rule{\textwidth}{0.4pt}

\vspace{-0.5cm}

\section{Introduction}

Analyzing data and extracting information about the data generation mechanism and the underlying data structure is an important task in machine learning, data mining and statistics. A common approach is to assume that the data has been generated by a statistical process and fit a statistical model to the data. \emph{Mixture models} \cite{bishop2006pattern} are a popular and flexible class of models that assume the data is generated by sampling from a set of component distributions. Specifically, it is assumed that the underlying density generating the data is a mixture density $f = \sum\limits_{i=1}^{K}\pi_i f_i$, where the mixture weights $\pi_i>0$ are such that $\sum\limits_{i=1}^{K} \pi_i=1$ and the component densities $f_1,...,f_K$ represent the different subpopulations. Mixture models are also widely used for clustering \cite{sampaio2024regularization}, where data points are assigned to clusters based on the component distribution that most probably generated them. The clustering procedure involves fitting a mixture model, often using the Expectation Maximization (EM) algorithm, and assigning each data point to the component with the highest posterior probability. Mixture models are flexible in treating data of different characteristics; however a major problem is the specification of the number of mixture components, which should be specified by the user.

The \emph{Gaussian Mixture Model} (GMM) assumes that all mixture components follow the Gaussian distribution. GMMs have been widely adopted as statistical models, since they are mathematically convenient to work with (providing closed form parameter updates), and have been shown to provide good results in practice.
However, the Gaussian assumption is quite restrictive and in several cases it does not provide a well-fitted solution for data generated by other distributions, e.g., uniform.

\emph{Data unimodality} \cite{dharmadhikari1988unimodality} could play a decisive role in building a successful statistical model, estimating the number of components and partitioning a dataset into clusters \cite{kalogeratos2012dip, schelling2020dataset}.
Assessing data unimodality means estimating whether the data has been generated by an arbitrary unimodal distribution like Gaussian, Student's t, triangular, etc. Even the uniform distribution is considered as a unimodal one.
Recognizing unimodality is fundamental for understanding data structure; for instance, clustering methods are irrelevant for unimodal data as it forms a single coherent cluster \cite{chasani2022uu}.

\emph{Mode estimation} also plays a crucial role in understanding complex datasets, especially when dealing with asymmetric or multimodal distributions. Unlike the mean or median, the mode identifies the most frequent values, providing a better measure of central tendency in skewed and asymmetric distributions. Similarly, valley detection reveals boundaries between modes, aiding in cluster identification and low-density region analysis. Both methods are essential for analyzing real-world data, where assumptions of symmetry or unimodality may not hold.

Few methods exist for assessing data unimodality, such as the well-known \emph{dip-test} \cite{hartigan1985dip} and the folding test \cite{siffer2018your}. Although those tests offer a decision on unimodality, they do not provide a unimodal statistical model of the data. Thus, one has to assume a specific parametric form of the density which is too restrictive. A solution to this major problem is provided by the \emph{UU-test} for unimodality \cite{chasani2022uu}. The unique feature of UU-test is that in case it decides that a dataset is unimodal, it also directly provides a statistical model of the data, in the form of a mixture of uniform distributions (i.e., a \emph{Uniform Mixture Model} (UMM)). Such a UMM model has been shown to be effective in modeling data generated from \emph{univariate unimodal distributions of any shape} \cite{chasani2022uu}. 

Since a UMM can be used to model univariate unimodal data, in this work, we propose a more general method, which builds a statistical mixture model that models adequately \emph{univariate multimodal data}, i.e., data generated by distributions with more than one mode (peak). This statistical model is called \emph{Unimodal Mixture Model} (UDMM). Its mixture components correspond to arbitrary unimodal distributions and each of them is modeled using a UMM provided by UU-test algorithm. Thus, UDMM is actually a hierarchical mixture model, since each component is also a uniform mixture model.
We also propose a technique, called \emph{UniSplit}, for determining valley points of univariate multimodal data achieving to split the original data into unimodal subsets. Our approach relies on the idea of unimodality. We introduce properties of critical points (gcm/lcm points) of the data empirical cumulative density function (ecdf) that provide indications on the existence of density valleys. These properties are exploited in the proposed UniSplit algorithm. Based on the computed valley points, the initial dataset is partitioned into unimodal subsets. Then we model each unimodal subset with a UMM and obtain the final UDMM as a mixture of the computed UMMs. In this way the number of UDMM components is automatically determined as a result of the unimodal data splitting procedure.

The proposed approach is very flexible, since it assumes no specific parametric form for each unimodal mixture component, enabling it to model data from various distributions (e.g., Gaussian and uniform). It requires no training and includes no user-specified hyperparameters (apart from a typical statistical significance level of a uniformity test), offering clear benefits compared to methods, such as GMMs or mean shift \cite{fukunaga1975estimation, cheng1995mean}, which require parameters like the number of components or kernel bandwidth.

We summarize the main contributions of this paper as follows:

\begin{itemize}
	\itemsep -0.1em
	\item A novel valley detection method (called UniSplit) is proposed for determining valley points of univariate multimodal datasets and obtaining partitions into unimodal subsets.			
	\item Significant properties of the ecdf are introduced based on critical points (called gcm and lcm) in the convex hull of the ecdf graph, which are related to the existence of density valleys.
	\item A statistical mixture model, called Unimodal Mixture Model (UDMM), is proposed, where each mixture component corresponds to a unimodal distribution.		
	\item The method is flexible, requires no training, while apart from the typical statistical significance level, it does not include user specified hyperparameters. The number of components in the UDMM model is automatically determined by the UniSplit algorithm, rather than being manually defined by the user.
	\item Experiments on synthetic and real datasets compare UniSplit with clustering methods, while UDMM performance in statistical modeling is also evaluated.
\end{itemize}

The outline of the paper is as follows: Section \ref{Related work} discusses related work on mixture models, mode estimation and unimodality-based clustering approaches. Section \ref{Notations} provides necessary definitions and notations, while Section \ref{Valleys} introduces the ideas implemented in our method for detecting valley points in data density. Section \ref{UUtest} briefly describes the UU-test for unimodality and the Uniform Mixture Model (UMM) it provides. Section \ref{UDMM} presents the proposed UniSplit methodology and defines the Unimodal Mixture Model (UDMM). Section \ref{Experiments} provides extensive experimental results on synthetic and real datasets to evaluate the splitting procedure and the performance of UDMM. Finally, in Section \ref{Conclusions} we provide conclusions and directions for future work.

\section{Related Work} \label{Related work}

Multimodal distributions often indicate multiple subpopulations, where the mean and median may not accurately represent the central tendency. Modal analysis, which involves identifying modes and estimating valleys, is crucial for understanding data structure. In multimodal cases, modeling each subpopulation using a mixture density approach (mixture models) such as GMMs, provides insights into the underlying distribution. A recent extension is proposed in \cite{scrucca2019transformation}, where GMMs carry out density estimation not on the original data but on appropriately transformed data in case of bounded variables. The basic idea is to use an invertible function to map a bounded variable to an unbounded support, estimate the density of the transformed variable, and then back-transform to the original scale. For particular applications, mixtures of distributions other than Gaussian have been explored for clustering \cite{li2006two, banerjee2005clustering}.

The proposed UniSplit algorithm is based on the detection of valley points of the underlying density distribution of the data. Valley detection (or modes estimation) has been considered in various methods. Many clustering algorithms assume that multiple modes indicate multiple clusters, while unimodality is a sign for a single cluster. Several classical clustering algorithms are based on partitioning the space around a pre-fixed number of central points (these are usually called partitioning methods, and include K-means clustering, for instance). In the recent times, however, there is a growing body of researchers that advocate that ``density needs to be incorporated in the clustering procedures'' \cite{chacon2019mixture}. In this spirit, mixture models have been successfully used for data clustering, while another density-based approach is clustering based on high density regions \cite{hartigan1975clustering}. In the last approach (also characterized as \emph{modal clustering} \cite{chacon2020modal,menardi2016review}) the clusters are taken as the “domains of attraction” of the density modes.

Kernel density estimation (KDE) is a non-parametric method that estimates the probability density function (pdf) of a dataset by a mixture of kernel functions, such as the Gaussian kernel. KDE assumes one kernel per data point, while the kernel function, the bandwidth of the kernel and other hyperparameters have to be chosen by the user. It is commonly used in modal clustering to identify modes in a dataset, which serve as cluster centers.

Mean shift \cite{fukunaga1975estimation, cheng1995mean} is a ``mode seeking'' clustering algorithm based on the idea of associating each data point to a mode of the underlying probability density function, which is modeled using kernel density estimation. The general idea is to shift each data point until it reaches its nearest peak of the data density, thus a cluster is formed around each peak. A major difficulty is that it includes a critical user defined hyperparameter which is the bandwidth of kernel function used in kernel density estimation. Many proposals for bandwidth selection have been made, however they do not always work successfully \cite{myhre2018robust}. Similarly, medoidshift \cite{sheikh2007mode} and density peaks \cite{rodriguez2014clustering} follow mode-seeking approaches, with medoidshift also requiring the kernel bandwidth and density peaks requiring the specification of several hyperparameters \cite{rasool2023overcoming}.

Two recent methods have been proposed in \cite{li2007nonparametric} and \cite{chacon2019mixture} for modal clustering. In \cite{li2007nonparametric}, the goal is to associate each data point with a local density maximum (mode) using KDE (Gaussian kernels). Modal clustering is achieved through the Modal EM (MEM) algorithm, a nonparametric EM-type method that identifies ``hilltops'' of the density. The algorithm is extended for hierarchical clustering by recursively estimating modes with increasing bandwidths and further improved in \cite{scrucca2021fast} for efficient implementation and flexible covariance matrix decomposition. Once modes are estimated, observations are assigned to their respective modes to form clusters. In \cite{chacon2019mixture} the ``modclust'' methodology integrates modal clustering with mixture modeling by applying the mean shift algorithm to a Gaussian mixture density estimated via the EM algorithm and the Bayesian Information Criterion (BIC) \cite{bishop2006pattern} for component selection. Both modclust and MEM, as any other mode-seeking procedure, depend on the quality of the underlying density estimate, with poorly estimated mixture model parameters leading to inferior solutions. A review on non-parametric modal clustering is provided in \cite{menardi2016review}.

While the above approaches rely on local density estimates for cluster identification, another research direction is to employ unimodality testing \cite{hartigan1985dip, silverman1981using, siffer2018your, chasani2022uu} to decide whether a dataset has a single or multiple modes. It should be emphasized that \emph{efficient unimodality tests currently exist only for univariate data}. 
Skinny dip \cite{maurus2016skinny} identifies clusters in noisy datasets by applying UniDip, which uses the dip-test to detect unimodal intervals in one-dimensional data, classifying points outside these intervals as noise. However, UniDip often misclassifies the tails of distributions as outliers. TailoredDip \cite{bauer2023extension} addresses this limitation by analyzing the spaces between clusters after running UniDip, improving its ability to capture distribution tails and enhancing overall performance.

Apart from clustering, modes estimation is often encountered in the field of image segmentation. The Fine to Coarse (FTC) segmentation algorithm \cite{delon2006nonparametric} detects modes in an image histogram by starting with the finest segmentation based on local minima and merging adjacent unimodal segments. This process continues until no further merging is possible. The segmentation solution should meet two criteria: each segment must be ``statistically unimodal'', and no union of consecutive segments should be unimodal.

\section{Notations and Definitions} \label{Notations}

In this section, we provide the main definitions required to present and clarify our method. The notations are similar to those described in \cite{chasani2022uu}. 
Let $X=\{x_1,...,x_N\}$, $x_i \in \mathbb{R}$ and $x_i < x_{i+1}$ an ordered 1-d dataset of distinct real numbers. 
For an interval [a,b], we define $X(a,b)=\{a \leq x_i \leq b, x_i\in X\}$ the subset of $X$ whose elements belong to that interval. 
Moreover, we denote as $F_X (x)$ the empirical cumulative distribution function (ecdf) of $X$, defined as:

\begin{displaymath}
F_X(x) = \frac{\textup{number\, of\, elements\, in\, the\, sample} \leq x}{N} =
\frac{1}{N} \sum_{i=1}^N I_{(-\infty,x)}(x_i)
\end{displaymath} 
$I_{(-\infty, x)}(x_i)$ is the indicator function:
$I_{(-\infty,x)}(x_i)=\left\{
\begin{array}{ll}
1, & \textup{if} \,\, x_i\leq x \\
0, & \textup{otherwise}\\
\end{array} 
\right. $ 
It also holds that $F_X(x)=0$ if $x < x_1$, $F_X(x)=1$ if $x\geq x_N$. Note also that $F_X(x)$ is \textit{piecewise constant}. For sake of clarity, we will refer to $F_X(x)$ as $F(x)$.

In what concerns the unimodality of a distribution there are two definition options. The first relies on the probability density function (pdf): a pdf is unimodal, if it has a single mode; a region where the density becomes maximum, while non-increasing density is observed when moving away from the mode. In other words, a pdf $f(x)$ is a unimodal function if for some value $m$, it is monotonically increasing for $x\leq m$ and monotonically decreasing for $x\geq m$. In that case, the maximum value of $f(x)$ is $f(m)$ and there are no other local maxima. The second definition option relies on the cumulative distribution function (cdf): a cdf $F(x)$ is unimodal if there exist two points $x_l$ and $x_u$ such that $F(x)$ can be divided into three parts: a) a convex part $(-\infty,x_l)$, b) a constant part $[x_l,x_u]$ and c) a concave part $(x_u,\infty)$. It is worth mentioning that it is possible for either the first two parts or the last two parts to be missing. It should be stressed that the uniform distribution is considered unimodal and its cdf is linear. A distribution that is not unimodal is called multimodal with two or more modes. Those modes typically appear as distinct peaks (local maxima) in the pdf plot. 

Let a subset $S=\{s_1,...,s_L\}$ of $X (s_i \in X)$ with $s_i \neq s_j$, $s_1=x_1$, $s_L=x_N$. We define the \textit{piecewise linear cdf} $PL_S(x)$ obtained by ``drawing'' the line segments from $(s_i, F(s_i))$ to $(s_j, F(s_j))$. Also, we assume that $PL_S(x)=0$ if $x < s_1$ and $PL_S(x)=1$ if $x\geq s_L$. It is important to note that using a piecewise linear cdf $PL_S(x)$ as data model, we make the assumption that the subset $X(s_i,s_{i+1})$ of data points  in each interval $[s_i,s_{i+1}]$ is uniformly distributed. Thus $PL_S(x)$ is actually the cdf of a Uniform Mixture Model (UMM). 

The subset $S$ will be called \textit{sufficient} if the cdf $PL_S(x)$ is a good statistical model of $X$. Since $PL_S(x)$ models the data in each interval using the uniform distribution, in order for $PL_S(x)$ to be a good statistical model of $X$, for each $i$ the subset $X(s_i,s_{i+1})$ should follow the uniform distribution as decided by a uniformity test. Thus in the case where $PL_S(x)$ is sufficient, the corresponding uniform mixture model fits well to the data. If $PL_S(x)$ is both unimodal and sufficient then the dataset $X$ is considered unimodal and $PL_S(x)$ provides a good statistical model of $X$. In order to address the unimodality issue of $PL_S(x)$ we confine our search to the gcm and lcm points of the ecdf, exploiting the idea used in the dip-test method for computing the dip statistic.

More specifically, we define the greatest convex minorant ($gcm$) of a function $F$ in $(-\infty,a]$ as $sup G(x)$ for $x\leq a$, where the $sup$ is taken over all functions $G$ that are convex in $(-\infty,a]$ and nowhere greater than $F$. Based on the above definition, we denote as $G_X(x)$ the gcm of ecdf $F(x)$. Note that, \textit{since $F(x)$ is piecewise constant, $G_X(x)$ is piecewise linear}. Let $G=\{g_1,...,g_{P_G}\} \subset X$ the set of gcm points (where $g_1=x_1, g_{P_G}=x_N$). The graph of $G_X(x)$ is defined by drawing line segments from ($g_i, F(g_i)$) to ($g_j, F(g_j)$). Based on the $PL$ definition, we can write: $G_X(x)=PL_G(x)$. It should be mentioned that the gcm function of $F(x)$ corresponds to the monotonically increasing part of a pdf plot. 

\begin{figure}[H]
	\centering
	\begin{subfigure}[b]{0.7\linewidth}
		\includegraphics[width=\linewidth]{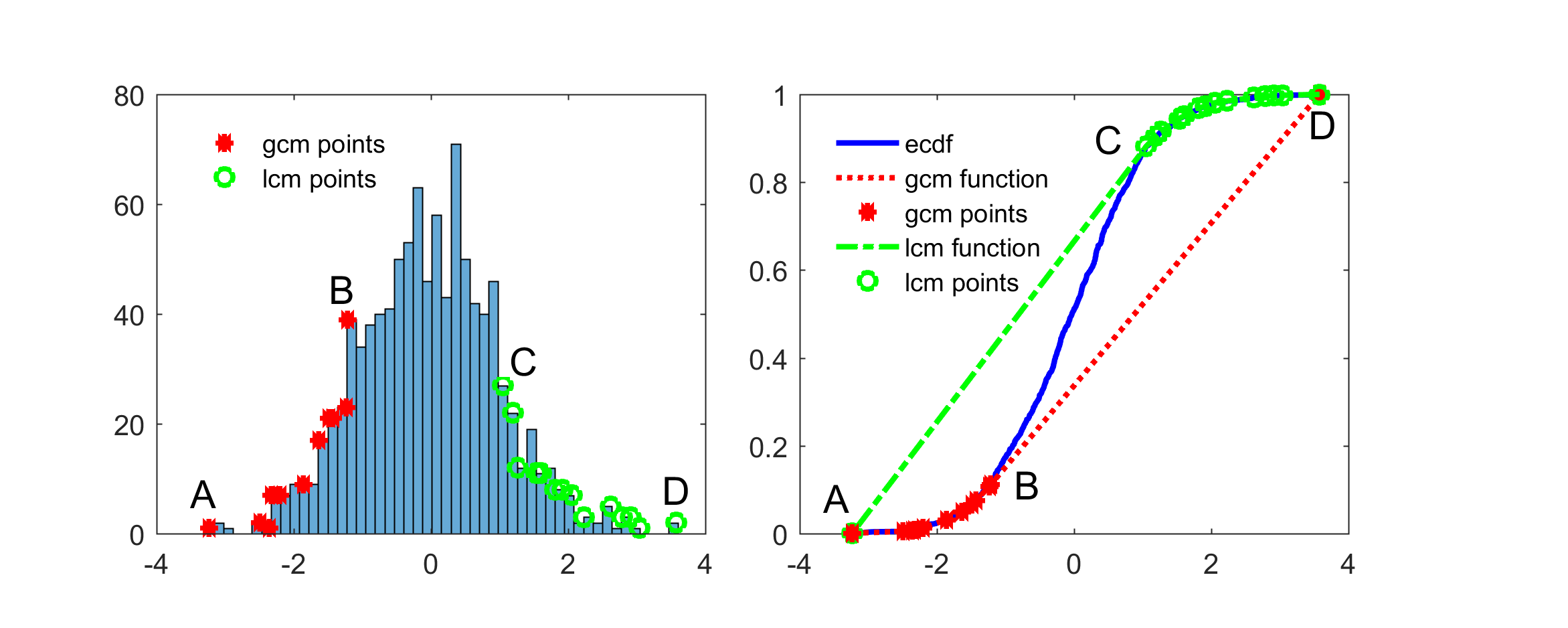}
	\end{subfigure}
	\caption{Histogram: gcm ($AB$ part) and lcm ($CD$ part) correspond to increasing and decreasing parts, respectively. Ecdf: $AB$, $BC$ and $CD$ correspond to the convex, intermediate and concave part, respectively.}
	\label{figure:fig2}
\end{figure}

Similarly, the least concave majorant ($lcm$) of a function $F$ in $[a,\infty)$ is defined as $inf L(x)$ for $x\geq a$,  where the $inf$ is taken over all functions $L$ that are concave in $[a,\infty)$ and nowhere less than $F$. We denote as $L_X(x)$ the lcm of ecdf $F(x)$. Since $F(x)$ is piecewise constant, \textit{$L_X(x)$ is piecewise linear}. Let $L=\{l_1,...,l_{P_L}\} \subset X$ the set of lcm points (where $l_1=x_1, l_{P_L}=x_N$). Its graph is defined by drawing line segments from ($l_i, F(l_i)$) to ($l_j, F(l_j)$) and we can write that $L_X(x)=PL_L(x)$.
The lcm function of $F(x)$ corresponds to the monotonically decreasing part of a pdf plot. 
The right plot of Fig.~\ref{figure:fig2} presents an ecdf (blue line) along with the gcm function $G_X(x)$ (red line), the corresponding set of gcm points $G$ (red stars), the lcm function $L_X(x)$ (green line) and the corresponding set of lcm points $L$ (green circles).

Given the sets of gcm ($G$) and lcm points ($L$) of ecdf $F(x)$, we define as $GL$ the ordered set of points obtained from the union of $G$ and $L$:
$GL=\{v_1, ..., v_M\}$, where $v_1=x_1$, $v_M=x_N$, $v_i < v_j$ if $i < j$ and either $v_i \in G$ or $v_i \in L$. Note that $v_1=x_1$ and $v_M=x_N$ belong to both $G$ and $L$.
We also define as $maxG=max(v_i | v_i \in G-\{x_N\})$ and $minL=min(v_i | v_i \in L-\{x_1\})$, the maximum value of $G$ and the minimum value of $L$ respectively, excluding the maximum and minimum elements of $X$.

\section{Detecting Valleys in Data Density} \label{Valleys}

The shape of the ecdf of a univariate dataset provides crucial information on the multimodality of the underlying data distribution. Gcm and lcm points constitute key points in the ecdf plot, since their location and the uniformity of intervals defined by successive gcm/lcm points constitute significant indicators related to the existence of density valleys in those intervals. We have identified and present below three main cases for intervals $[a,b]$ defined by successive gcm or lcm points: 

\begin{enumerate}[label=(\alph*)]
	\item Uniformity of $X(a,b)$ indicates no density valley in $[a,b]$.
	\item If $X(a,b)$ is non-uniform and unimodal, a single density valley exists in $[a,b]$.
	\item If $X(a,b)$ is non-uniform and multimodal, multiple density valleys exist in $[a,b]$.
\end{enumerate}
We clarify below in detail each of the above cases and present illustrative figures. 
 
(a) \textit{Uniform $X(a,b)$ indicates no density valley in $[a,b]$:} in case $X(a,b)$ is uniform, the corresponding ecdf segment is linear. Based on the type of $a$ and $b$ (gcm or lcm), they both lie on the increasing or decreasing part of the same mode on a histogram plot, respectively.
Fig.~\ref{figure:fig2} illustrates the histogram and ecdf plots of a unimodal dataset. The ecdf segments between the gcm/lcm points are linear, indicating uniformity. On the histogram plot, the gcm points (between $A$ and $B$) lie on the increasing part of the mode, and the lcm points (between $C$ and $D$) lie on the decreasing part, thus no density valleys are detected between gcm or between lcm points.

\begin{figure}[!t]
	\hspace{-0.8cm}
	\begin{subfigure}[b]{0.63\linewidth}
		\includegraphics[width=\linewidth]{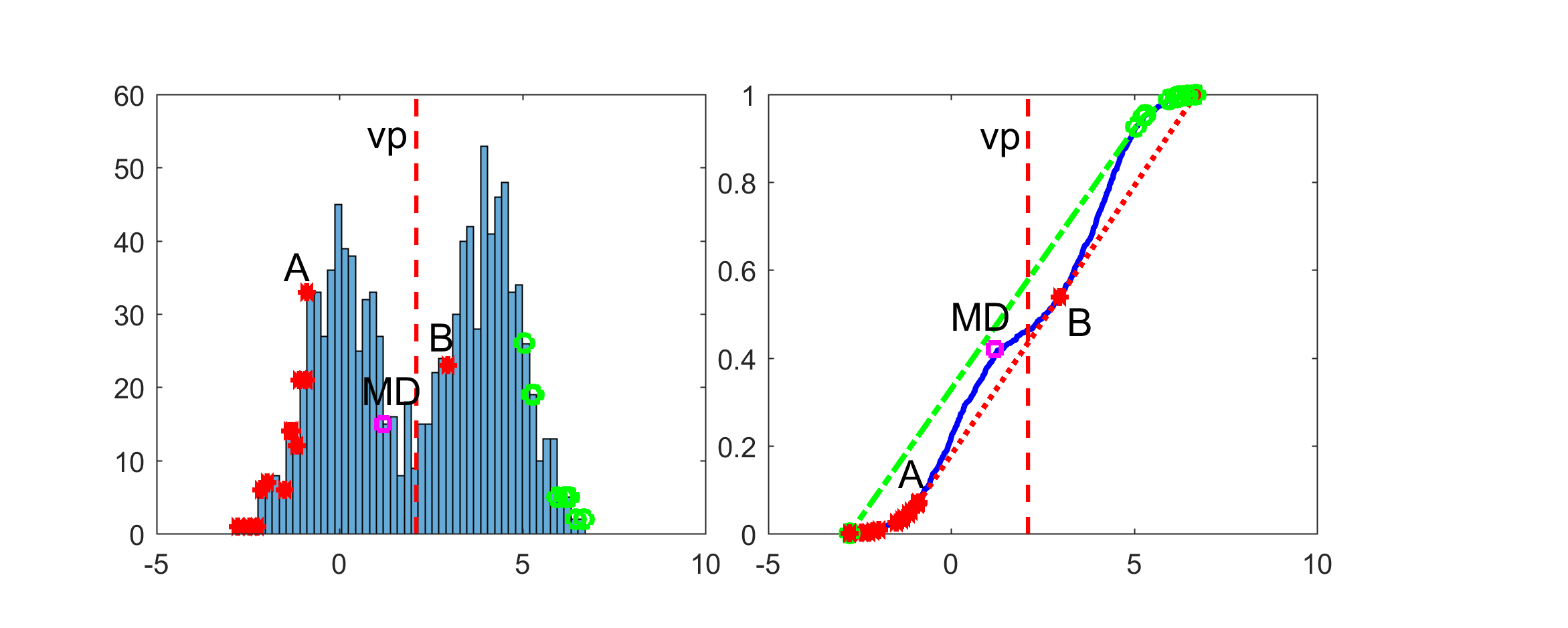}
		\caption{$A$,$B$ are gcm points on increasing parts\newline of successive modes.}
		\label{figure:fig4a}
	\end{subfigure} 
	\hspace{-1.28cm}
	\begin{subfigure}[b]{0.63\linewidth}
		\includegraphics[width=\linewidth]{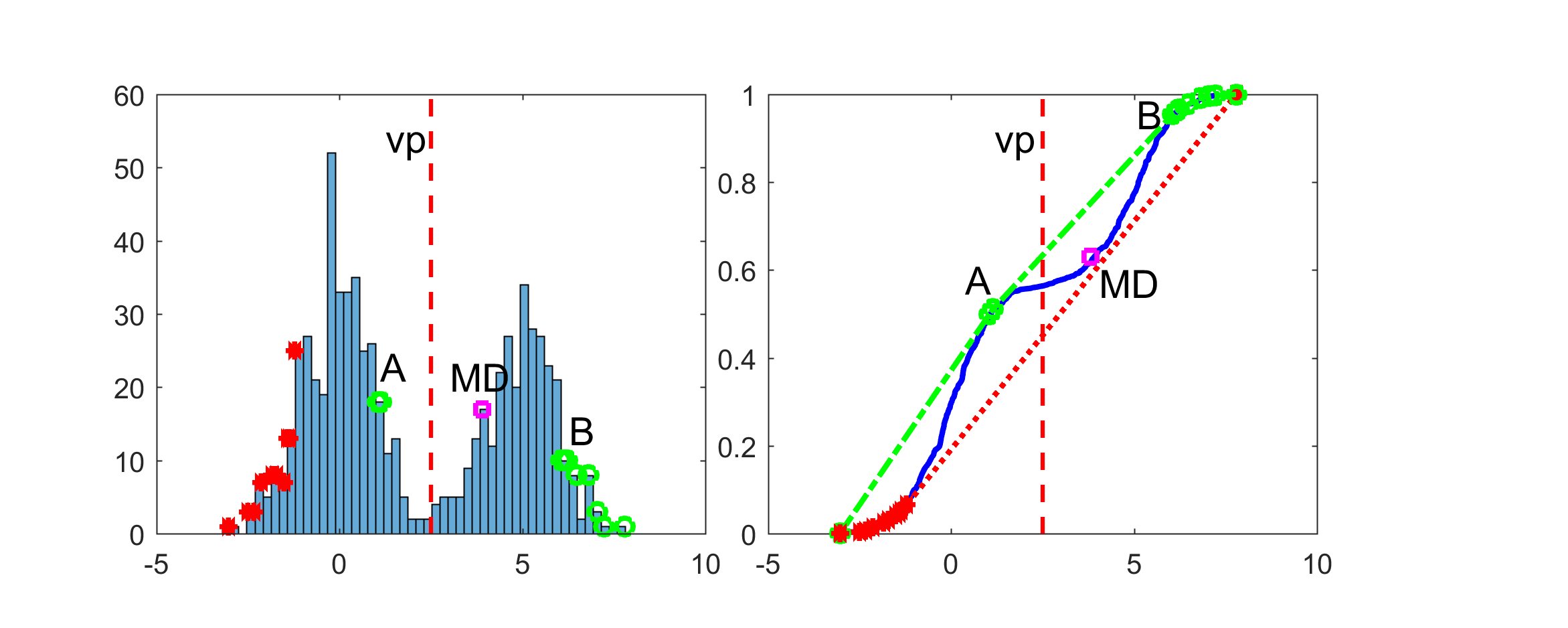}
		\caption{$A$,$B$ are lcm points on decreasing parts of\newline successive modes.}
		\label{figure:fig4b}
	\end{subfigure}
	\vspace{-0.6cm}
	\caption{Histogram and ecdf of a bimodal dataset. The non-uniform and unimodal $X(x_A,x_B)$ indicates a density valley between $A$ and $B$. $MD$ is a point close to the valley. $vp$ is the valley point.}
	\label{figure:fig4}
\end{figure}

(b) \textit{Non-uniform and unimodal $X(a,b)$ indicates a single density valley in $[a,b]$:} non-uniformity of $X(a,b)$ indicates the existence of non-linear ecdf segments (i.e., convex and/or concave ecdf segments) within $[a,b]$. If an interval $[a,b]$ exists, where $a, b$ are successive gcm points and $X(a,b)$ is non-uniform and unimodal, this implies that the ecdf segment is exclusively concave. This property is ensured, since, if it were partially convex and partially concave, then $X(a,b)$ would be multimodal. Thus, $a$ and $b$ lie on increasing parts, while the concave segment corresponds to a decreasing part between $a$ and $b$ on a histogram plot, with one valley (and one peak) being detected in $[a,b]$. It is evident that $a$ and $b$ lie on increasing parts of successive modes. Fig.~\ref{figure:fig4a} illustrates the histogram and ecdf plots of a bimodal dataset sampled from two close Gaussians. On the ecdf plot, we can see that $A, B$ are gcm points and $X(x_A,x_B)$ is not uniformly distributed ($AB$ is non-linear), since the ecdf segment $AB$ is concave (unimodal $X(x_A,x_B)$). It is evident that one density valley is formed between $A$ and $B$ on the histogram plot. Similarly, in case $a, b$ are successive lcm points, the ecdf segment is convex. The lcm points $a$ and $b$ lie on successive decreasing parts of different modes, while the convex segment corresponds to an increasing part between $a$ and $b$ on a histogram plot. Thus one density valley (and one peak) is detected in $[a,b]$. Fig.~\ref{figure:fig4b} illustrates the histogram and ecdf plots of a bimodal dataset. On the ecdf plot, $AB$ is convex, while on the histogram plot $A$ and $B$ lie on the decreasing parts of different modes with one density valley (and one peak) being formed between them.

\begin{figure}[!t]
	\centering
	\begin{subfigure}[b]{0.65\linewidth}
		\includegraphics[width=\linewidth]{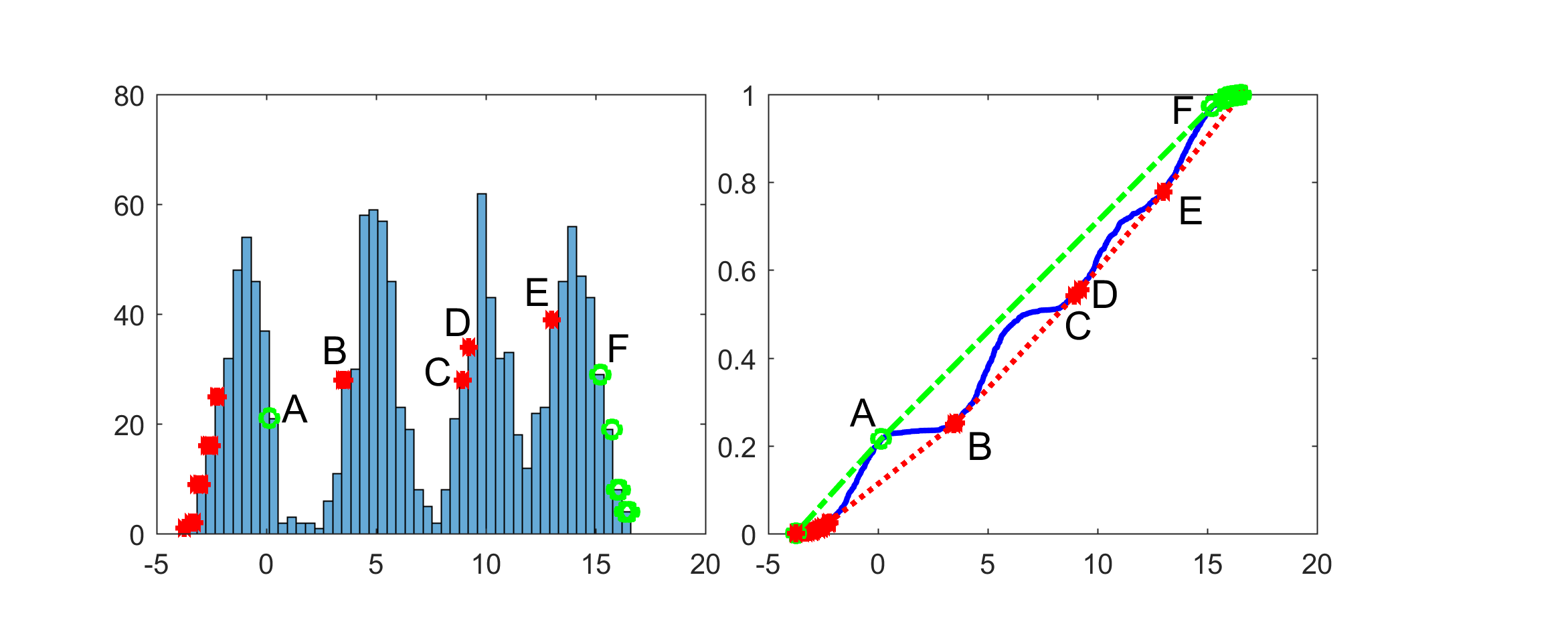}
		\caption{Multiple density valleys in non-uniform and multimodal $X(x_A,x_F)$.}
		\label{figure:fig5a}
	\end{subfigure} 
	\begin{subfigure}[b]{0.65\linewidth}
		\includegraphics[width=\linewidth]{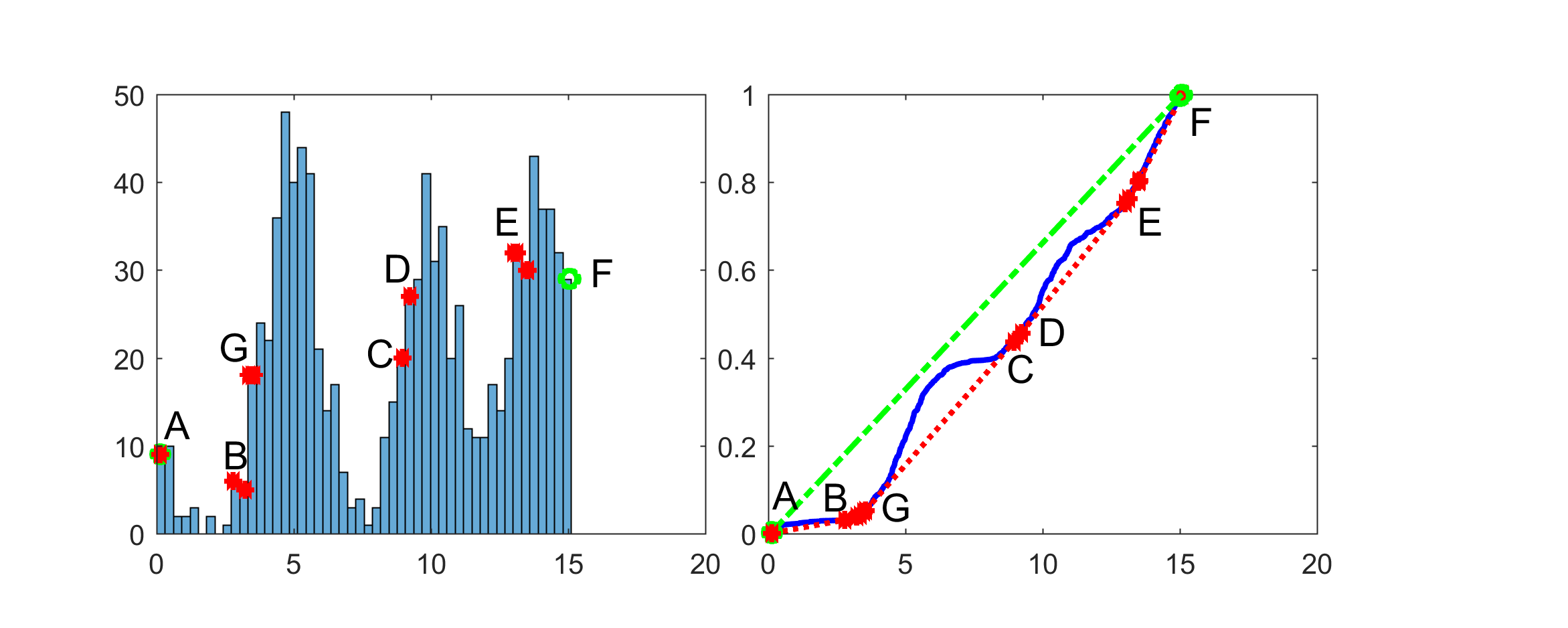}
		\caption{Candidate splitting intervals $[x_A,x_B], [x_G,x_C]$, $[x_D,x_E]$ in zoomed set $X(x_A,x_F)$.Best splitting interval: $[x_G,x_C]$.}
		\label{figure:fig5b}
	\end{subfigure}
	\begin{subfigure}[b]{0.65\linewidth}
		\includegraphics[width=\linewidth]{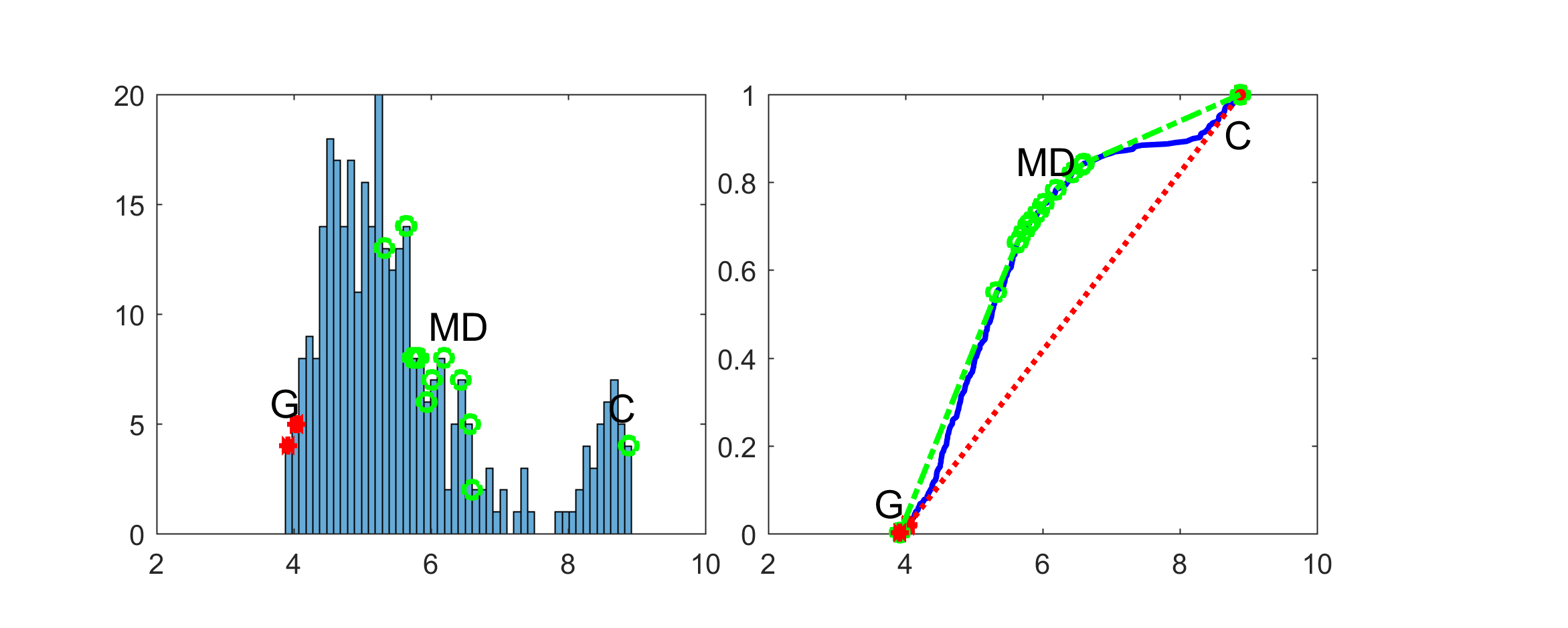}
		\caption{Non-uniform and unimodal set $X(x_G,x_C)$ with a density valley being formed between $G$ and $C$. MD point is also illustrated.}
		\label{figure:fig5c}
	\end{subfigure}
	\caption{Histogram and ecdf plot of a multimodal dataset with its best splitting intervals, processed recursively until a non-uniform and unimodal interval containing a single valley point is detected.}
	\label{figure:fig5}
\end{figure}

(c) \textit{Non-uniform and multimodal $X(a,b)$ indicates multiple density valleys in $[a,b]$:} similarly to case (b), non-uniformity of $X(a,b)$ corresponds to a non-linear ecdf segment, and since $X(a,b)$ is multimodal, the corresponding ecdf is expected to include convex and concave segments. Thus, multiple increasing/decreasing parts exist on a histogram plot, i.e., multiple density valleys are formed. We should note here that all multimodal sets $X(a,b)$ are also non-uniform. We choose to refer both properties of multimodality and non-uniformity for sake of clarity. In Fig.~\ref{figure:fig5a} the histogram and ecdf plot of a multimodal dataset are illustrated. On the ecdf plot, $A$ and $F$ are successive lcm points with $X(x_A,x_F)$ being non-uniform and more specifically multimodal. Multimodality is evident by the multiple peaks and density valleys on the histogram plot and the multiple convex/concave parts between $A$ and $F$ on the ecdf plot.

\subsection{Multimodality Degree}

To apply our splitting algorithm, we need a fast and easy way to assess of the \emph{degree of multimodality} of a data subset. To define the multimodality degree, we consider the \emph{distance among the peaks} and the \emph{depth of the valley between the peaks on the data histogram}. As the distance and the depth become larger, the degree should be higher. Since multimodality implies that at least two peaks (and at least one valley) exist, it is strongly related to non-uniform intervals (considering cases (b) and (c)). Thus we measure the multimodality degree of the data in an interval $[a,b]$ as the distance from uniformity. Let $F_U(x)$ be the cdf of the uniform distribution in $[a,b]$. Then the value $d=\max\limits_{x\in X(a,b)} (|F(x)-F_U(x)|)$ computes the maximum distance (deviation) of the ecdf from uniformity. In plots we denote as $MD(x_{MD}, F(x_{MD}))$ the point of maximum distance. 

\begin{figure}[H]
	\hspace{-0.8cm}
	\begin{subfigure}[b]{0.63\linewidth}
		\includegraphics[width=\linewidth]{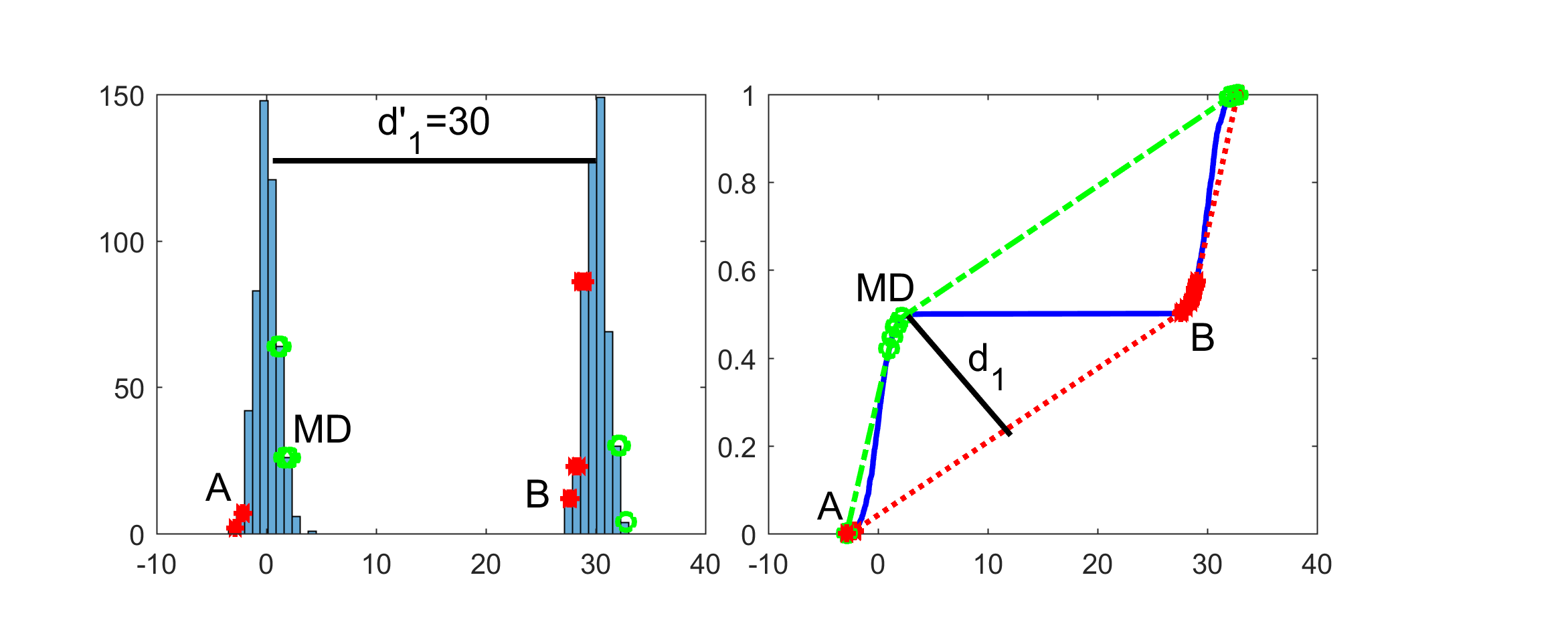}
	\end{subfigure} 
	\hspace{-1.28cm}
	\begin{subfigure}[b]{0.63\linewidth}
		\includegraphics[width=\linewidth]{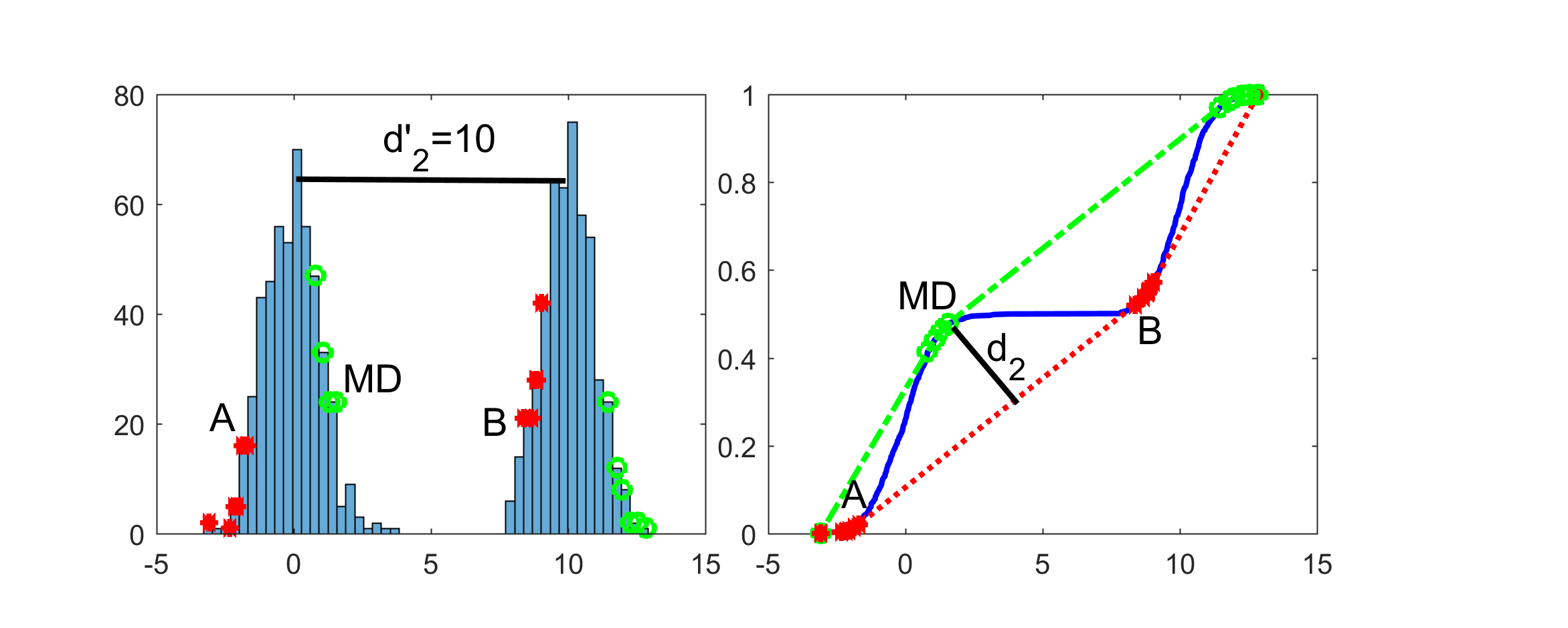}
	\end{subfigure}
	\vspace{-0.6cm}
	\caption*{(a) Closer peaks demonstrate a lower degree of non-uniformity.}
	\hspace{-0.8cm}
	\begin{subfigure}[b]{0.63\linewidth}
		\includegraphics[width=\linewidth]{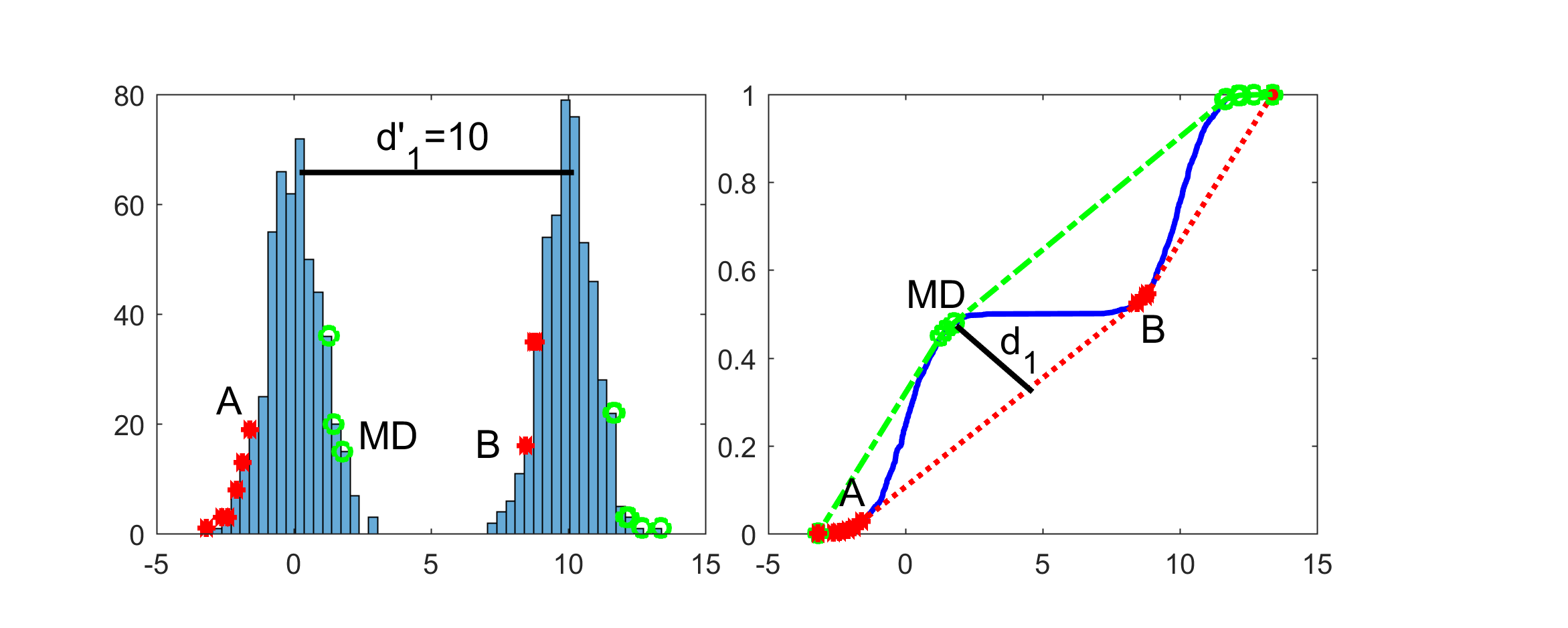}
	\end{subfigure} 
	\hspace{-1.28cm}
	\begin{subfigure}[b]{0.63\linewidth}
		\includegraphics[width=\linewidth]{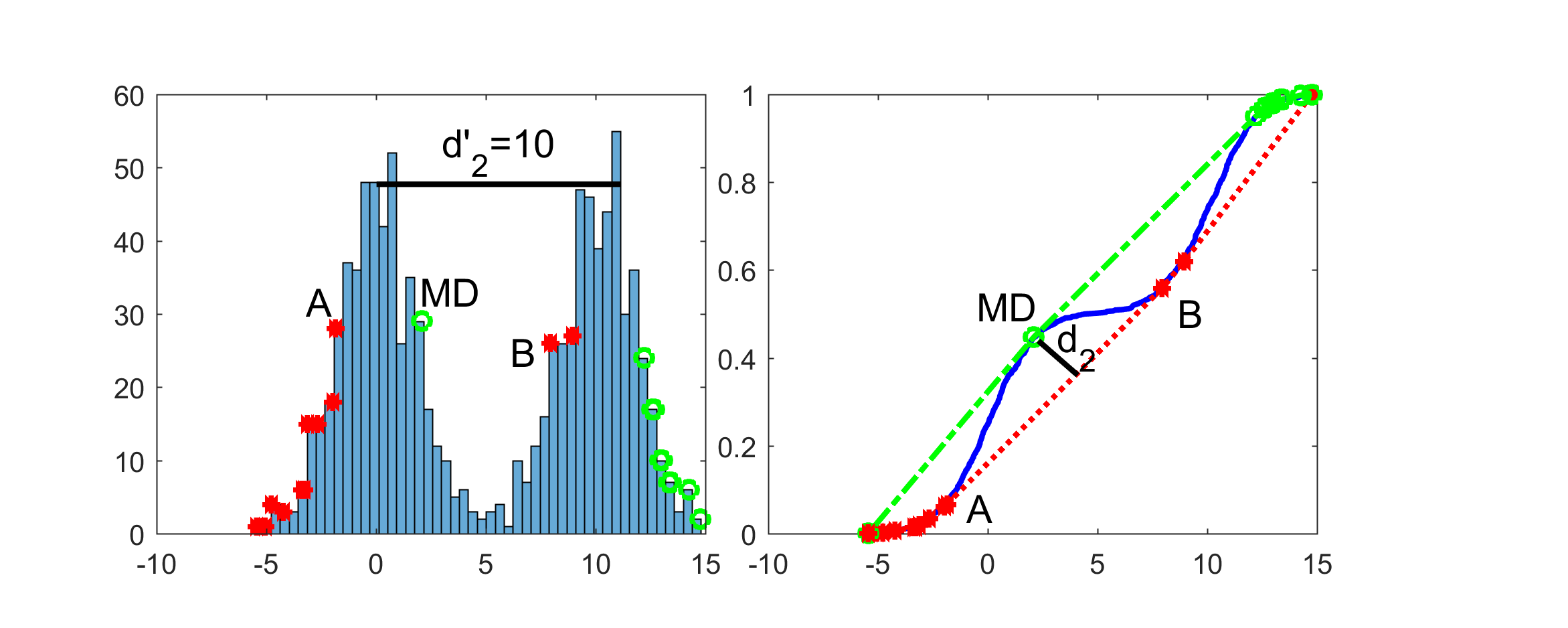}
	\end{subfigure}	
	\caption*{(b) Smaller valley depth corresponds to lower degree of non-uniformity.}
	\caption{Histogram and ecdf plots of bimodal datasets with varying peak distances and valley depths. The black segments on the pdfs correspond to the horizontal distances ($d'_1$ and $d'_2$) between the two peaks, while on the ecdfs correspond to the max distances ($d_1$ and $d_2$) of $MD$ from line segment $AB$.}
	\label{figure:fig6}
\end{figure}

Fig.~\ref{figure:fig6}a shows two bimodal datasets with peaks at $0$ and $30$ (left) and at $0$ and $10$ (right). Both datasets are multimodal due to non-linear ecdf segments $AB$. The black segments on the ecdf plots illustrate the maximum distances ($d_1$ and $d_2$) of $MD$ from the line segment connecting $(x_A, F(x_A))$ to $(x_B, F(x_B))$, with $d'_1$ and $d'_2$ representing the distances between the peaks on the pdf plots. The peaks in the dataset on the right are closer ($d'_2 < d'_1$), which is also evident on the ecdf plots where $d_2 < d_1$.
Valley depth is also related to the distance from uniformity as shown in Fig.~\ref{figure:fig6}b. Two bimodal datasets with equally spaced peaks (distance = $10$) are illustrated. The left histogram shows a deeper valley than the right, with the deeper valley corresponding to a higher degree of non-uniformity on the ecdf plots ($d_1 > d_2$).

In what concerns the location of the maximum deviation ($MD$) point, in case there exists a single valley in $[a,b]$, the $MD$ point will be very close to the valley point. In particular, if $a, b$ are successive gcm points, then a sequence of increasing (containing $a$), decreasing (containing $MD$) and again increasing (containing $b$) parts is formed. It is clear that $[x_{MD}, b]$ defines the valley region, since a valley exists between a decreasing part and an increasing part. 
Similarly, if $a,b$ are successive lcm points, the valley region will be $[a, x_{MD}]$, since a sequence of decreasing (containing $a$), increasing (containing $MD$), decreasing (containing $b$) parts exists. 
Fig.~\ref{figure:fig4a} and Fig.~\ref{figure:fig4b} illustrate these cases with histograms and ecdf plots, showing the relationship between the MD point and valley regions. In Fig.~\ref{figure:fig4a}, $A$ and $B$ are gcm points, thus between $MD$ and $B$ a valley is identified, while in Fig.~\ref{figure:fig4b} a valley is identified between $A$ and $MD$, since $A$ and $B$ are lcm points.

\vspace{-0.2cm}

\section{UU-test for Unimodal Data Modeling} \label{UUtest}

As mentioned in Section \ref{Valleys}, uniform ecdf segments between successive gcm and lcm points indicate the absence of density valleys, thus unimodality is suggested (case (a)). UU-test \cite{chasani2022uu} is a method which decides data unimodality by exploiting the above property of uniformity (i.e., ecdf linearity) between gcm and lcm points. It aims to construct a piecewise linear ($PL$) approximation of the ecdf by exploiting the gcm-lcm points of the ecdf. It takes the $GL$ set (ordered union of sets of gcm and lcm points) and tries to construct a $PL$ function, which is unimodal (gcm points preceding lcm points).
For each ordered subset $S$ of $GL$, $PL_S(x)$ is unimodal. 
However, in order for $PL_S(x)$ to be a good approximation of the ecdf, it should also model the data sufficiently (i.e., $PL_S(x)$ should be sufficient). This means that the data in each interval  $[s_i, s_{i+1}], s_i \in S$ are well-fitted by the uniform distribution as decided by a uniformity test. 

In case UU-test succeeds in finding such a solution, it also provides a statistical model of the unimodal data in the form of a Uniform Mixture Model (UMM). If $S$ is the final set of points returned by UU-test, the cdf of the statistical model is $PL_S(x)$, which is both unimodal and sufficient approximation of the ecdf. Since the cdf model is piecewise linear, it defines a UMM in which each component is the uniform distribution. More specifically, let $S=\{s_1,\ldots, s_{M+1}\}$, then a UMM with $M$ components is defined, where each component $i$ is uniformly distributed in the interval $[s_i,s_{i+1}]$, $(i=1,...,M)$. If $N$ is the size of $X$ and $N_i$ is the number of data points in each interval $[s_i,s_{i+1})$, then the UMM pdf is defined as follows \cite{chasani2022uu}:

\begin{displaymath} 
p(x)= \sum_{i=1}^{M} \frac{\pi_i }{s_{i+1}-s_i} I(x\in [s_i,s_{i+1})), \quad \pi_i=N_i/N
\end{displaymath}

In Fig.~\ref{figure:fig2}, the gcm points precede lcm points and the ecdf segments between gcm/lcm points are linear. Thus, UU-test can determine a $PL$ approximation of the ecdf.
In case UU-test fails, it ends deciding multimodality. This occurs when there exist intervals where data is multimodal, specifically when the set \( X(a, b) \) is non-uniform ($a, b$ being successive gcm or lcm points), indicating density valleys (case (b) and (c) in Section \ref{Valleys}). We use these non-uniform intervals to detect valley points and split the data into unimodal subsets in order to build a unimodal mixture model.

\section{The Unimodal Mixture Model (UDMM)} \label{UDMM}

In this section, we propose a method that builds a statistical mixture model for modeling univariate multimodal data. In this model, each component is unimodal as determined by UU-test for unimodality. First, we present a technique, called \emph{UniSplit}, that splits multimodal data into unimodal sets. To achieve this, we identify an interval with high degree of multimodality and then compute an appropriate valley point inside this interval. Based on the computed valley points, we recursively partition the data until unimodal segments are obtained. Finally, we provide the formulation for the Unimodal Mixture Model (UDMM) where each component constitutes a statistical model of a unimodal subset in the form of a uniform mixture model. 

\subsection{The UniSplit Algorithm}

Based on Section \ref{Valleys}, a dataset is characterized as multimodal, when at least one non-uniform interval $[a,b]$ defined by successive gcm or lcm points of the ecdf exists. In that case, at least one valley is noted inside the interval.
In our method, we aim to compute valley points in the density of multimodal datasets, thus we need to detect non-uniform intervals between successive gcm or lcm points in the ecdf. These intervals constitute \emph{candidate splitting intervals}, since they contain at least one valley. To detect candidate splitting intervals, the UU-test algorithm is applied, which utilizes a uniformity test (Kolmogorov-Smirnov \cite{dodge2008kolmogorov}), to decide whether a set of points follows the uniform distribution or not. As happens with every statistical test, the uniformity test requires a user-defined statistical significance level as input (we use the value equal to $0.01$ in our experiments). We should note here that apart from the uniformity significance level, our approach does not include any other user specified hyperparameters.

Our method starts by calling UU-test that takes the initial dataset $X$ as input. In case $X$ is unimodal and since no valley points are detected in unimodal datasets, the algorithm terminates and returns the corresponding UMM. Let $G$ and $L$ be the ordered sets of gcm and lcm points respectively, and $GL$ be the ordered union of them. Let also $maxG$ and $minL$ be the maximum value of $G$ and minimum value of $L$, respectively. In case $X$ is multimodal, we search for non-uniform intervals defined by successive gcm or lcm points to detect valley points (cases (b) and (c) in Section \ref{Valleys}). 

A special case occurs when $maxG < minL$, i.e., all gcm points precede all lcm points. If $X(maxG, minL)$ is uniform (linear ecdf) then it is ensured that no valleys exist in $[maxG, minL]$. Otherwise, we compute the gcm set $G'$ and lcm set $L'$ of $X(maxG, minL)$ to detect possible valley points in non-uniform intervals defined by successive gcm (or lcm) points in $G'$ (or in $L'$). Thus, we augment the original $GL$ set with the new gcm and lcm points, $GL := G \cup G' \cup L' \cup L$.

Based on the computed $GL$ set, UU-test detects and finally returns a set $I$ of candidate splitting (multimodal) intervals where at least one valley exists. Next, we determine the multimodality degree of each candidate interval and select the one with the highest degree, called as the 
\emph{best splitting interval}. Let $T = [a^*,b^*]$ be the best splitting interval. If $X(a^*,b^*)$ is unimodal, a single valley is formed in $T$ (case (b)), otherwise, multiple valleys are detected (case (c)).

In case of a single valley in $T = [a^*,b^*]$, the following strategy is used to determine a point in the valley region. We compute the MD point of $T$ and subsequently, use the $x_{MD}$, $a^*$ and $b^*$ values to compute a valley point. If $a^*, b^*$ are gcm points, the valley point lies in the middle of $x_{MD}$ and $b^*$. Since $MD$ is on the decreasing part of the mode and $b^*$ is on the increasing part of the next mode, a valley is formed between them, thus their middle point seems a reasonable location for the valley point. Similarly, if $a^*, b^*$ are lcm points, a valleys exists between the decreasing part ($a^*$) of the mode and the increasing part ($MD$) of the next mode, thus we compute the valley point as the middle point between $a^*$ and $x_{MD}$.
In Fig.~\ref{figure:fig4} the best splitting interval $T=[x_A,x_B]$ and the $MD$ point of $T$ are illustrated. In Fig.~\ref{figure:fig4a}, $A$ and $B$ are gcm points. On the histogram plot a valley exists between $MD$ and $B$, thus the middle point ($vp$) between $x_{MD}$ and $x_B$ is considered as a reasonable location for the valley point. Similarly, in Fig.~\ref{figure:fig4b} $A, B$ are lcm points with the average of $x_A$ and $x_{MD}$ denoting the $vp.$

In case where $X(a^*,b^*)$ is multimodal, multiple valleys exist in $T = [a^*,b^*]$. For an accurate valley point computation, we aim at detecting an interval with a single valley. Thus, we focus on the multimodal set $X(a^*,b^*)$ and work recursively, until we detect a non-uniform and unimodal interval. Such an interval will contain a single valley, thus we can follow the previously described methodology to compute a valley point.
Fig.~\ref{figure:fig5a} shows the histogram and ecdf plots of a multimodal dataset with its best splitting interval being \([x_A, x_F]\), identified as non-uniform and multimodal. Focusing on $X(x_A,x_F)$ (Fig.~\ref{figure:fig5b}), three candidate intervals are identified: \([x_A, x_B]\), \([x_G, x_C]\), and \([x_D, x_E]\). Among these, \(T=[x_G, x_C]\) demonstrates the highest degree of non-uniformity (largest distance of the ecdf of $X(x_G,x_C)$ from the line segment $GC$) and is unimodal, since a single peak is formed (histogram in Fig.~\ref{figure:fig5c}). Thus, $T$ contains a single valley, making it the final splitting interval for valley point computation. \(MD\) on the ecdf plot (Fig.~\ref{figure:fig5c}) is a close point to the valley region and helps us compute the valley point.
Algorithm \ref{algo:algo1} presents the steps of computing a valley point of a univariate multimodal dataset $X$. It takes $X$ as input and returns an appropriate valley point ($vp$).

After computing a valley point $vp$, we split the data into two subsets: a left subset $X_L$ (points on the left of $vp$) and a right subset $X_R$ (points on the right of $vp$). Then, the method runs recursively on each subset, until all obtained subsets are unimodal. The whole method (UniSplit algorithm) is described in Algorithm \ref{algo:algo2}, which takes a univariate dataset $X$ and a list of valley points ($vp\_list$) as input and returns an updated $vp\_list$ that partitions the data domain into adjacent unimodal intervals.

 \begin{algorithm}[!t]
	\caption{$vp$ = find\_vp$(X)$ // $X$ is multimodal} 
	\label{algo:algo1}
	\begin{algorithmic} 
		\State{Compute $GL$ set of $X$}
		\State{$I \leftarrow$ set of candidate splitting intervals of $GL$}
		\State{$T=[a^*,b^*] \leftarrow$ best splitting interval}
		\If{$X(a^*,b^*)$ is unimodal}
		\State{$x_{MD} \leftarrow$ compute $MD$ point of $T$}
		\If{$a^*,b^*$ gcm points}
		\State{$vp \leftarrow \frac{x_{MD}+ b^*}{2}$}
		\Else \quad // $a^*,b^*$ lcm points
		\State{$vp \leftarrow \frac{a^* + x_{MD}}{2}$}
		\EndIf
		\State{\Return{$vp$}}
		\Else \quad // $X(a^*,b^*)$ is multimodal
		\State{$vp \leftarrow$ find\_vp$(X(a^*,b^*)$)}
		\EndIf
	\end{algorithmic}
\end{algorithm}

\subsection{Merging Adjacent Intervals}

It should be noted that there exist cases where \emph{oversplitting} may occur, due to low density variations at the tails of unimodal subsets. This results in unnecessary splittings that define subsets with a small number of data points. To tackle this issue, we follow the typical merging procedure: Let our dataset $X$ has been splitted into $R$ adjacent unimodal subsets:  $X=\{X_1,X_2,...,X_R\}$. We iteratively merge the two first sets into one set and check its unimodality. In case it is unimodal we replace the two sets with their union, otherwise we merge the next two sets and repeat the procedure. The iterations stop when there is no unimodal union of successive sets. In this way a minimal unimodal partition is obtained, i.e., there is no union of successive subsets resulting in a unimodal set. 

 \begin{algorithm}[!t]
	\caption{$vp\_list$=UniSplit($X,vp\_list$)}
	\label{algo:algo2}
	\begin{algorithmic} 		
		\State{result $\leftarrow$ UU-test($X$)}
		\If{result = unimodal}
		\State{\Return{$vp\_list$}}
		\EndIf
		\State{$vp \leftarrow$ find\_vp$(X)$}
		\State{$vp\_list \leftarrow vp\_list \cup \{vp\}$}
		\State{$X_L \leftarrow$ $X(x_1, vp), \quad X_R \leftarrow X(vp, x_N$)}
		\State{$vp\_list \leftarrow$ UniSplit$(X_L,vp\_list)$}
		\State{$vp\_list \leftarrow$ UniSplit$(X_R,vp\_list)$}
		\State{\Return{$vp\_list$}}	
		\State{// First call: $vp\_list$=UniSplit($X,\emptyset$)}		
	\end{algorithmic}
\end{algorithm}

\begin{figure}[h]
	\centering 
	\begin{subfigure}[b]{0.3\linewidth}
		\includegraphics[width=\linewidth]{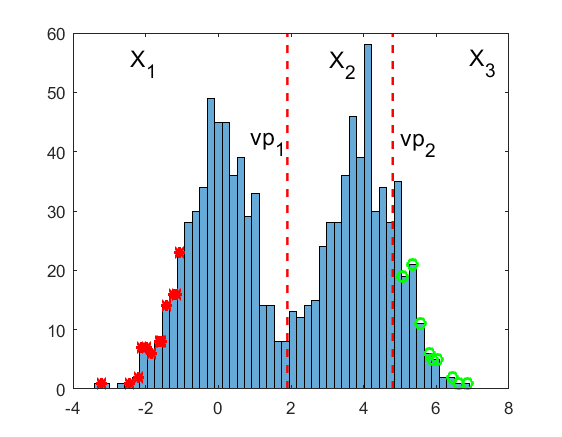}
		\caption{}
		\label{figure:fig9a}
	\end{subfigure}
	\begin{subfigure}[b]{0.3\linewidth}
		\includegraphics[width=\linewidth]{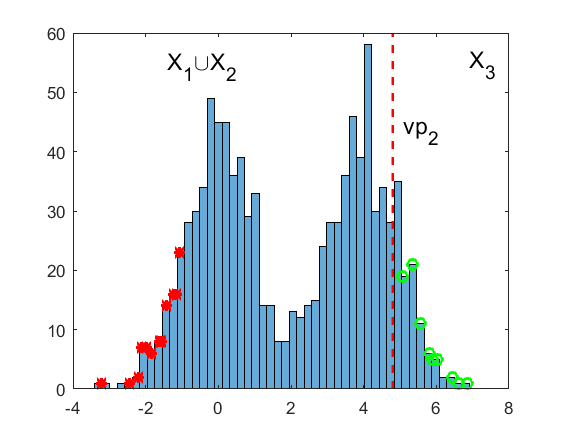}
		\caption{}
		\label{figure:fig9b}
	\end{subfigure}  
		\begin{subfigure}[b]{0.3\linewidth}
		\includegraphics[width=\linewidth]{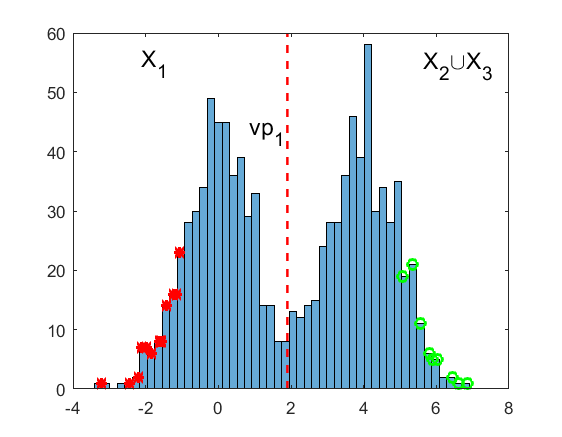}
		\caption{}
		\label{figure:fig9c}
	\end{subfigure}
	\caption{(a) Bimodal dataset with two computed valley points by UniSplit. (b) Omitting $vp_1$ leads to a multimodal set $X_1 \cup X_2$, thus $vp_1$ is necessary. (c) Merging $X_2$ and $X_3$ (omitting $vp_2$) leads to a unimodal set, thus $vp_2$ can be deleted. $vp_1$ is the final valley point.}
	\label{figure:fig9}
\end{figure}

Fig.~\ref{figure:fig9a} illustrates the histogram of a bimodal dataset with a single density valley. However, two valley points ($vp_1,vp_2$) have been determined by UniSplit method with the resulting unimodal subsets being $X_1$, $X_2$ and $X_3$. In Fig.~\ref{figure:fig9b} we merge sets $X_1$ and $X_2$ (by omitting $vp_1$) resulting to a multimodal set $X_1 \cup X_2$. This means that $vp_1$ is a required split point and cannot be omitted. Next, we merge $X_2$ with $X_3$ (omitting $vp_2$), which results to a unimodal set $X_2 \cup X_3$ (Fig.~\ref{figure:fig9c}). In such case, we delete $vp_2$ and our final solution contains a single valley point ($vp_1$).

\subsection{Computational Complexity}

The computational complexity mainly depends on determining the gcm/lcm points of the ecdf, which can be computed in \(O(n \log n)\) using the convex hull of the ecdf plot \cite{chasani2022uu}. In case the data is unsorted, an additional \(O(n \log n)\) is needed. Once the gcm/lcm points are computed, calculating the multimodality degree of a subset requires \(O(n)\), and the valley point is computed in \(O(1)\). Thus, computing the first valley point has a total complexity of \(O(n \log n)\). As the method iterates through subsets after each split, with far fewer splits than \(n\), the overall complexity remains \(O(n \log n)\). Additionally, the merging procedure incurs \(O(n \log n)\) complexity due to the UU-test for unimodality.

\subsection{UDMM formulation}

Based on the result of the UniSplit algorithm which splits multimodal data into unimodal subsets, a mixture model can be defined with each component modeling the unimodal data of each subset. More specifically, given a univariate dataset $X$, we first apply UniSplit method to obtain unimodal subsets of $X$. Then we employ UU-test to generate a UMM that models each unimodal set. Thus we obtain a hierarchical statistical model in the form of a \textit{mixture of UMMs}, where each component is unimodal. We call such a model as \textit{Unimodal Mixture Model} (UDMM). Let we split $X$ into $K$ unimodal subsets, i.e., $X=\{X_1,...,X_K\}$. Thus we can build a UDMM with $K$ components where each component $j$ is unimodal with $j=1,...,K$. 

Let $N$ be the size of $X$ and $N_j$ be the size of $X_j$. For each unimodal subset $X_j$, UU-test provides the set $S_j=\{s_1^j,\ldots, s_{M_j+1}^j\}$. Then a UMM with $M_j$ components is computed for $X_j$, where each UMM component $i$ is uniformly distributed in the range $[s_i^j,s_{i+1}^j]$, $(i=1,...,M_j)$. Let $N_{ij}$ be the the number of data points of $X_j$ in each interval $[s_i^j,s_{i+1}^j]$. The UDMM pdf of the multimodal set $X$ is defined as follows \cite{chasani2022uu}: 

\begin{displaymath} 
p(x)=  \sum_{j=1}^{K} w_j \sum_{i=1}^{M_j} \frac{\pi_{ij} }{s_{i+1}^j-s_i^j} I(x\in [s_i^j,s_{i+1}^j))  , \quad w_j = \frac{N_j}{N}, \quad \pi_{ij}=\frac{N_{ij}}{N_j} 
\end{displaymath}

It should be noted that the computed UDMM could also be used to \emph{generate synthetic data samples following the same multimodal distribution as the original dataset}.

\section{Experimental Results} \label{Experiments}

This section presents the experimental evaluation of our method across various tasks. At first, the modeling performance of UDMM was assessed using synthetic and real datasets. Next, its effectiveness in splitting tasks, mode estimation, and splitting quality was evaluated. UDMM’s applicability to image segmentation based on pixel intensity was tested, followed by its use as a probability density model in the Naive Bayes \cite{bishop2006pattern} classification method, where class distributions of each feature are modeled by a UDMM. Finally, we provide examples involving noise and outliers, demonstrating the robustness of our method and discuss the impact of the statistical significance level ($\alpha$) on our method.

\subsection{Modeling Multimodal Data with UDMM}

We conducted a series of experiments using synthetic and real datasets to evaluate the statistical modeling capabilities of UDMM against GMM, KDE and GMDEB\footnote{GMDEB is implemented using the mclustAddons package \cite{scrucca2016mclust, scrucca2023model}.} \cite{scrucca2019transformation}. We have generated synthetic datasets by sampling from various univariate multimodal distributions defined as mixtures of different unimodal distributions: i) $N(\mu, \sigma,n)$: Gaussian (Normal) distribution with mean $\mu$ and standard deviation $\sigma$, ii) $U(a, b, n)$: uniform distribution between $a$ and $b$, iii) $Tr(l,d,u,n)$: triangular distribution with lower limit $l$, mode $d$ and upper limit $u$, iv) $St(\nu, l, s, n)$: Student’s t distribution with $\nu$ degrees of freedom, location $l$ and scale $s$, v) $C(l, s, n)$: Cauchy distribution with location $l$ and scale $s$, and vi) $\Gamma(k, \theta, l, n)$: Gamma distribution with shape $k$, scale $\theta$ and location $l$. In all cases, the parameter $n$ indicates the dataset size.

Specifically, the synthetic datasets were generated from 12 multimodal distributions (D1 - D12), as shown in Table~\ref{table:table2}. The size of each distribution is also presented as a multiple of $m$, where $m=100$. We also evaluated the four models on 9 real datasets \cite{JSSv097i09}. The size and the description of each real dataset is also provided in Table~\ref{table:table2}.

For each synthetic distribution, 50 datasets were generated, and the four models were fitted to each dataset. While UDMM automatically estimates the number of components, GMM requires this number as input. Two criteria were used for this task: BIC \cite{bishop2006pattern} and the silhouette score \cite{rousseeuw1987silhouettes}. For BIC, we fit the dataset under consideration using several GMMs with components \(k\) ranging from 1 to 10, and the GMM corresponding to the \(k\) value yielding the lowest BIC was considered the best solution for the dataset. For the silhouette score, \(k\) ranged from 2 to 10 (as silhouette does not support \(k = 1\)), with the \(k\) achieving the highest silhouette score selected as the best GMM solution.

Since GMDEB \cite{scrucca2019transformation} utilizes GMMs for density estimation, $k$ is estimated using BIC, as with the typical GMM.
For KDE, we used a Gaussian kernel, and considered two rules for the bandwidth estimation: Scott's rule \cite{scott2015multivariate} and Silverman's rule \cite{silverman1986density}.

\begin{table}[H]
	\renewcommand{\arraystretch}{1} 
	\caption{Characteristics of synthetic and real datasets.}
	\label{table:table2}
	\resizebox{\textwidth}{!}{%
		\begin{tabular}{lll}
			\hline
			Name               & \multicolumn{2}{l}{Parameters}                                                                                                                                                                \\ \hline
			\underline{Synthetic}          & \multicolumn{2}{l}{}                                                                                                                                                                          \\
			D1                 & \multicolumn{2}{l}{$N(0,1,5m) \cup N(6,1,8m)$}                                                                                                                                                \\
			D2                 & \multicolumn{2}{l}{$N(-1,0.8,20m) \cup N(4,1.5,25m)$}                                                                                                                                         \\
			D3                 & \multicolumn{2}{l}{$St(2,0,1,5m) \cup U(4,7,2m) \cup N(10,1,4m)$}                                                                                                                             \\
			D4                 & \multicolumn{2}{l}{$Tr(-5,-4,0,3m) \cup Tr(1,5,6,5m) \cup U(7,10,2m)$}                                                                                                                          \\
			D5                 & \multicolumn{2}{l}{$\Gamma(1,2,0,5m) \cup Tr(5,6,7,5m) \cup N(10,0.2,5m)\cup St(10,15,1,8m)$}                                                                                                  \\
			D6                 & \multicolumn{2}{l}{$C(0,2,m) \cup U(50,55,3m) \cup U(100,105,3m)\cup St(1,200,1,m)$}                                                                                                          \\
			D7                 & \multicolumn{2}{l}{$U(-1,1,10m) \cup U(2,7,12m)$}                                                                                                                                             \\
			D8                 & \multicolumn{2}{l}{$St(1,-10,1,2m) \cup St(2,0,1,3m) \cup St(1,5,1,3.5m)\cup St(3,15,1,2.5m) \cup St(5,20,1,4m)$}                                                                             \\
			D9                 & \multicolumn{2}{l}{$U(-20,-15,10m) \cup U(-10,0,25m) \cup U(1,10,30m)\cup U(12,14,20m) \cup U(20,50,15m) \cup U(55,60,5m)$}                                                                   \\
			D10                & \multicolumn{2}{l}{$U(-15,-7,50m) \cup N(-2,4,40m) \cup N(9,3,30m)\cup U(15,20,20m)$}                                                                                                         \\
			D11                & \multicolumn{2}{l}{$St(5,-2,1,2m) \cup N(5,0.5,2m) \cup U(7,10,2m) \cup \Gamma(2,3,12,2m) \cup U(25,30,2m) \cup Tr(40,45,50,2m) \cup Tr(55,56,60,2m)$}                                          \\
			D12                & \multicolumn{2}{l}{$St(1,-50,1,m) \cup C(0,2,m) \cup U(30,60,m)$}                                                                                                                             \\                                                                                                              
			\underline{Real}      & \underline{Size}      & \underline{Description}                                                                                                                                                              \\ 
			suicide            & $n = 86$  & Lengths of spells of psychiatric treatment undergone by control patients in a suicide study.   
			\\
			racial             & $n = 56$  & Proportion of white student enrollment in school districts in Nassau County (Long Island, New York), for the 1992-1993 school year.
			\\
			acidity            & $n = 155$ & Acidity index measured in a sample of lakes in the Northeastern United States.                                         
			\\
			faithful eruptions & $n = 272$ & Eruption duration of the Old Faithful Geyser in Yellowstone National Park, Wyoming, USA.
			\\
			faithful waiting   & $n = 272$ & Waiting time in between eruptions of the Old Faithful Geyser in Yellowstone National Park, Wyoming, USA.
			\\
			galaxy             & $n=82$    & Velocities of distant galaxies, diverging from our own galaxy.  
			\\
			enzyme             & $n=245$   & Distribution of enzymatic activity in the blood, for an enzyme involved in the metabolism of carcinogenic substances.    
			\\
			stamps             & $n=485$   & Thickness measurements (in millimeters) of unwatermarked used white wove stamps of the 1872 Hidalgo stamp issue of Mexico.
			\\
			geyser             & $n = 272$ & Interval times between the starts of the geyser eruptions on the Old Faithful Geyser.
			\\ \hline
		\end{tabular}%
	}
\end{table}

To evaluate the quality of the obtained UDMM, GMM, KDE and GMDEB solutions, we used the two-sample Kolmogorov-Smirnov (KS) test criterion. The two-sample KS test computes the maximum absolute difference between the ecdfs of two datasets. In the case of synthetic datasets, in each experiment we used a dataset generated from the ground truth distribution and compared it (using the two-sample KS test) with a dataset generated from each of the four models fitted on the generated dataset. In the case of real datasets we compared the original dataset with a dataset generated from each of the four fitted models. We repeated the above procedure $50$ times and obtained the average distance (KS statistic) and the average number of components ($k$) for each model. The smaller the distance provided by the KS test, the better the obtained statistical model. The results are presented in Table~\ref{table:table3}.

\begin{table}[h]
	\renewcommand{\arraystretch}{1} 
	\caption{Statistical model evaluation using the two-sample KS test (the lower the better). Bold values indicate the best model in each row. The ground truth number of components ($k^\star$) (in case of synthetic datasets) and the average estimated number of components ($k$) are also provided.}
	\label{table:table3}
	\resizebox{\textwidth}{!}{%
		\begin{tabular}{lrrrrrr||rrrrr}
			\hline
			\multicolumn{1}{r}{} & \multicolumn{6}{c}{Average KS statistic}                                                                                   &           & \multicolumn{4}{c}{Average number of components ($k$)} \\ \hline
			Name                 & GMM (BIC)            & GMM (Sil)            & KDE (Scott)          & KDE (Silverman)      & GMDEB                & UDMM                 & $k^\star$ & GMM (BIC)   & GMM (Sil)      & GMDEB & UDMM          \\ \hline
			\underline{Synthetic}            & \multicolumn{1}{l}{} & \multicolumn{1}{l}{} & \multicolumn{1}{l}{} & \multicolumn{1}{l}{} & \multicolumn{1}{l}{} & \multicolumn{1}{l||}{} &           &             &                &       &               \\
			D1                   & 0.031                & 0.031                & 0.026                & \textbf{0.025}       & 0.053                & 0.032                & 2         & \textbf{2}  & \textbf{2}     & 2.82  & \textbf{2}    \\
			D2                   & \textbf{0.013}       & \textbf{0.013}       & 0.014                & \textbf{0.013}       & 0.019                & 0.016                & 2         & \textbf{2}  & \textbf{2}     & 4.1   & \textbf{2}    \\
			D3                   & \textbf{0.026}       & 0.044                & 0.027                & 0.029                & 0.172                & 0.027                & 3         & 4.96        & 3.94           & 1.78  & \textbf{3.1}  \\
			D4                   & 0.030                & 0.037                & 0.030                & 0.027                & 0.047                & \textbf{0.025}       & 3         & 6.14        & \textbf{3}     & 4.26  & \textbf{3}    \\
			D5                   & 0.021                & 0.022                & 0.030                & 0.032                & 0.099                & \textbf{0.017}       & 4         & 7.26        & \textbf{4}     & 3.18  & \textbf{4}    \\
			D6                   & 0.037                & 0.061                & 0.032                & 0.032                & 0.267                & \textbf{0.031}       & 4         & 8.78        & 4.94           & 1.78  & \textbf{4.1}  \\
			D7                   & 0.021                & 0.042                & 0.025                & 0.025                & 0.022                & \textbf{0.020}       & 2         & 8.66        & 2.04           & 8.16  & \textbf{2}    \\
			D8                   & 0.026                & 0.140                 & 0.024                & 0.022                & 0.306                & \textbf{0.018}       & 5         & 8.84        & 2.62           & 1.38  & \textbf{5.06} \\
			D9                   & 0.016                & 0.079                & 0.009                & 0.010                & 0.017                & \textbf{0.007}       & 6         & 10          & 2              & 9.9   & \textbf{6}    \\
			D10                  & 0.011                & 0.049                & 0.007                & 0.007                & 0.011                & \textbf{0.006}       & 4         & 9.55        & 2              & 8.45  & \textbf{4.05} \\
			D11                  & 0.033                & 0.051                & 0.024                & 0.024                & 0.072                & \textbf{0.017}       & 7         & 8.28        & 6.24           & 3.78  & \textbf{7}    \\
			D12                  & 0.061                & 0.132                & 0.053                & 0.052                & 0.281                & \textbf{0.048}       & 3         & 7.32        & \textbf{3.02}  & 1.48  & \textbf{3.02} \\
			\underline{Real}        & \textbf{}            &                      &                      &                      &                      &                      &           &             &                &       &               \\
			suicide              & 0.097                & 0.135                & 0.098                & 0.094                & 0.115                & \textbf{0.011}       &           & 6           & 3              & 2     & 1             \\
			racial               & 0.145                & 0.145                & 0.372                & 0.381                & 0.225                & \textbf{0.120}       &           & 2           & 2              & 1     & 1             \\
			acidity              & \textbf{0.082}       & \textbf{0.082}       & 0.100                & 0.100                & 0.165                & 0.090                &           & 2           & 2              & 2     & 1             \\
			faithful eruptions   & 0.070                & 0.070                & 0.080                & 0.080                & 0.157                & \textbf{0.050}       &           & 2           & 2              & 2     & 2             \\
			faithful waiting     & 0.070                & 0.070                & 0.080                & 0.090                & 0.168                & \textbf{0.050}       &           & 2           & 2              & 2     & 2             \\
			galaxy               & 0.121                & 0.121                & \textbf{0.048}       & 0.085                & 0.390                & 0.109                &           & 3           & 3              & 1     & 2             \\
			enzyme               & 0.085                & 0.093                & 0.220                & 0.240                & 0.155                & \textbf{0.044}       &           & 2           & 3              & 2     & 2             \\
			stamps               & 0.475                & 0.565                & 0.478                & 0.478                & 0.099                & \textbf{0.058}       &           & 3           & 2              & 2     & 1             \\
			geyser               & \textbf{0.051}       & \textbf{0.051}       & 0.058                & 0.073                & 0.161                & 0.062                &           & 2           & 2              & 2     & 1             \\ \hline
		\end{tabular}%
	}
\end{table}

The results in Table~\ref{table:table3} demonstrate that UDMM effectively models univariate multimodal data. While GMM and KDE excel for datasets generated by Gaussian distributions (e.g., D1 and D2), UDMM’s performance is comparable. In other cases, UDMM outperforms, accurately estimating the true number of components ($k^\star$), unlike GMM (BIC), which often overestimates, and GMDEB, which provides less accurate results. For real data, UDMM performs well except for acidity, galaxy, and geyser datasets, where it uses fewer components than GMM. However, it is noteworthy that UDMM achieves its performance using only a single component for the acidity and geyser datasets, while GMMs employ two components. Overall, UDMM is a successful statistical model for univariate multimodal data, correctly estimating components in synthetic datasets and providing accurate modeling solutions for real data with fewer components compared to other methods.

\begin{figure}[H]
	\centering 
	\begin{subfigure}[b]{0.5\linewidth}
		\includegraphics[width=\linewidth]{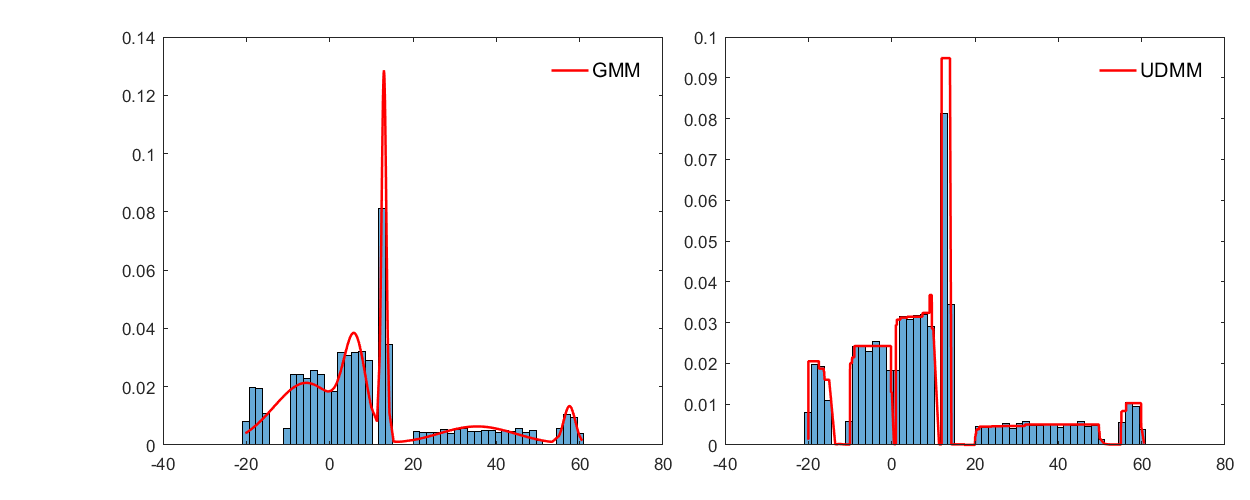}
		\caption*{D9}
		\label{figure:fig11a}
	\end{subfigure} 
	\hspace{-0.2cm}
	\begin{subfigure}[b]{0.5\linewidth}
		\includegraphics[width=\linewidth]{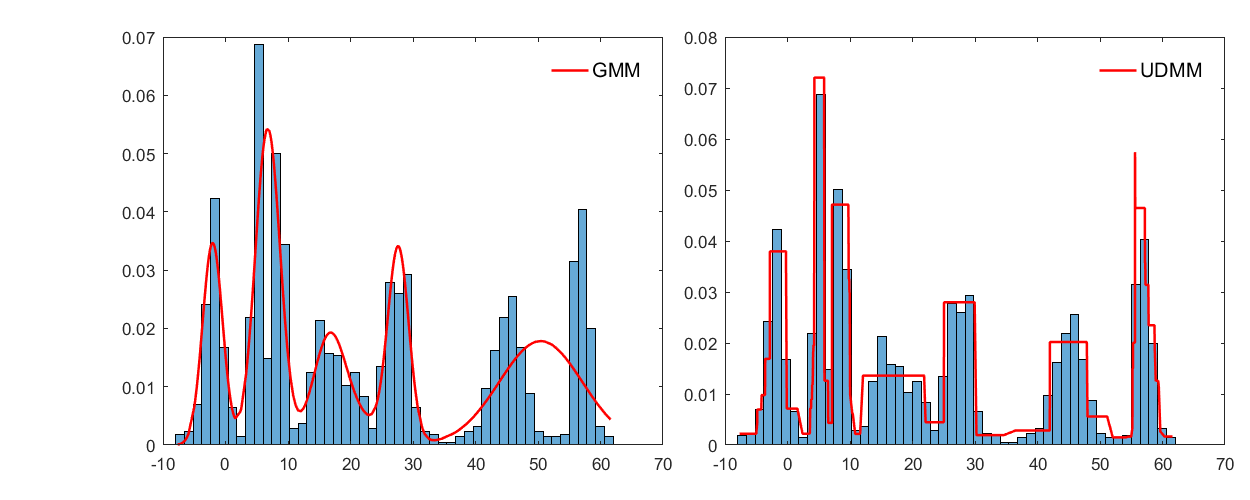}
		\caption*{D11}
		\label{figure:fig11c}
	\end{subfigure}
\end{figure}
\vspace{-0.8cm}
\begin{figure}[H]\ContinuedFloat
	\centering 
	\begin{subfigure}[b]{0.5\linewidth}
		\includegraphics[width=\linewidth]{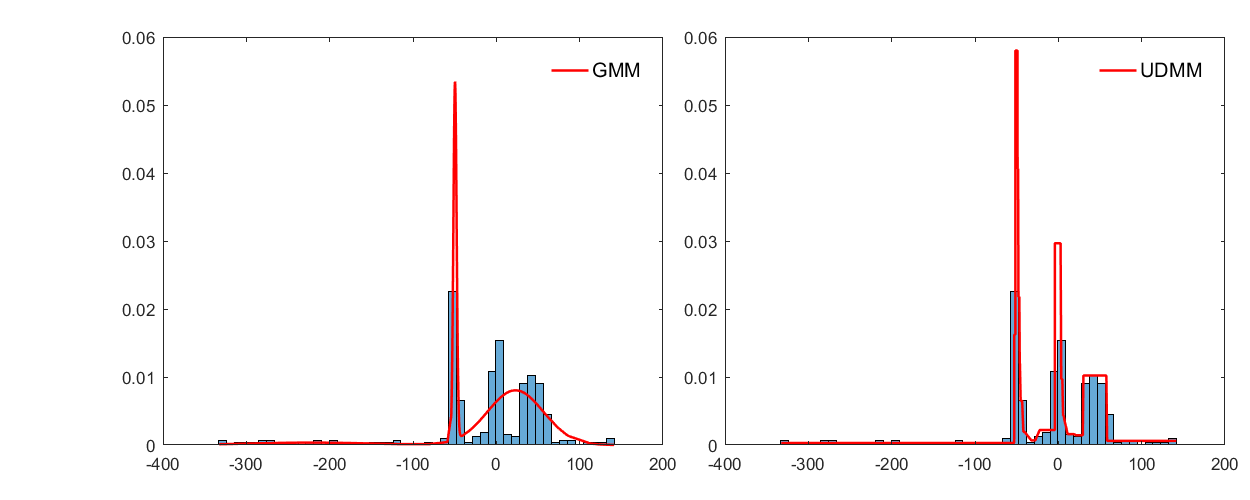}
		\caption*{D12}
		\label{figure:fig11d}
	\end{subfigure} 
	\hspace{-0.2cm}
	\begin{subfigure}[b]{0.5\linewidth}
		\includegraphics[width=\linewidth]{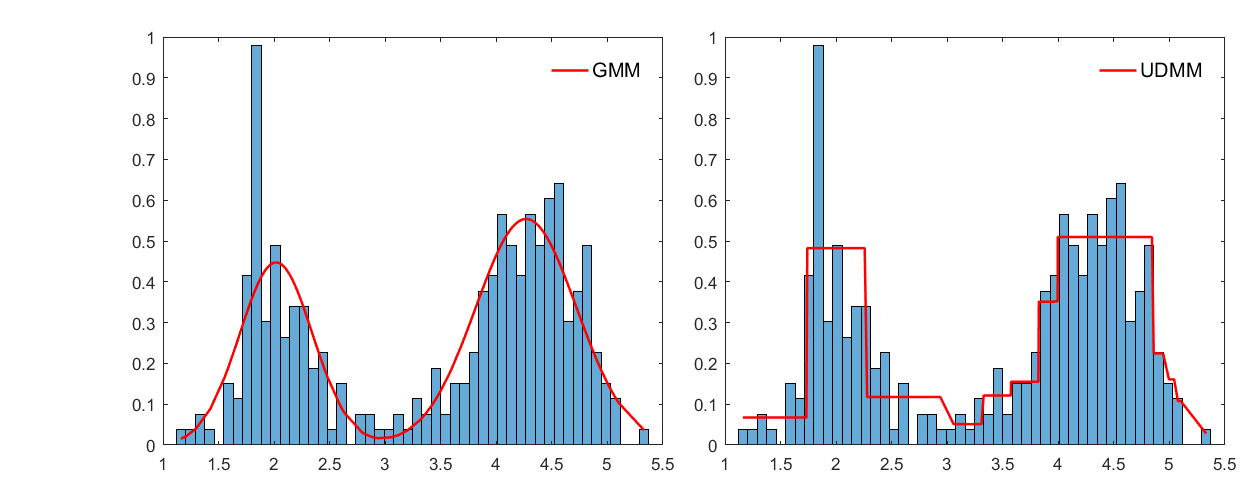}
		\caption*{faithful eruptions}
		\label{figure:fig11h}
	\end{subfigure}
	\caption{Examples of statistical model fitting results on several datasets using GMM and UDMM.}
	\label{figure:fig11}
\end{figure}

In Fig.~\ref{figure:fig11}, we present the histogram and pdf plots of the solutions provided by the two best-performing models, namely GMM (left plot) and UDMM (right plot), for some of the datasets from Table~\ref{table:table2}. For the synthetic datasets, GMMs were trained using the true number of components ($k^\star$), since the estimated number of components ($k$) for GMM (BIC) and GMM (Sil) is averaged in Table~\ref{table:table3}. For the real datasets, GMMs were trained using the minimum number of components provided by GMM (BIC) and GMM (Sil) in Table~\ref{table:table3}. It is evident that the obtained UDMMs constitute accurate statistical models for the datasets, whereas the GMMs do not always provide adequate solutions. For instance, in the plots of D11 and D12 in Fig.~\ref{figure:fig11}, although GMM uses the ground truth number of components ($k^\star=7$ and $k^\star=3$, respectively), it fails to accurately fit the two rightmost components. In contrast, UDMM successfully captures these components without requiring prior knowledge of $k^\star$.

\subsection{Multimodal Data Splitting}

We also assessed the performance of UniSplit on partitioning univariate synthetic data, focusing on the accurate estimation of the number of modes and the quality of data splitting. We compare UniSplit with TailoredDip\footnote{TailoredDip is implemented using the ClustPy package \cite{leiber2023benchmarking} in Python.} \cite{bauer2023extension}, FTC\footnote{FTC is implemented in Matlab as described in \cite{delon2006nonparametric}.} \cite{delon2006nonparametric}, mean shift\footnote{For mean shift we use the sklearn package in Python.} \cite{fukunaga1975estimation, cheng1995mean}, modclust\footnote{The implementation of modclust is available in \href{http://matematicas.unex.es/~jechacon}{http://matematicas.unex.es/~jechacon}.} \cite{chacon2019mixture} and MEM\footnote{MEM is implemented using the mclustAddons package \cite{scrucca2016mclust, scrucca2023model} in R.} \cite{li2007nonparametric, scrucca2021fast}. TailoredDip and FTC rely on exploiting unimodality as UniSplit does, while the remaining three methods (mean shift, modclust and MEM) focus on identifying modes and their corresponding clusters within a distribution, making them well-suited for modal clustering tasks.

We generated synthetic datasets by sampling from various univariate multimodal distributions (D13 - D22 in Table~\ref{table:table4}). For each distribution, $100$ datasets were created, and the six methods were applied to cluster the generated data. Ground truth clustering information was available for each dataset, thus the methods were evaluated in terms of splitting (clustering) using the Normalized Mutual Information (NMI) score. NMI ranges from $0$ to $1$, with values closer to $1$ indicating better clustering performance.

The parameters of each method are initialized as follows. For UniSplit and TailoredDip, the significance level is set to $0.01$.
TailoredDip also requires a \emph{factor} parameter, which defines the maximum difference in sample size during the merge test of two clusters, while FTC requires a segmentation parameter $e$, with large and small values resulting in coarse and finer segmentation, respectively. We have tuned both parameters taking into account the NMI value, and finally selected: $factor = 0.5$ and $e = 0$.
In mean shift, the bandwidth was calculated based on distances between points, scaling it according to a quantile (0.3) of nearest neighbor distances. Finally, for GMMs employed in modclust and MEM, we use the $k$ value (ranging from $1$ to $10$) that yields the lowest BIC score.

Table~\ref{table:table5} provides the average and standard deviation of NMI values, along with the ground truth ($k^\star$) and estimated number of modes ($k$) for each method across 100 datasets generated by each distribution. UniSplit outperforms most methods in both splitting performance (NMI) and estimating $k$. In Gaussian mixture distributions, such as D13 and D19, UniSplit's NMI values are slightly lower than the best-performing methods but remain close, with the estimated $k$ being closely to the ground truth. Interesting examples include D20, D21, and D22, where the number of uniform components increases. In these cases, UniSplit improves, while other methods, such as modclust and MEM, deteriorate. Overall, UniSplit shows robust performance, accurately estimating the number of modes across different multimodal distributions.

\subsection{Image Segmentation}

\begin{table}[t]
	\renewcommand{\arraystretch}{1} 
	\centering
	\small
	\caption{Characteristics of synthetic datasets.}
	\label{table:table4}
	\begin{tabular}{ll}
		\hline
		Name      & Distribution Parameters                                           \\ \hline
		D13  & $N(0,1.7,700) \cup N(5,1,500)$                                    \\
		D14  & $U(-1,3,300) \cup U(8,10,200)$                                    \\
		D15  & $Tr(0.8,1,5,1000) \cup Tr(3,7.8,8,1000)$                          \\
		D16  & $N(0,1,1000)$ (right part) $\cup N(4,1,1000)$ (left part)         \\
		D17  & $Tr(-3.3,1,2.5,1000) \cup N(4,1,1000)$                            \\
		D18  & $U(-2,0,200) \cup U(1,5,300) \cup U(6,7,450)$                     \\
		D19  & $N(0,1,500) \cup N(6,1,80) \cup N(12,1,500) \cup N(18,1,100)$     \\
		D20  & $N(0,1,500) \cup N(4,1,300) \cup N(11,1,500) \cup U(14,15,50)$    \\
		D21  & $N(0,1,500) \cup N(4,1,300) \cup U(10,11,100) \cup U(14,15,50)$   \\
		D22 & $N(0,1,500) \cup U(2.5,4,200) \cup U(10,11,100) \cup U(14,15,50)$ \\ \hline
	\end{tabular}%
\end{table}

\begin{table}[t]
	\renewcommand{\arraystretch}{1} 
	\caption{Partition evaluation of multimodal datasets. The average and standard deviation for NMI, the ground truth number of modes ($k^\star$) and the average number of detected modes ($k$) are provided.}
	\label{table:table5}
	\resizebox{\textwidth}{!}{%
		\begin{tabular}{lrrrrrr||rrrrrrr}
			\hline
			& \multicolumn{6}{c}{Mean NMI}                                                                                                               & \multicolumn{1}{l}{} & \multicolumn{6}{c}{Average Number of Detected Modes ($k$)}                         \\ \hline
			Distributions & UniSplit               & TailoredDip   & FTC                    & Mean shift             & Modclust               & MEM                    & $k^\star$            & UniSplit      & TailoredDip & FTC        & Mean shift & Modclust      & MEM        \\ \hline
			D13      & 0.78$\pm$0.02          & 0.71$\pm$0.07 & 0.77$\pm$0.03          & 0.71$\pm$0.07          & 0.75$\pm$0.03          & \textbf{0.79$\pm$0.02} & 2                    & \textbf{2}    & \textbf{2}  & 2.05       & 2.46       & \textbf{2}    & \textbf{2} \\
			D14      & \textbf{1.00$\pm$0.00} & 0.94$\pm$0.08 & \textbf{1.00$\pm$0.00} & 0.87$\pm$0.12          & 0.41$\pm$0.04          & 0.86$\pm$0.16          & 2                    & \textbf{2}    & \textbf{2}  & \textbf{2} & 2.56       & 5.58          & 2.44       \\
			D15      & \textbf{0.71$\pm$0.02} & 0.65$\pm$0.07 & 0.64$\pm$0.13          & 0.61$\pm$0.05          & 0.38$\pm$0.06          & 0.40$\pm$0.04          & 2                    & \textbf{2}    & \textbf{2}  & 1.97       & 2.83       & 4.74          & 4.29       \\
			D16     & \textbf{0.73$\pm$0.03} & 0.71$\pm$0.05 & 0.58$\pm$0.29          & 0.64$\pm$0.07          & 0.41$\pm$0.06          & 0.55$\pm$0.10          & 2                    & \textbf{2}    & \textbf{2}  & 1.8        & 2.7        & 4.2           & 3.16       \\
			D17      & \textbf{0.84$\pm$0.02} & 0.69$\pm$0.11 & 0.78$\pm$0.03          & 0.75$\pm$0.09          & 0.82$\pm$0.02          & \textbf{0.84$\pm$0.02} & 2                    & \textbf{2}    & \textbf{2}  & \textbf{2} & 2.54       & \textbf{2}    & \textbf{2} \\
			D18      & \textbf{0.99$\pm$0.01} & 0.92$\pm$0.11 & 0.96$\pm$0.04          & 0.91$\pm$0.02          & 0.51$\pm$0.04          & 0.80$\pm$0.04          & 3                    & \textbf{3}    & \textbf{3}  & 3.32       & 3.06       & 8.24          & 4.43       \\
			D19      & 0.97$\pm$0.03          & 0.89$\pm$0.07 & 0.97$\pm$0.04          & \textbf{0.99$\pm$0.01} & \textbf{0.99$\pm$0.01} & \textbf{0.99$\pm$0.01} & 4                    & 3.9           & 3.3         & 3.8        & 4.07       & \textbf{4}    & \textbf{4} \\
			D20      & 0.91$\pm$0.04          & 0.84$\pm$0.03 & 0.89$\pm$0.06          & 0.86$\pm$0.02          & \textbf{0.93$\pm$0.02} & \textbf{0.93$\pm$0.02} & 4                    & 3.7           & 3           & 3          & 3.01       & \textbf{3.96} & 3.93       \\
			D21      & \textbf{0.89$\pm$0.08} & 0.80$\pm$0.08 & 0.80$\pm$0.11          & 0.73$\pm$0.14          & 0.77$\pm$0.12          & 0.78$\pm$0.06          & 4                    & \textbf{3.92} & 3.76        & 3.5        & 3.5        & 4.56          & 5.48       \\
			D22     & \textbf{0.93$\pm$0.02} & 0.89$\pm$0.06 & 0.92$\pm$0.07          & 0.70$\pm$0.00          & 0.68$\pm$0.10          & 0.75$\pm$0.07          & 4                    & \textbf{4.02} & 3.92        & 3.88       & 3          & 5.92          & 5.66       \\ \hline
		\end{tabular}%
	}
\end{table}

A widely studied statistical modeling task concerns image segmentation where the objective is to identify and differentiate various objects or regions within an image based on pixel intensities. We have applied UniSplit, TailoredDip, FTC, mean shift, modclust and MEM to solve this task and tested their performance in estimating the number of segments and their ability to accurately segment the image. To apply the methods for rgb (colored) images, each rgb image is first converted to grayscale, thus a univariate dataset is obtained containing the gray values of the pixels. Then we applied each compared method to the resulting dataset and obtained a segmentation of the image, i.e., a partition of the pixels into subsets.

We tested the performance of the six methods on rgb images, where the ground truth number of colors can be easily determined through visual inspection. Once the ground truth value of colors ($k^\star$) has been specified, we used the k-means algorithm to obtain the ground truth partition for each image, which is subsequently used to evaluate the quality of the obtained solutions using the NMI score. The parameters of each method are set as they were in the previous experiments.

\begin{table}[H]	
	\renewcommand{\arraystretch}{1} 
	\caption{Image segmentation results: i) Estimated number of colors ($k$), ii) NMI values with respect to a ground truth solution obtained by applying k-means with the ground truth number of colors ($k^{\star}$).}	
	\label{table:table6}
	\resizebox{\textwidth}{!}{%
		\begin{tabular}{llrrrrrr}
			\hline
			Images                         & $k^\star$ / NMI & UniSplit       & TailoredDip    & FTC            & Mean shift     & Modclust & MEM            \\ \hline
			
			\multirow{2}{*}{France flag}   & $k^\star=3$     & $k=5$          & $k=6$          & $k=6$          & $k=4$          & $k=8$    & $k=3$          \\
			& NMI             & \textbf{0.969} & 0.967          & 0.967          & 0.960          & 0.736    & 0.740          \\ \hline
			
			\multirow{2}{*}{Europe flag}   & $k^\star=2$     & $k=2$          & $k=2$          & $k=2$          & $k=11$         & $k=2$    & $k=9$          \\
			& NMI             & 0.936          & \textbf{0.965} & 0.656          & 0.730          & 0.853    & 0.119          \\ \hline
			
			\multirow{2}{*}{Face}          & $k^\star=3$     & $k=3$          & $k=3$          & $k=3$          & $k=5$          & $k=5$    & $k=4$          \\
			& NMI             & \textbf{0.998} & 0.963          & 0.936          & 0.950          & 0.830    & 0.877          \\ \hline
			
			\multirow{2}{*}{Flower}        & $k^\star=6$     & $k=6$          & $k=6$          & $k=6$          & $k=3$          & $k=7$    & $k=6$          \\
			& NMI             & \textbf{0.998} & 0.996          & 0.992          & 0.770          & 0.851    & 0.995          \\ \hline
			
		\end{tabular}%
	}
\end{table}

In Table~\ref{table:table6} we present the NMI values and the obtained number of colors for each image as provided by the six methods. In the second column, we provide the ground truth value of colors ($k^\star$).
In general, the differences in the highest NMI values for each image are small, indicating that some methods provide similar segmentations. An interesting case is the flag of Europe, where mean shift and MEM fail to provide correct segmentation, detecting 11 and 9 colors, respectively, instead of the correct 2, while FTC, despite its correct estimation, it does not achieve the optimal NMI value. As shown in Table~\ref{table:table6}, it is clear that UniSplit achieves very high NMI values ($>0.93$) for all images and provides accurate or very close estimates of $k$ compared to $k^{\star}$.

\begin{figure}[H]
	\centering
	\begin{subfigure}[b]{\linewidth}
		\centering
		\includegraphics[width=0.13\linewidth]{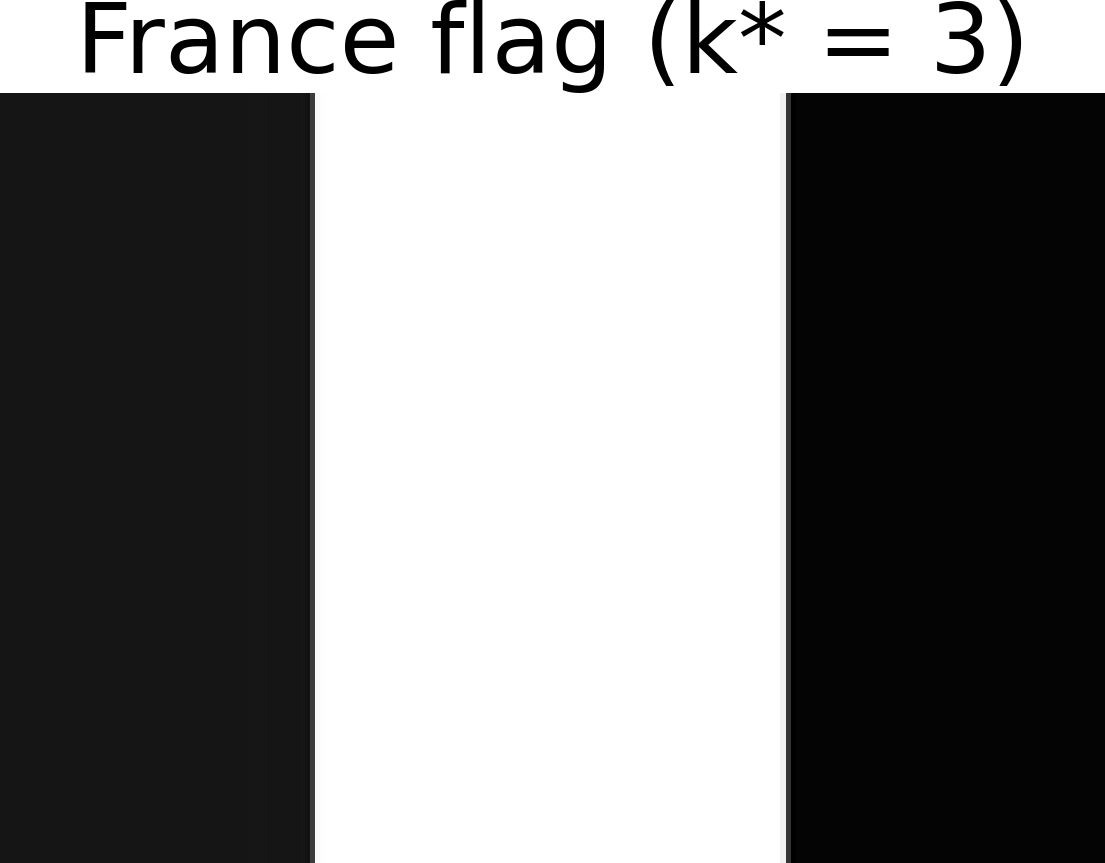}
		\includegraphics[width=0.13\linewidth]{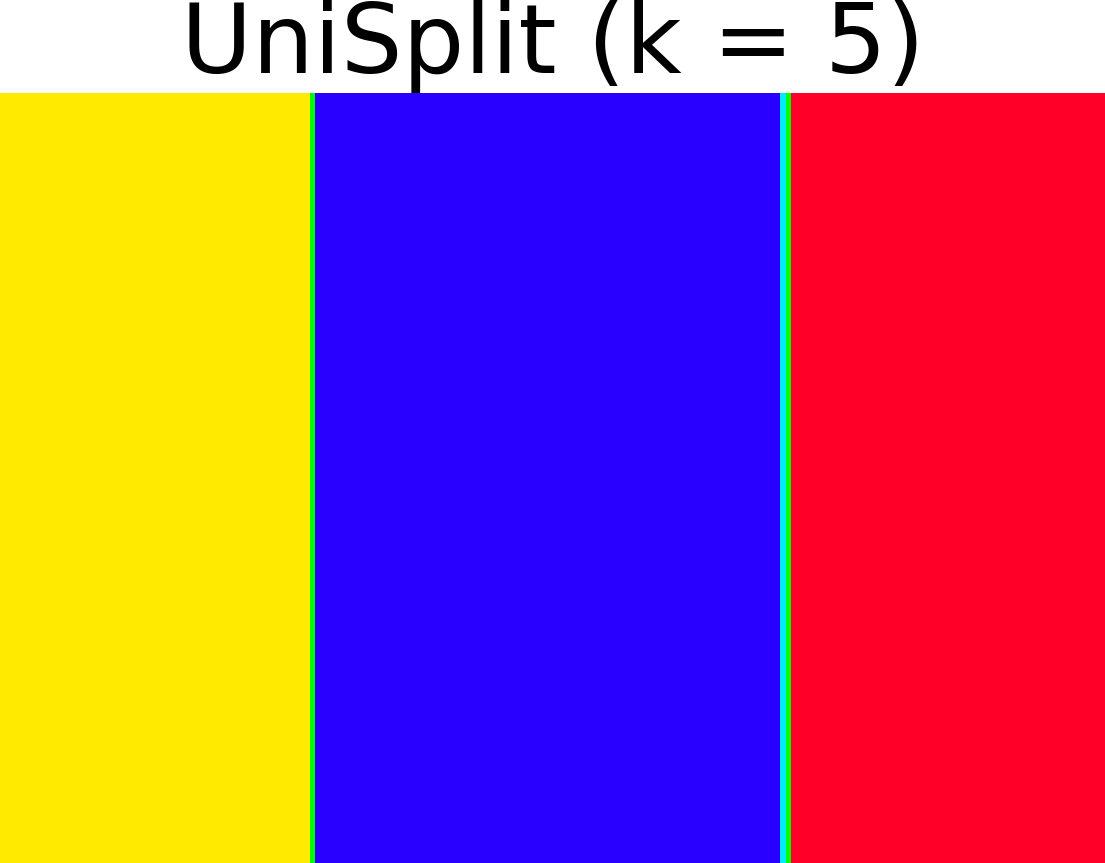}
		\includegraphics[width=0.13\linewidth]{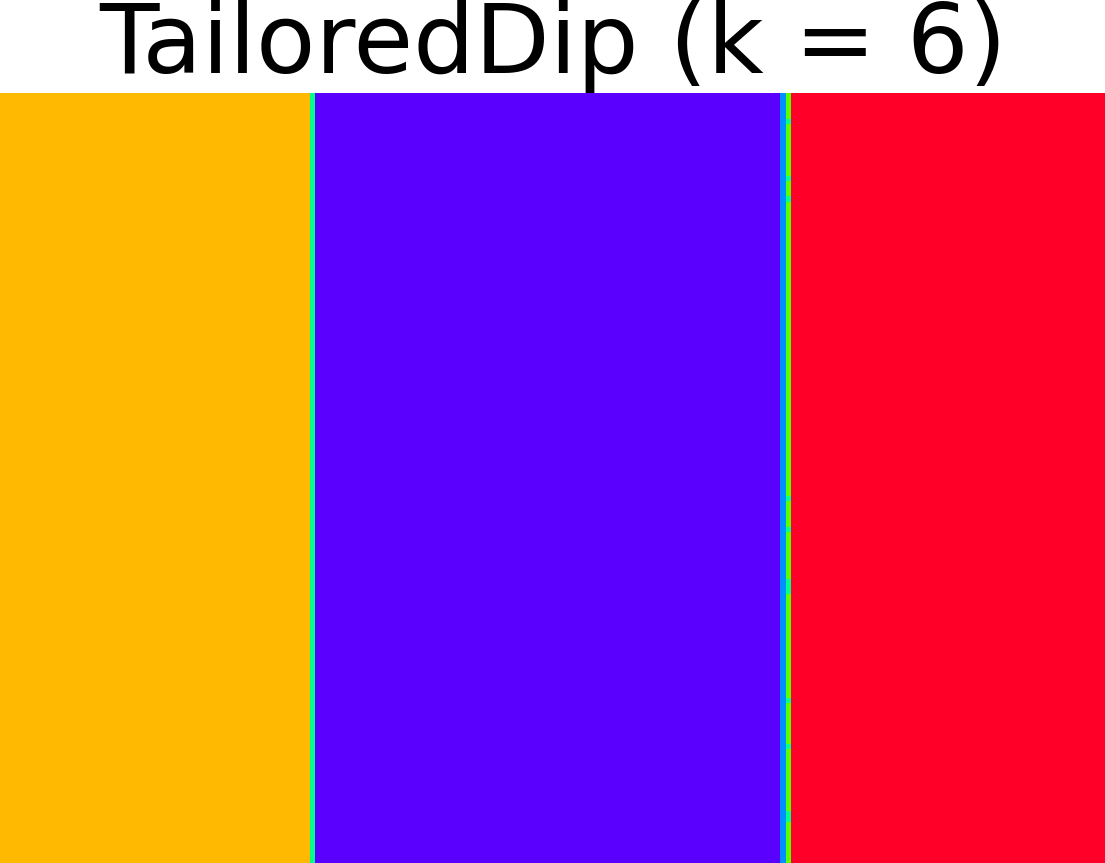}
		\includegraphics[width=0.13\linewidth]{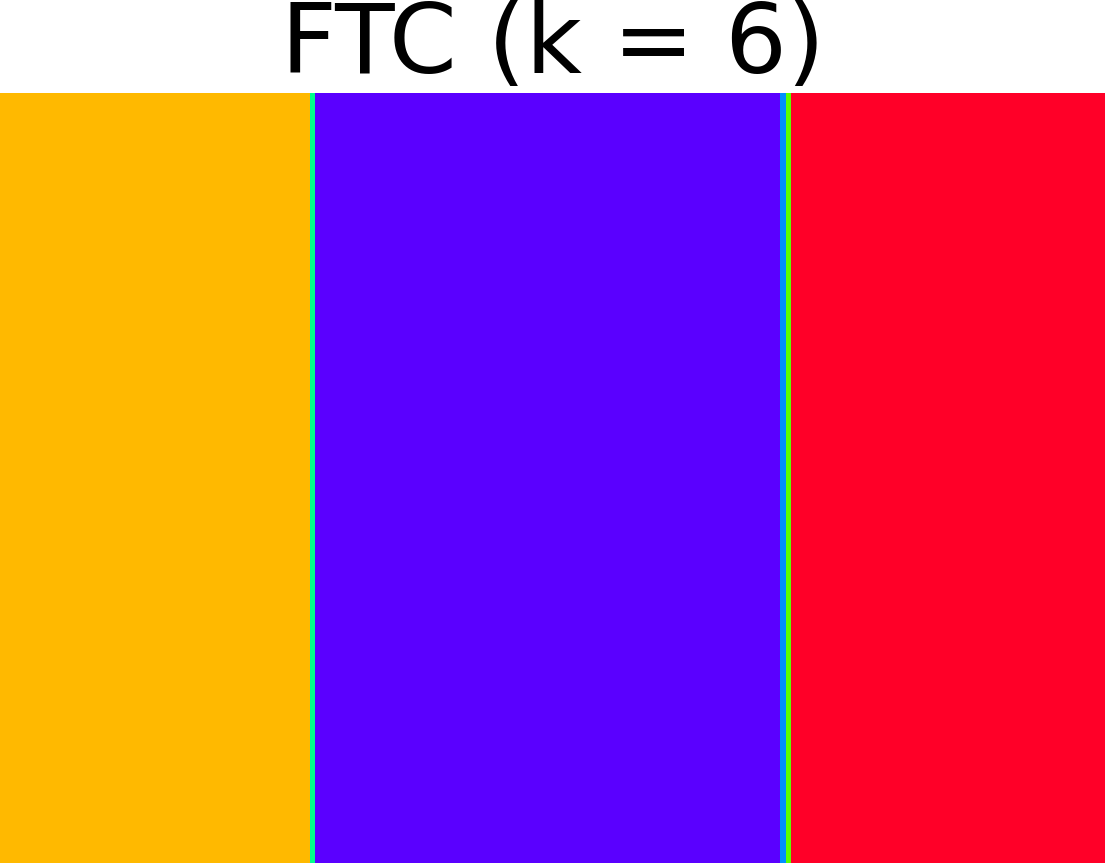}
		\includegraphics[width=0.13\linewidth]{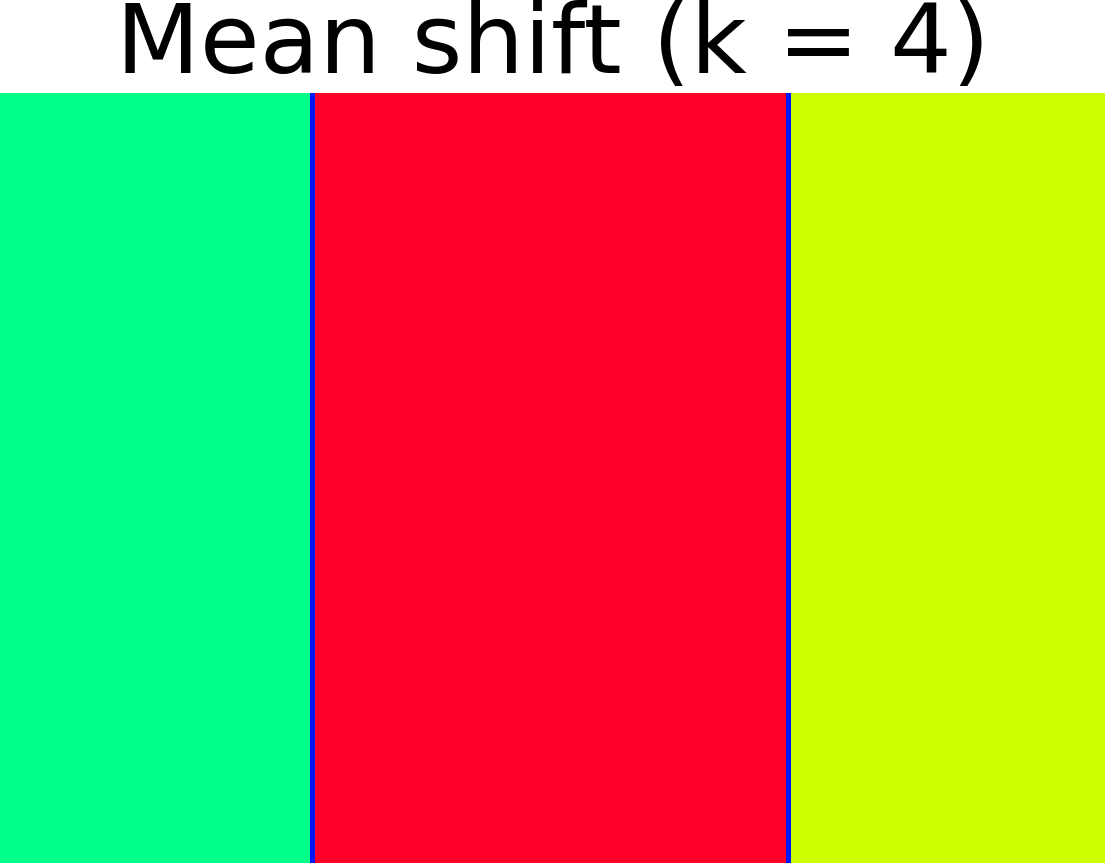}
		\includegraphics[width=0.13\linewidth]{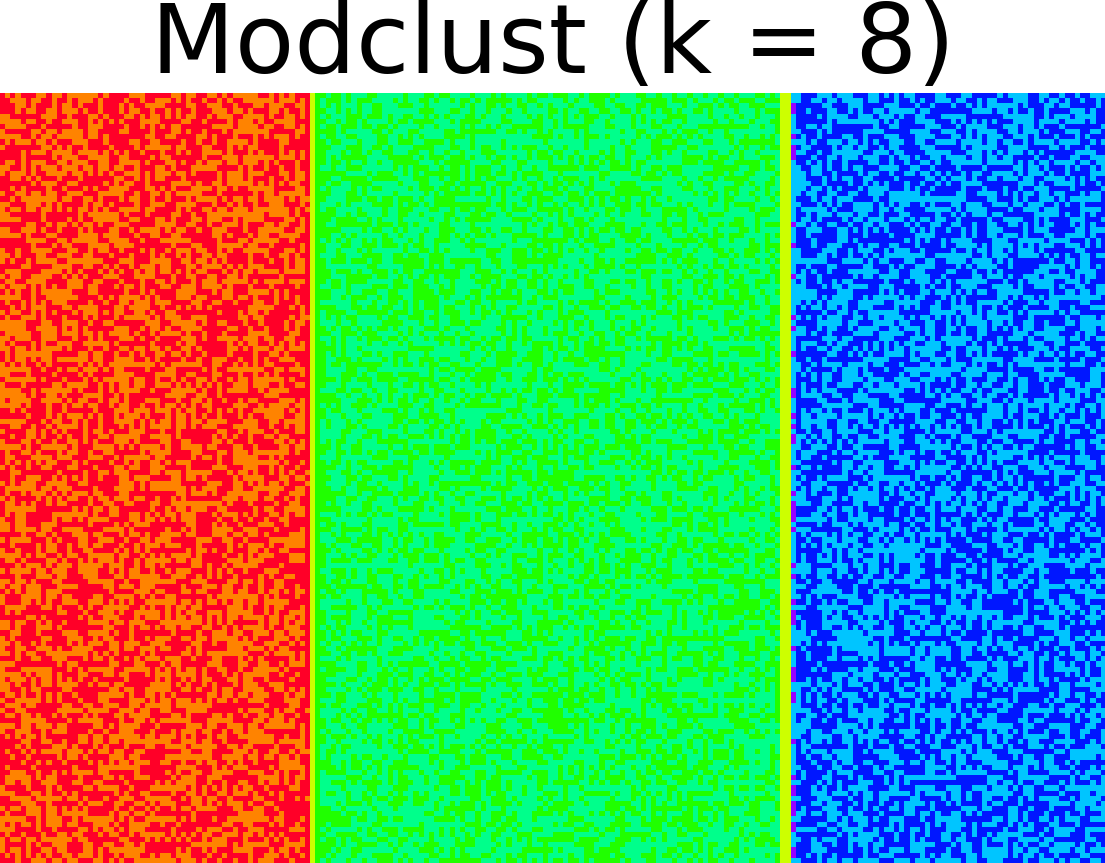}
		\includegraphics[width=0.13\linewidth]{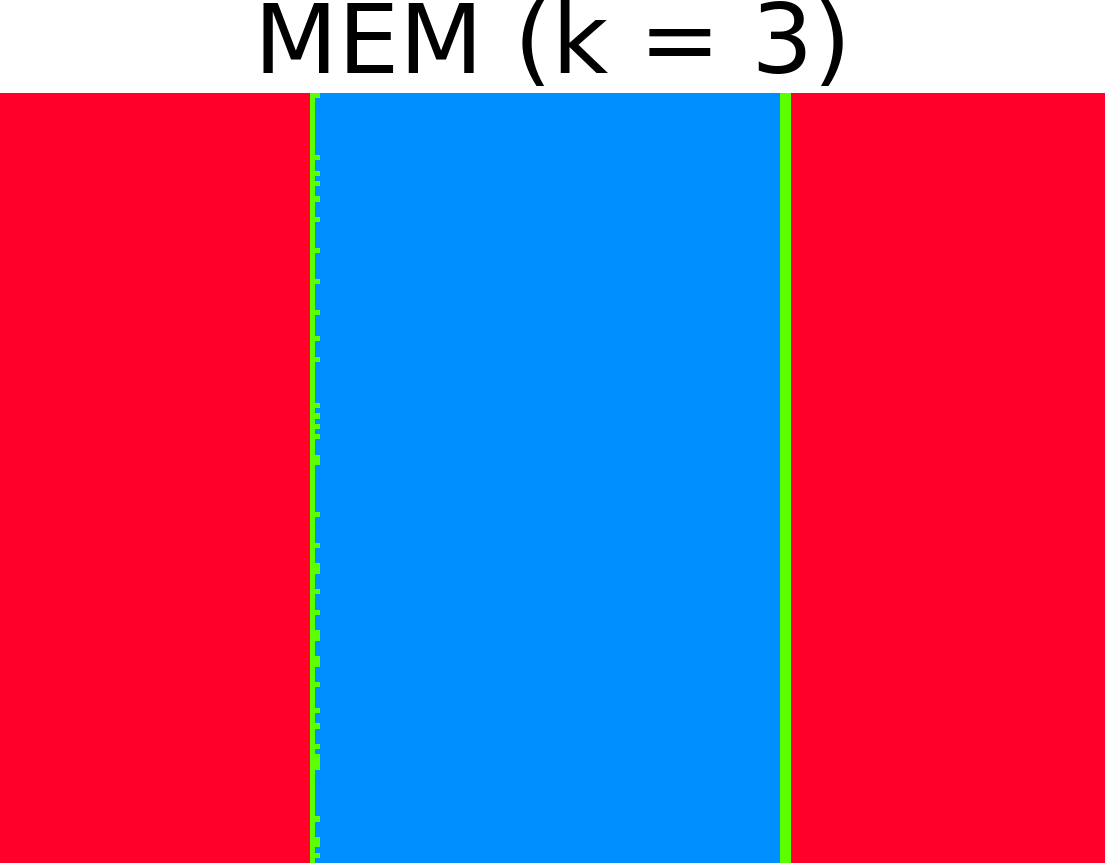}
		\caption*{}
		\label{figure:fig12b}
	\end{subfigure}
\end{figure}
\vspace{-1.2cm}
\begin{figure}[H]\ContinuedFloat
	\centering
	\begin{subfigure}[b]{\linewidth}
		\centering
		\includegraphics[width=0.13\linewidth]{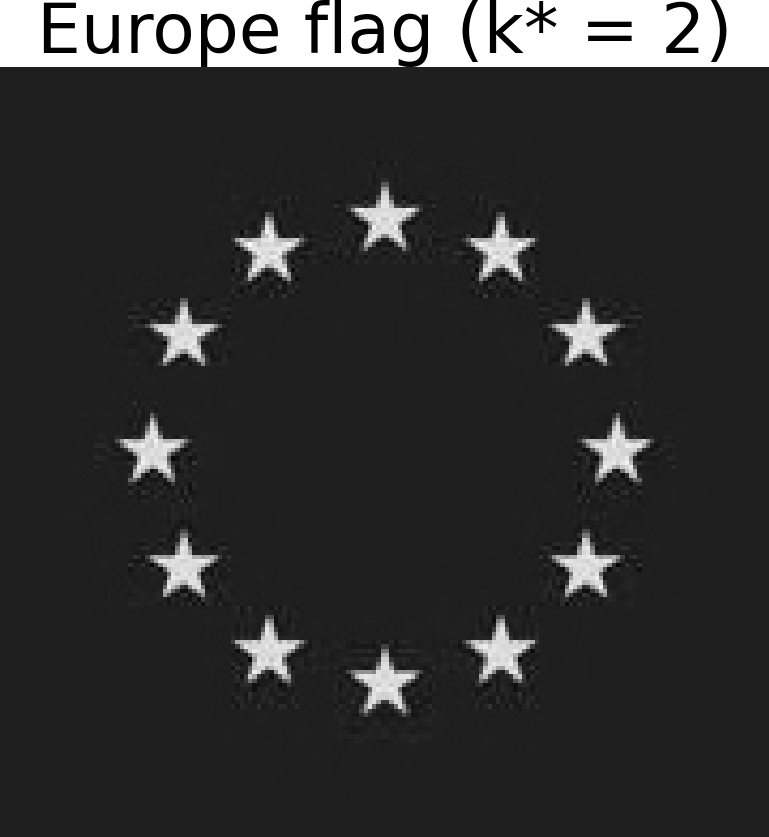}
		\includegraphics[width=0.13\linewidth]{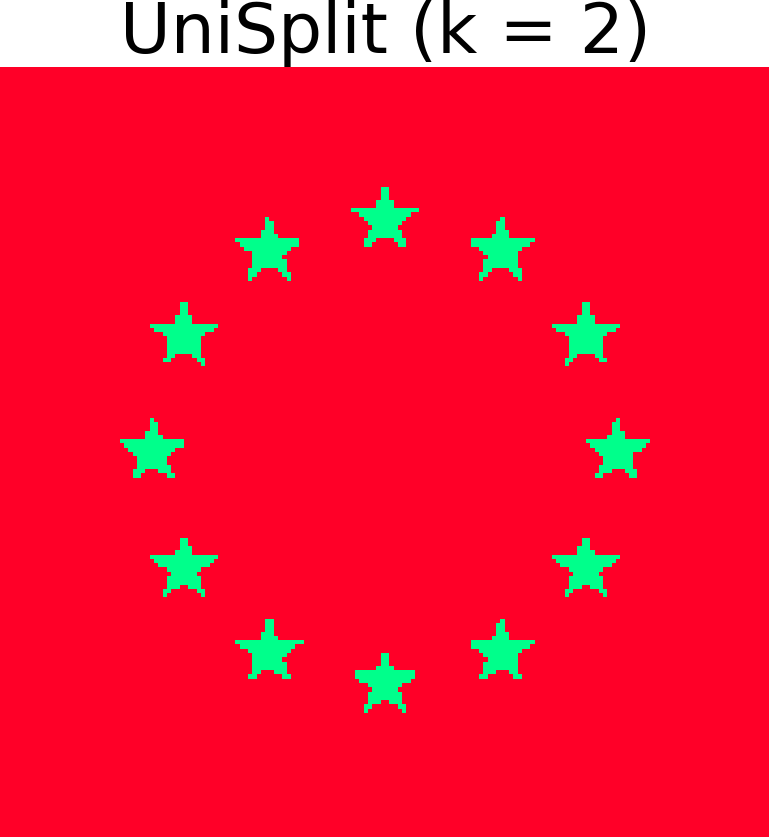}
		\includegraphics[width=0.13\linewidth]{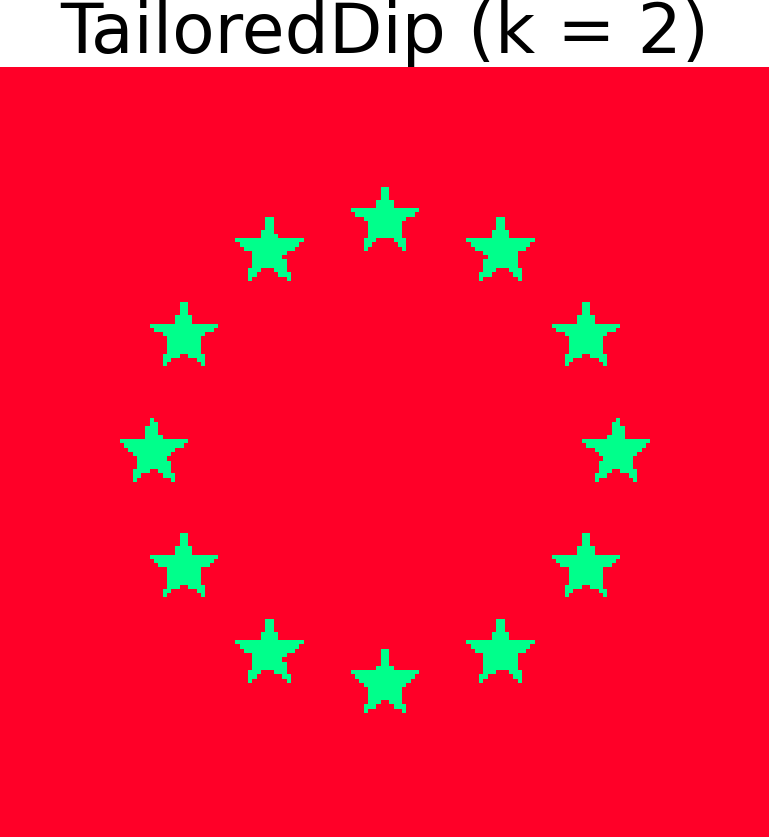}
		\includegraphics[width=0.13\linewidth]{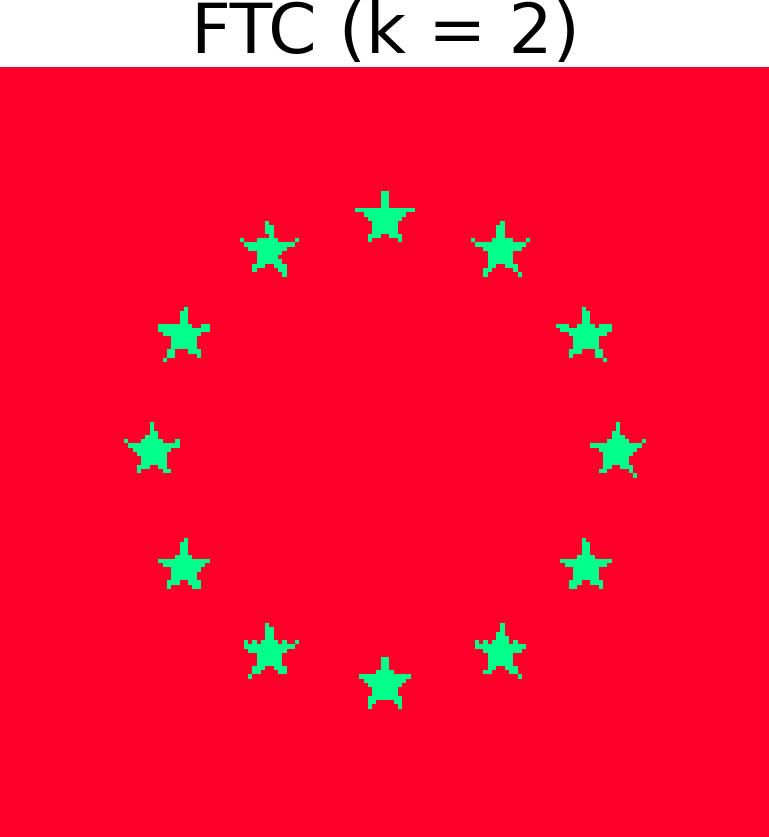}
		\includegraphics[width=0.13\linewidth]{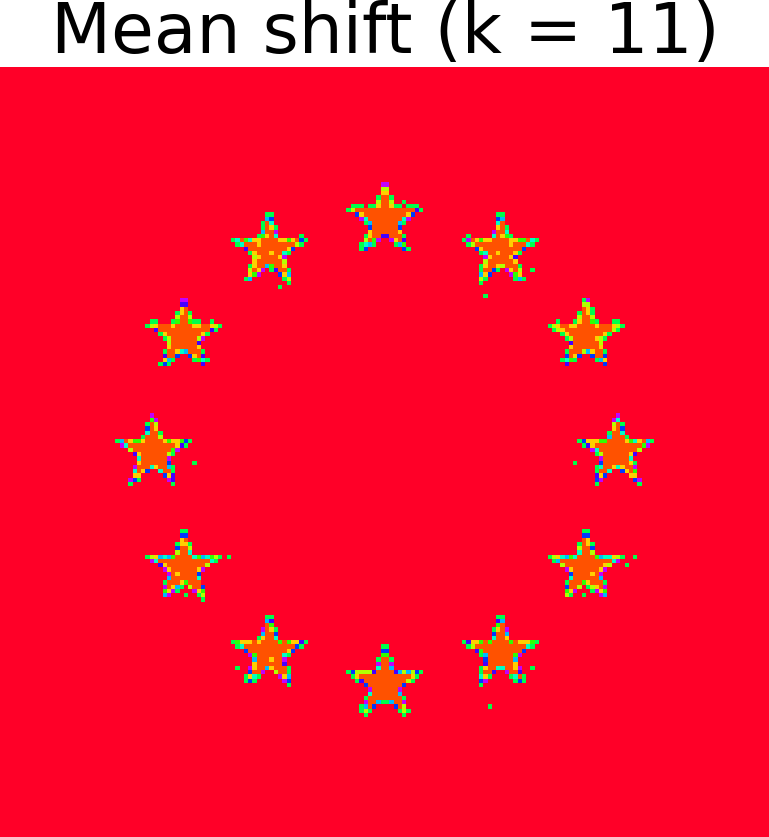}
		\includegraphics[width=0.13\linewidth]{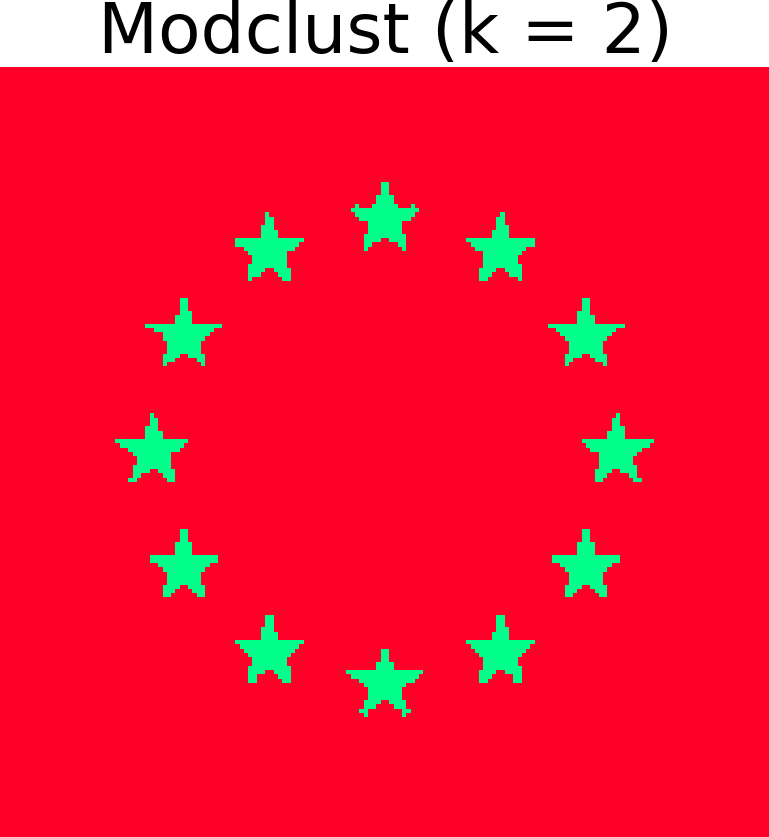}
		\includegraphics[width=0.13\linewidth]{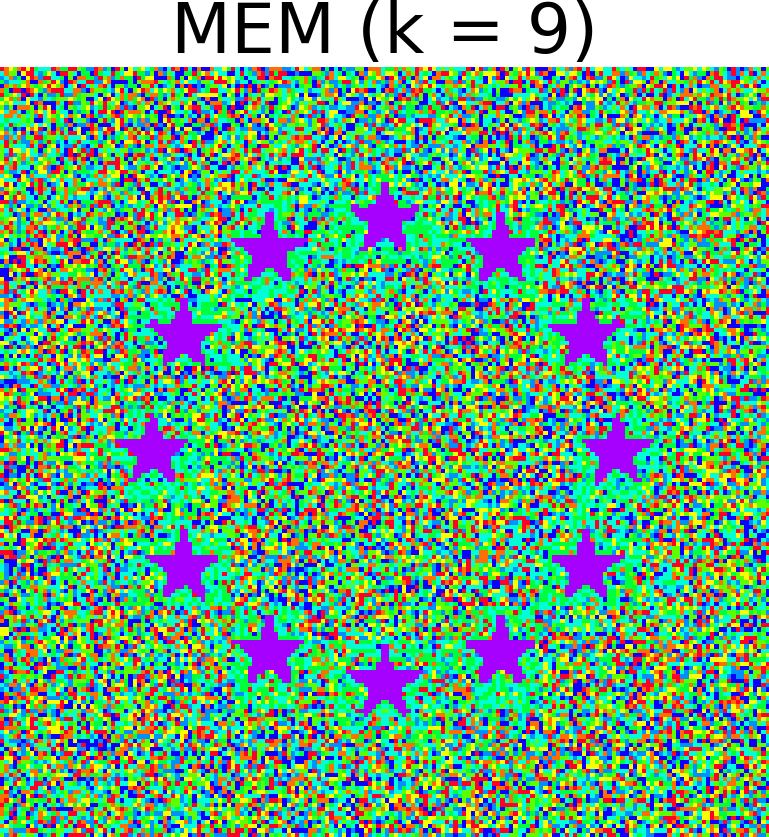}
		\caption*{}
		\label{figure:fig12c}
	\end{subfigure}
\end{figure}
\vspace{-1.2cm}
\begin{figure}[H]\ContinuedFloat
	\centering
	\begin{subfigure}[b]{\linewidth}
		\centering
		\includegraphics[width=0.13\linewidth]{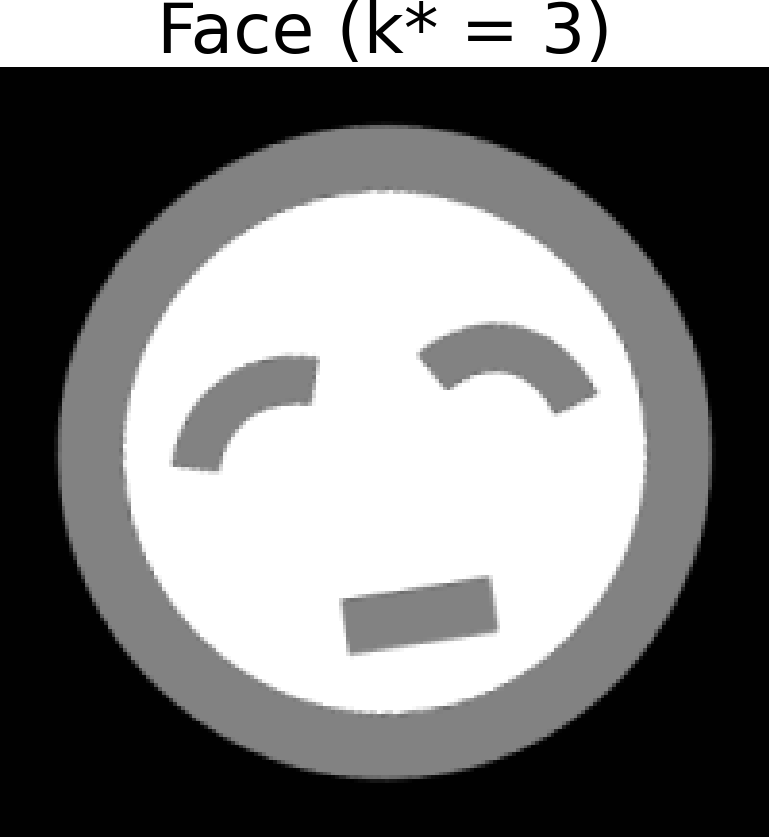}
		\includegraphics[width=0.13\linewidth]{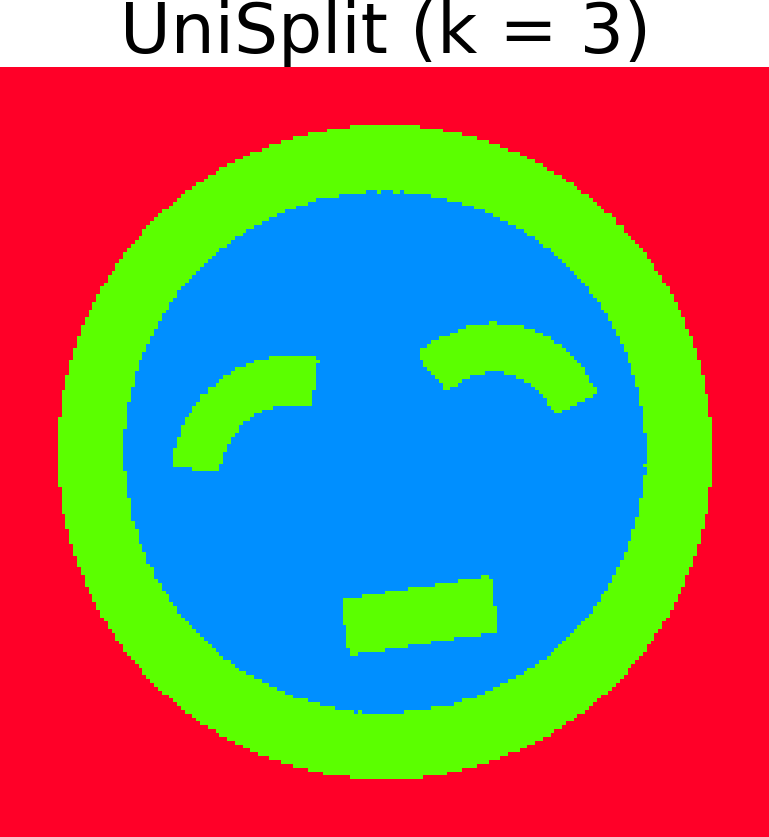}
		\includegraphics[width=0.13\linewidth]{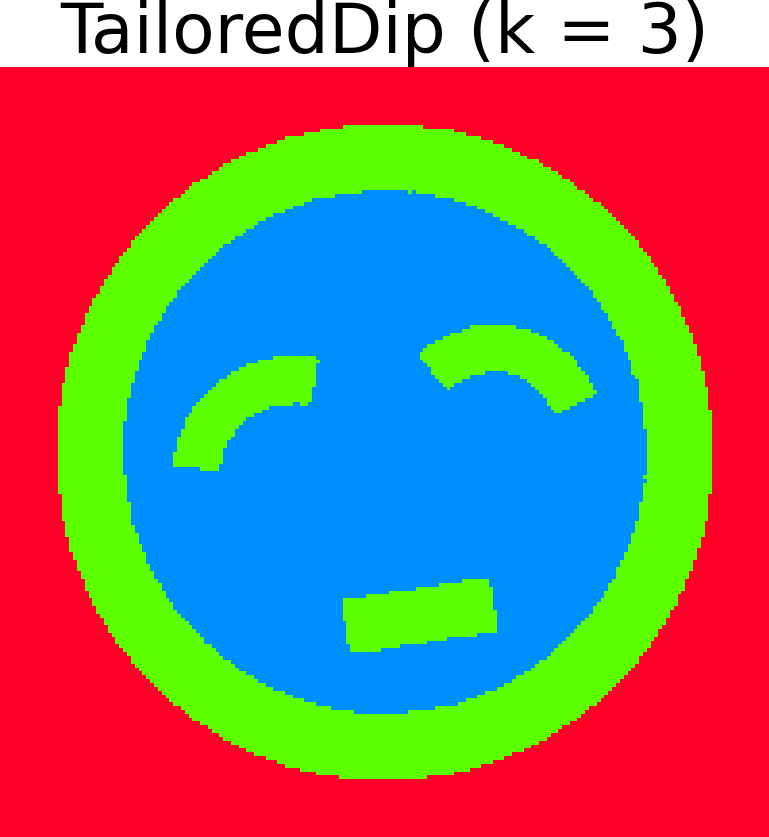}
		\includegraphics[width=0.13\linewidth]{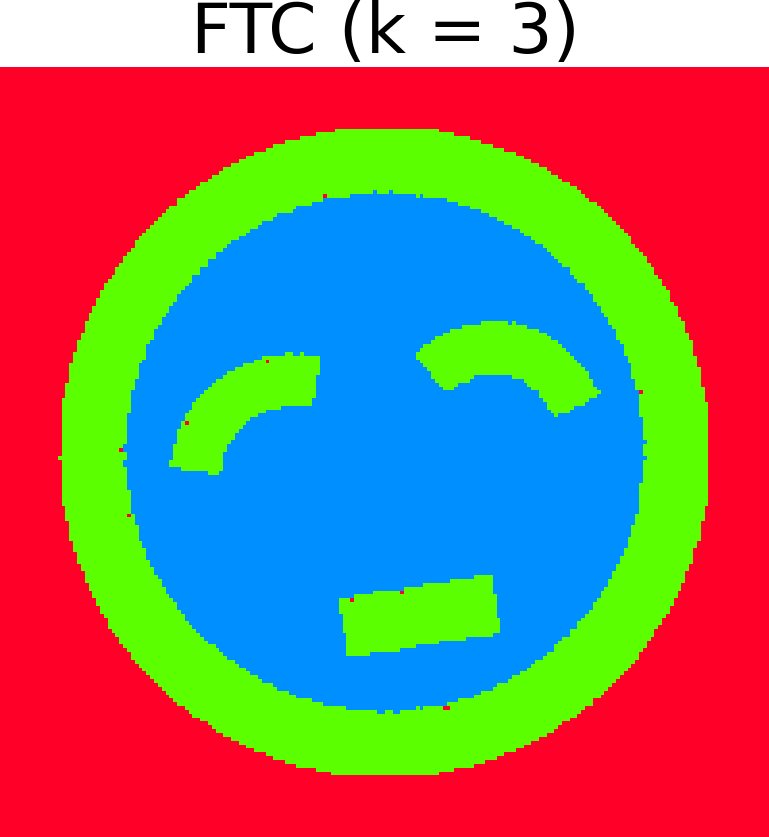}
		\includegraphics[width=0.13\linewidth]{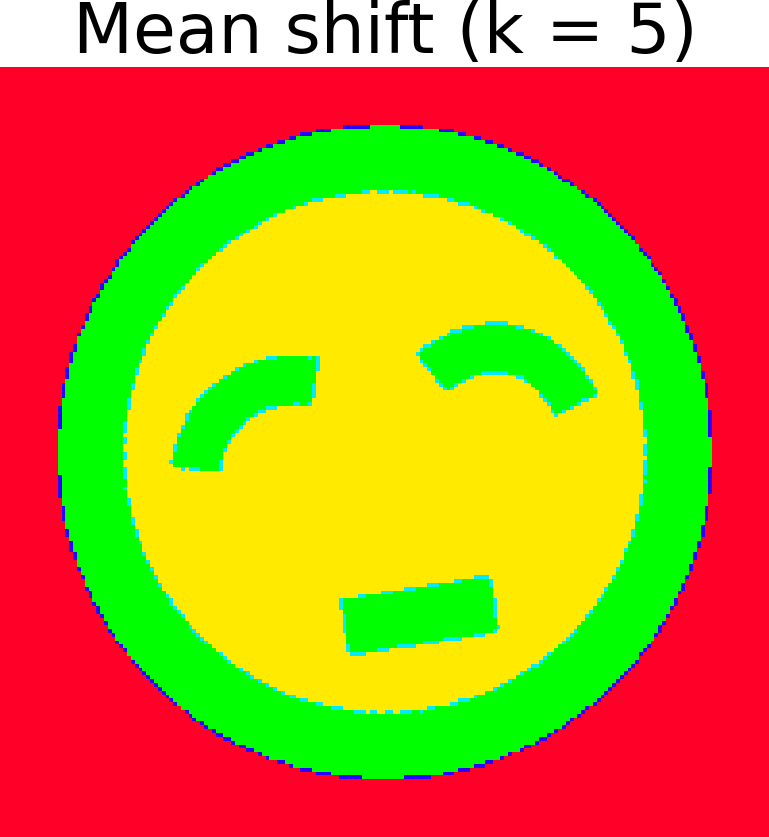}
		\includegraphics[width=0.13\linewidth]{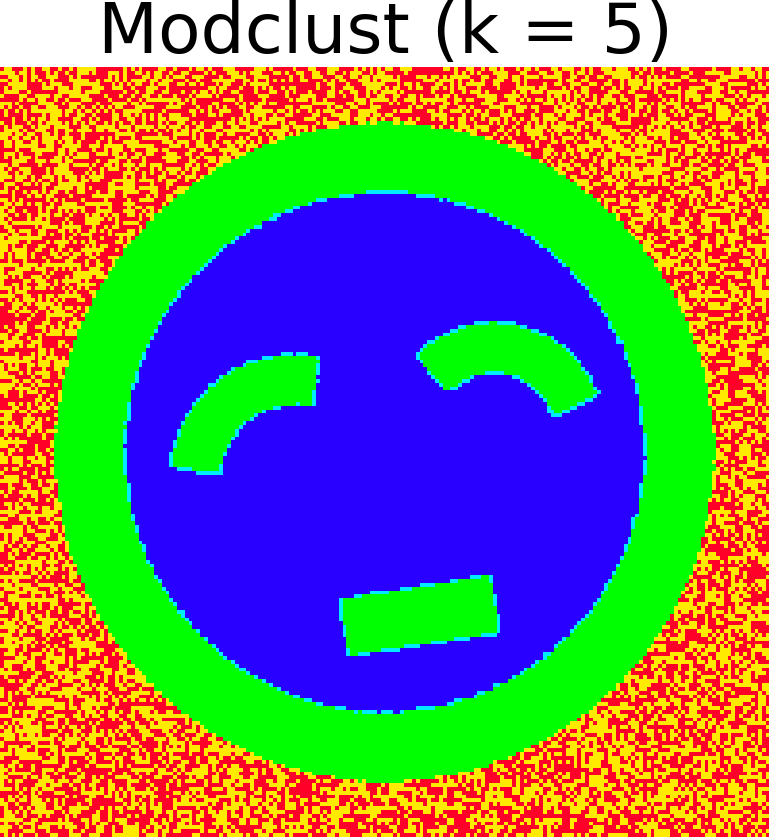}
		\includegraphics[width=0.13\linewidth]{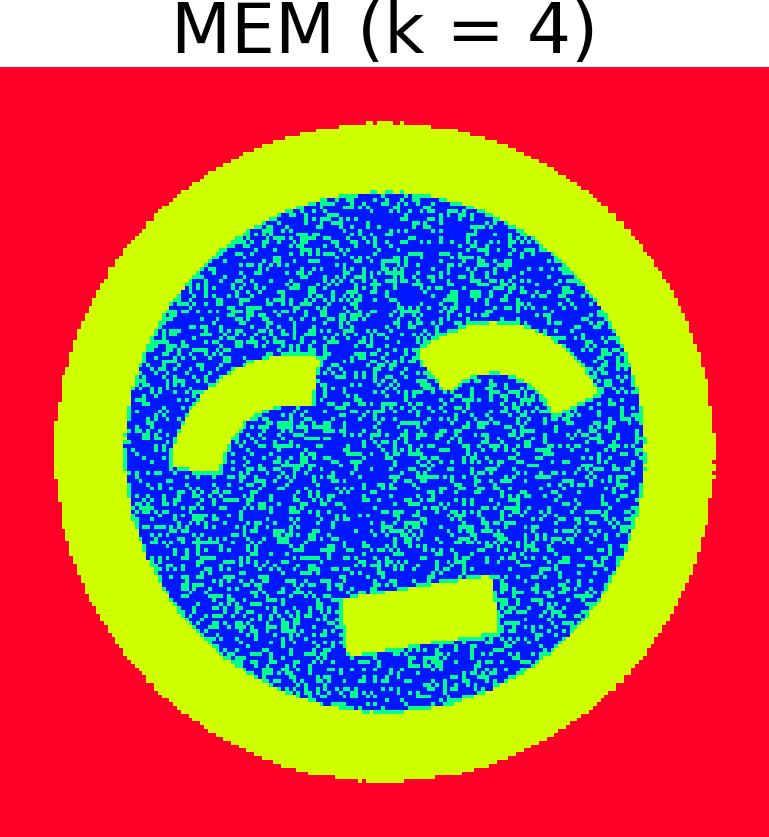}
		\caption*{}
		\label{figure:fig12e}
	\end{subfigure}
\end{figure}
\vspace{-1.2cm}
\begin{figure}[H]\ContinuedFloat
	\centering
	\begin{subfigure}[b]{\linewidth}
		\centering
		\includegraphics[width=0.13\linewidth]{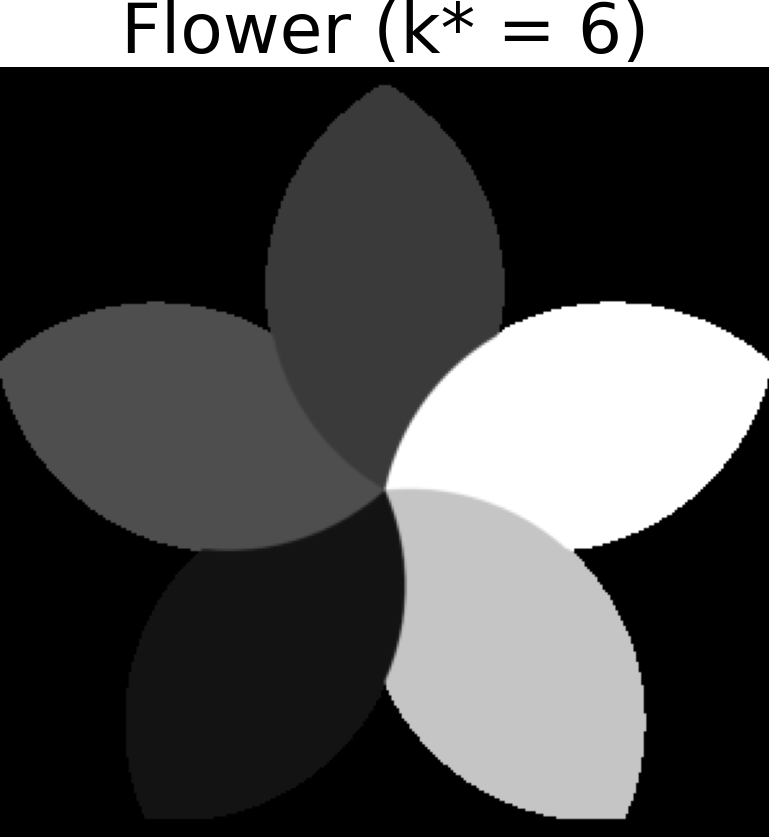}
		\includegraphics[width=0.13\linewidth]{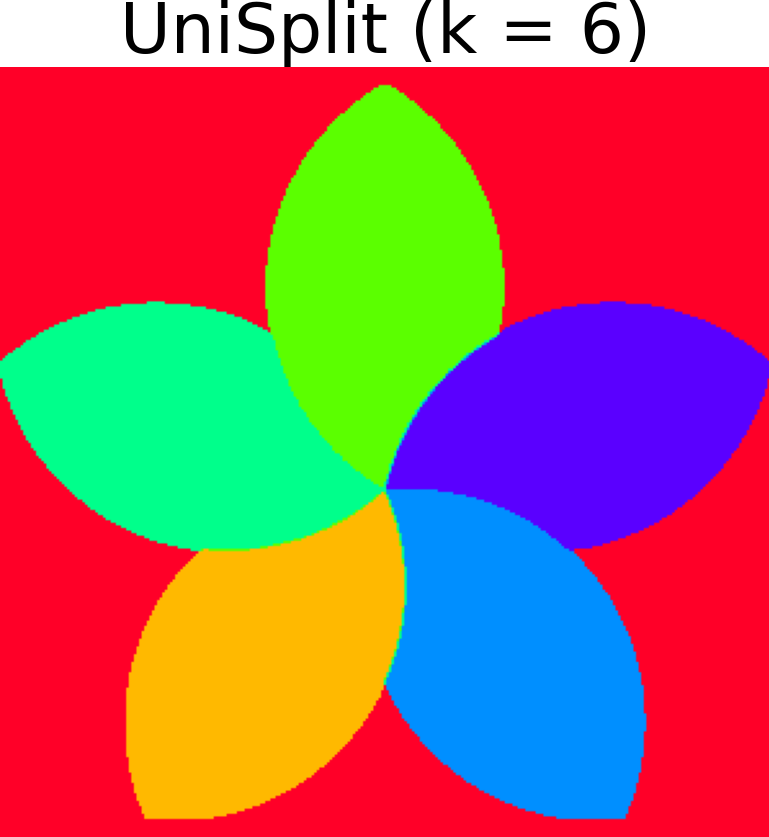}
		\includegraphics[width=0.13\linewidth]{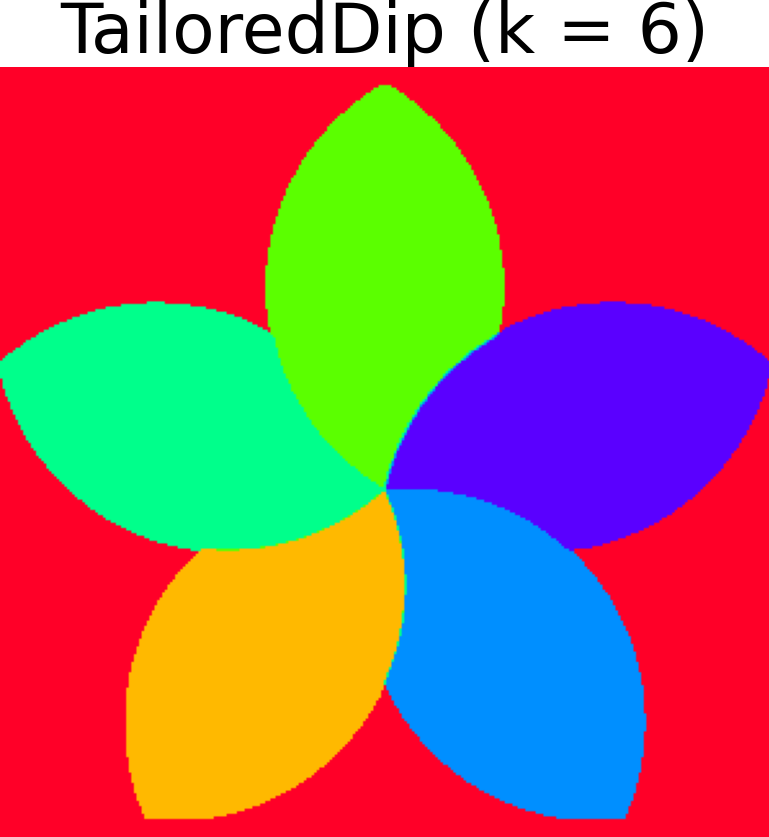}
		\includegraphics[width=0.13\linewidth]{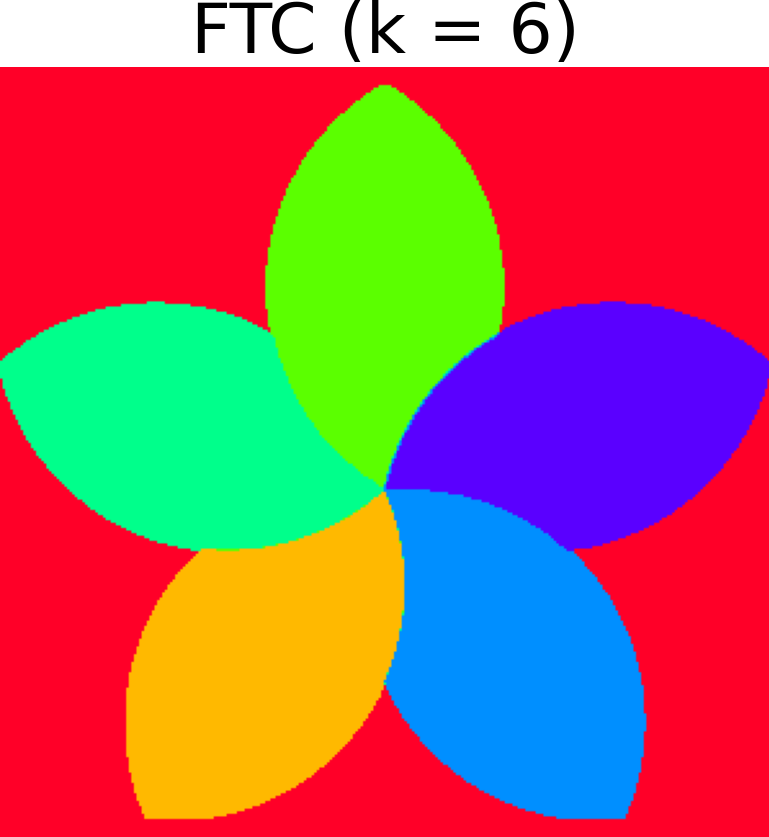}
		\includegraphics[width=0.13\linewidth]{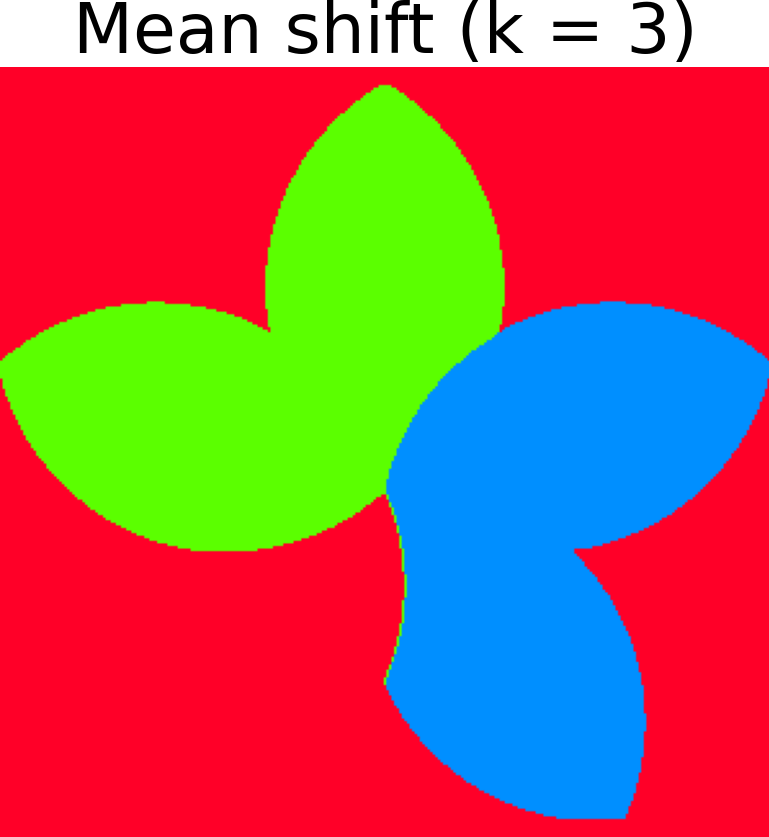}
		\includegraphics[width=0.13\linewidth]{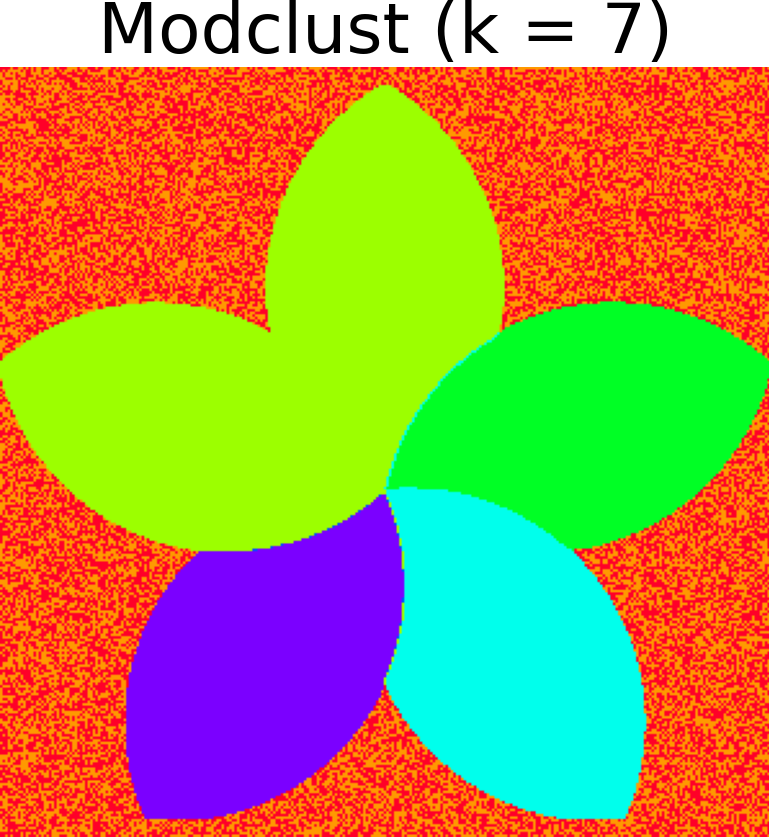}
		\includegraphics[width=0.13\linewidth]{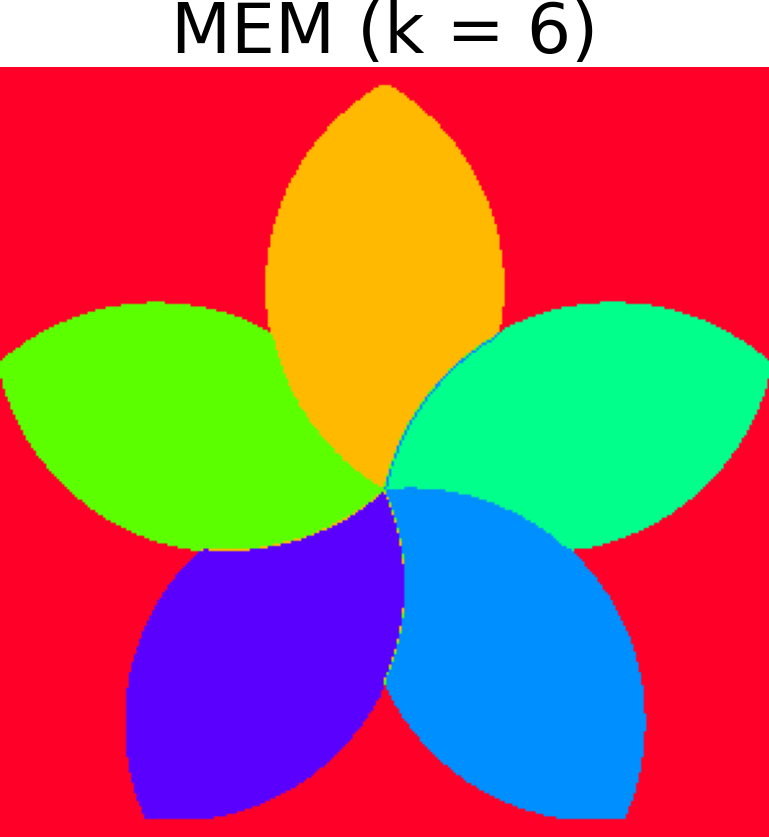}
		\caption*{}
		\label{figure:fig12f}
	\end{subfigure}
\end{figure}
\vspace{-1.2cm}
\begin{figure}[H]\ContinuedFloat
	\centering
	\begin{subfigure}[b]{\linewidth}
	\centering
	\includegraphics[width=0.13\linewidth]{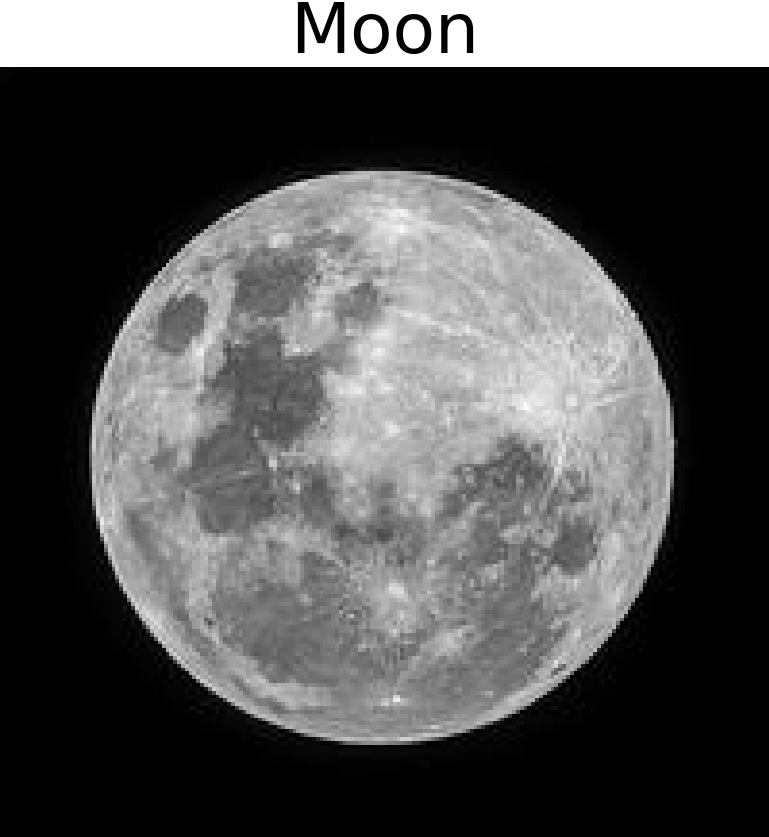}
	\includegraphics[width=0.13\linewidth]{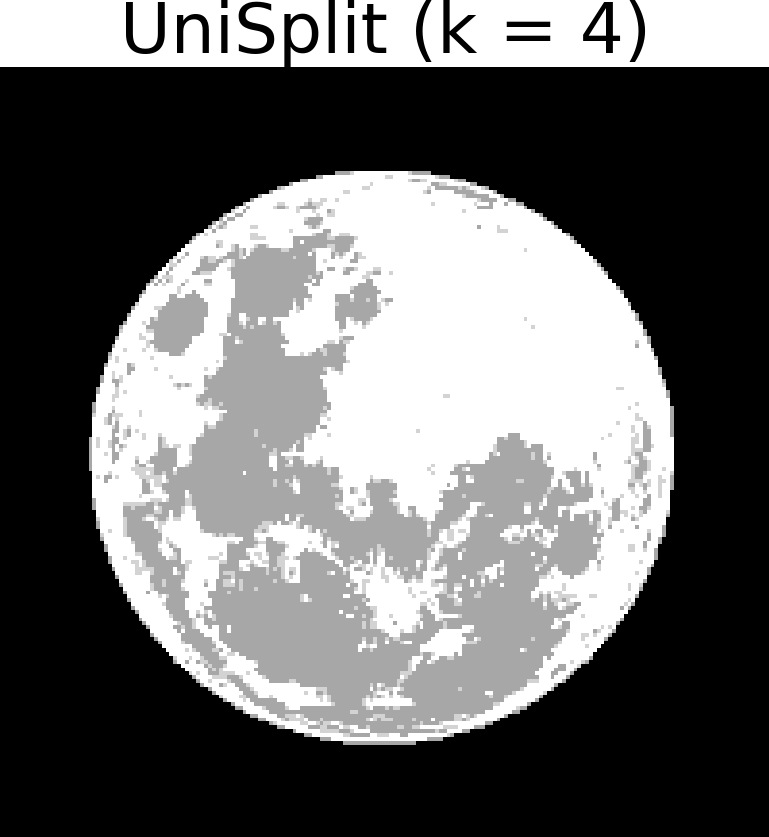}
	\includegraphics[width=0.13\linewidth]{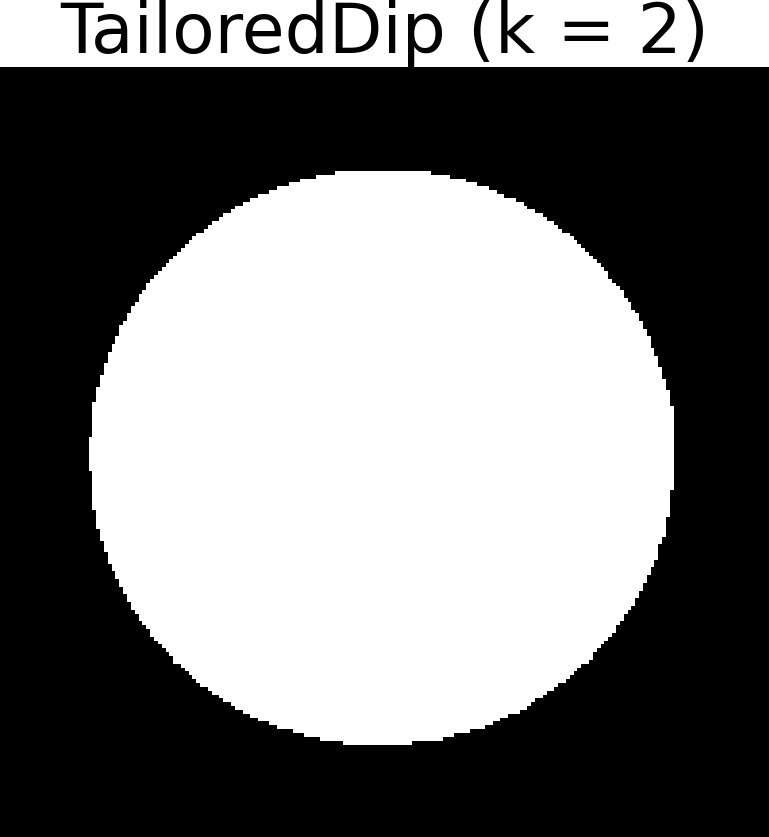}
	\includegraphics[width=0.13\linewidth]{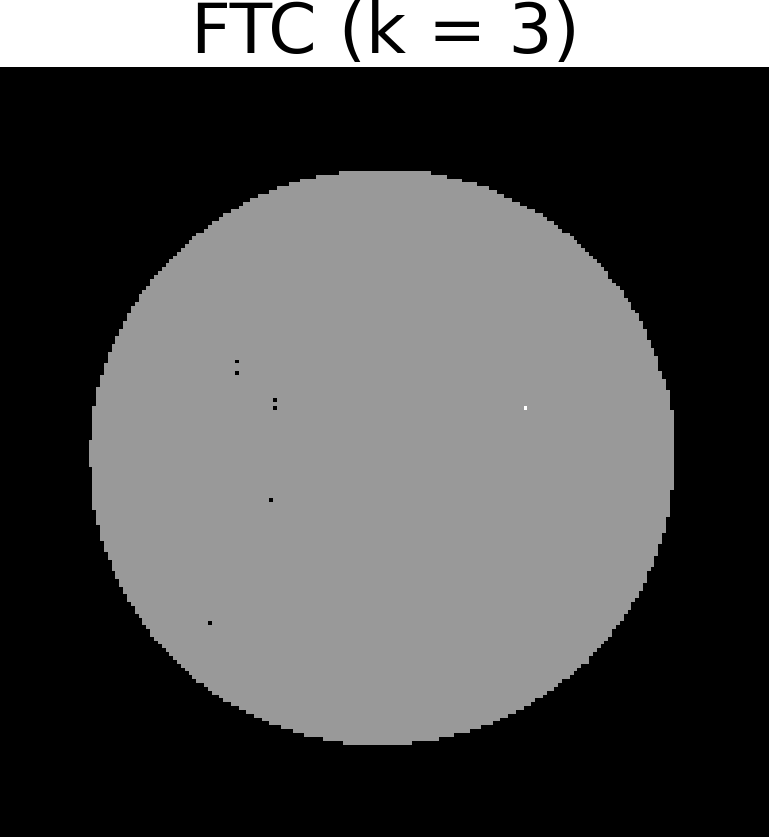}
	\includegraphics[width=0.13\linewidth]{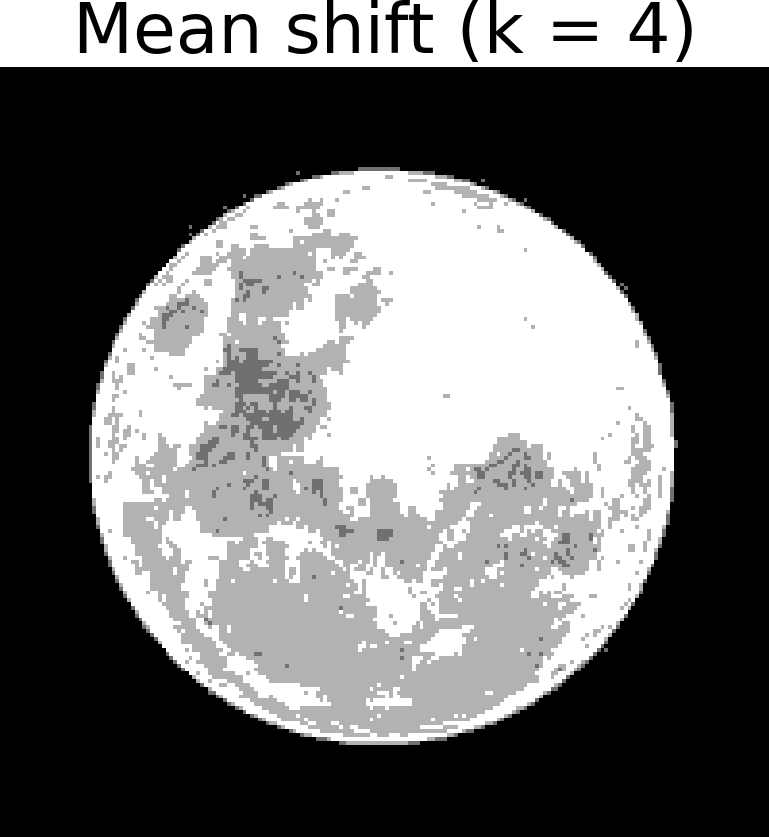}
	\includegraphics[width=0.13\linewidth]{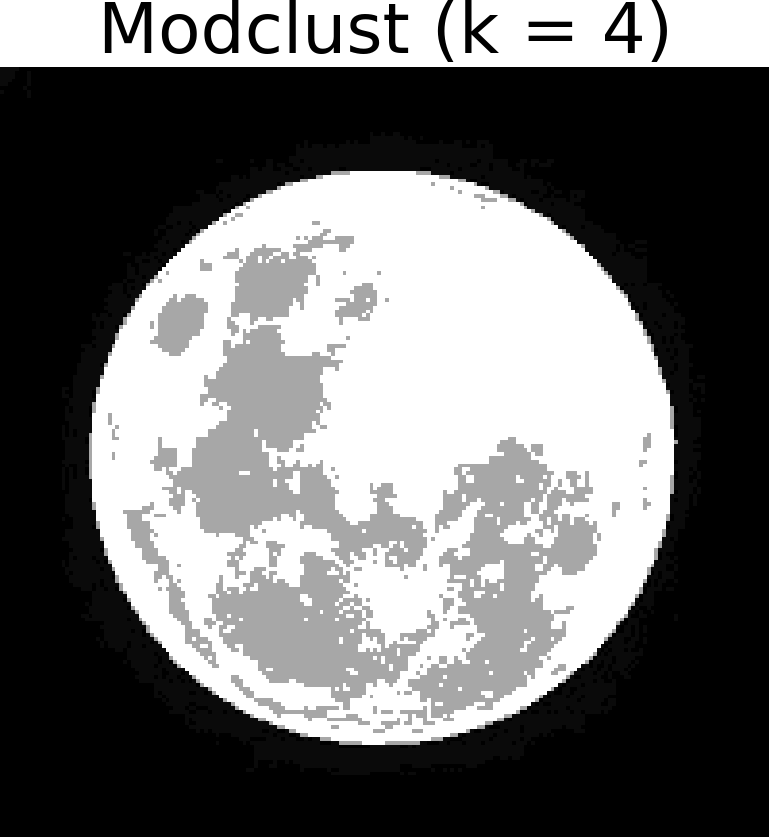}
	\includegraphics[width=0.13\linewidth]{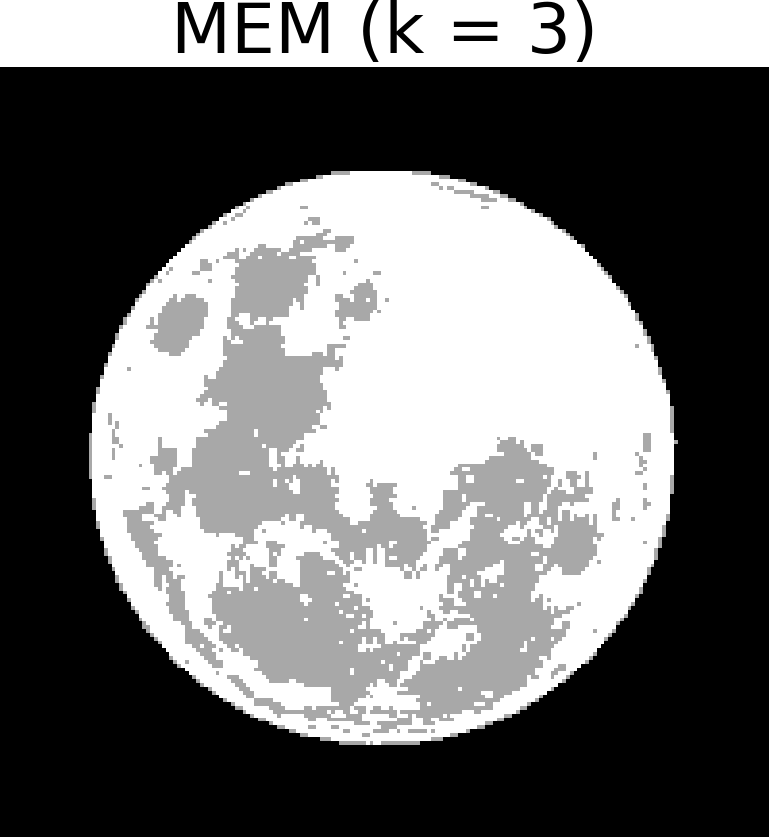}
	\caption*{}
	\label{figure:fig13c}
\end{subfigure}
\end{figure}
\vspace{-1.2cm}
\begin{figure}[H]\ContinuedFloat
	\centering
	\begin{subfigure}[b]{\linewidth}
	\centering
	\includegraphics[width=0.13\linewidth]{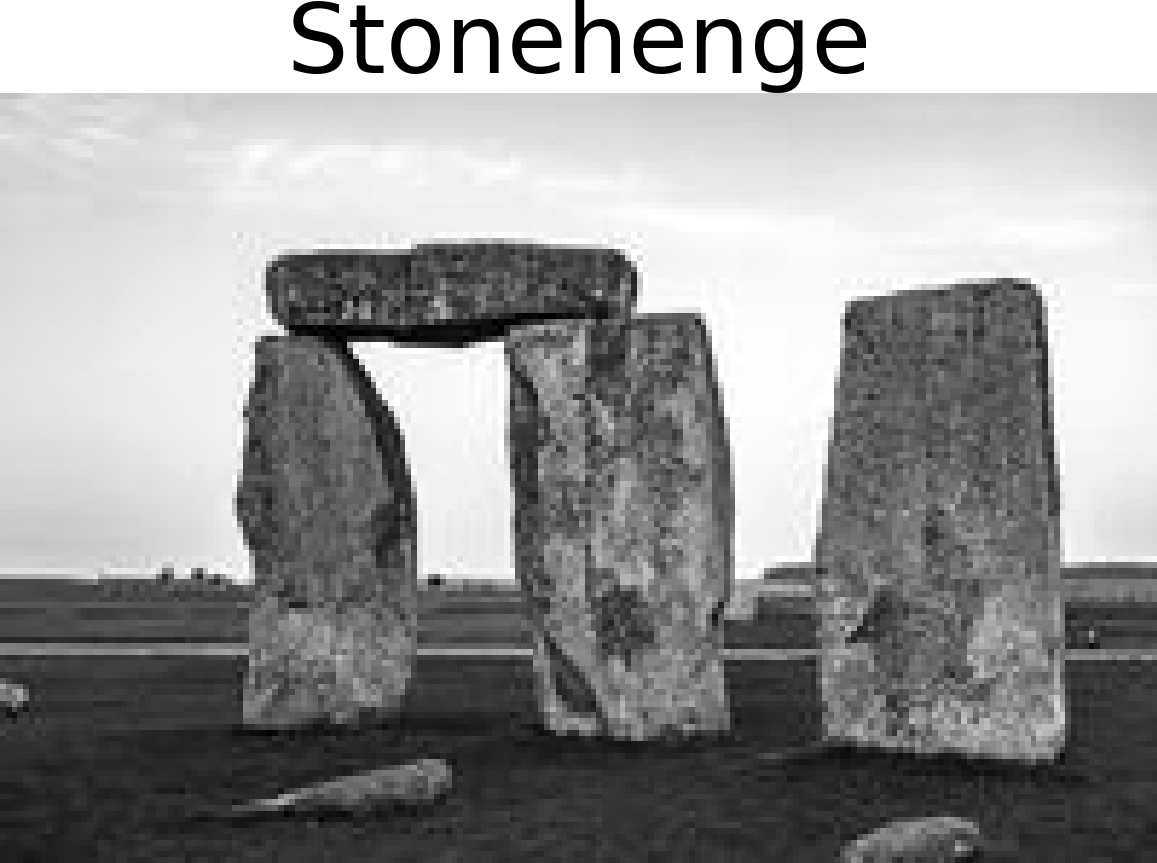}
	\includegraphics[width=0.13\linewidth]{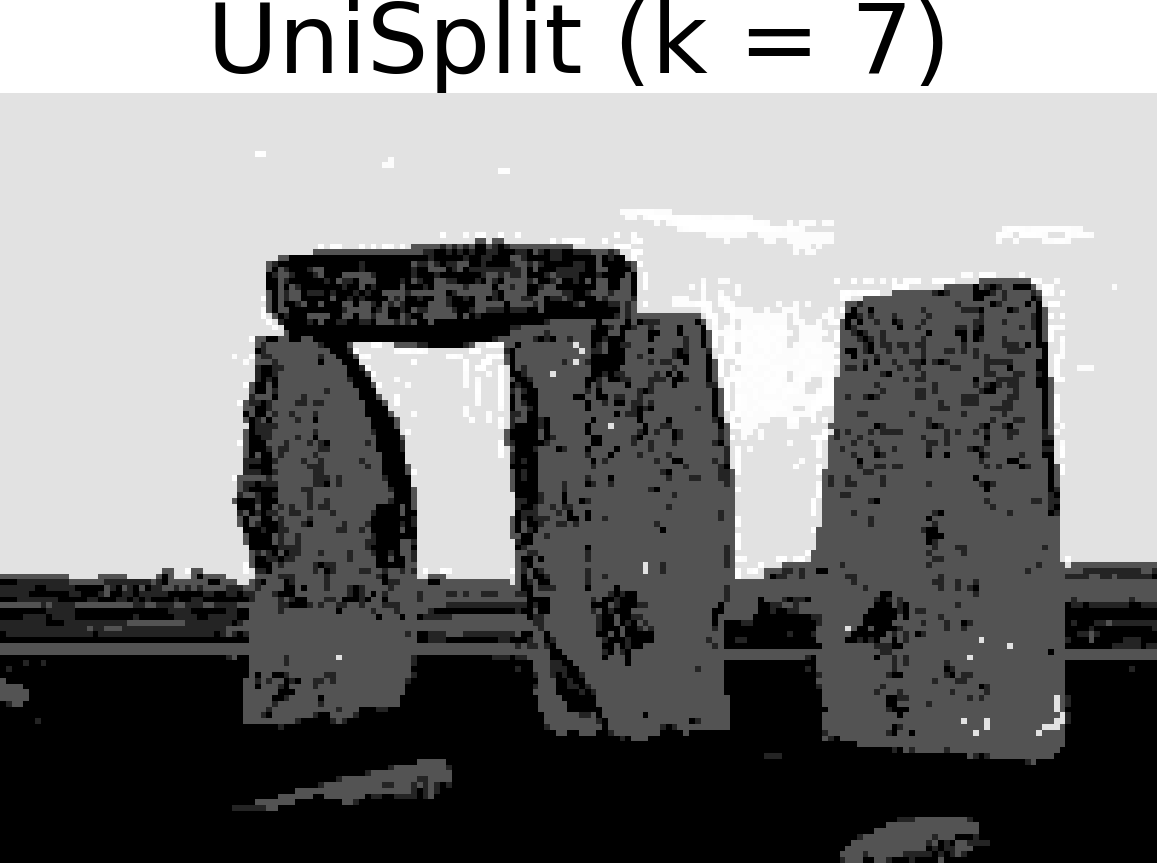}
	\includegraphics[width=0.13\linewidth]{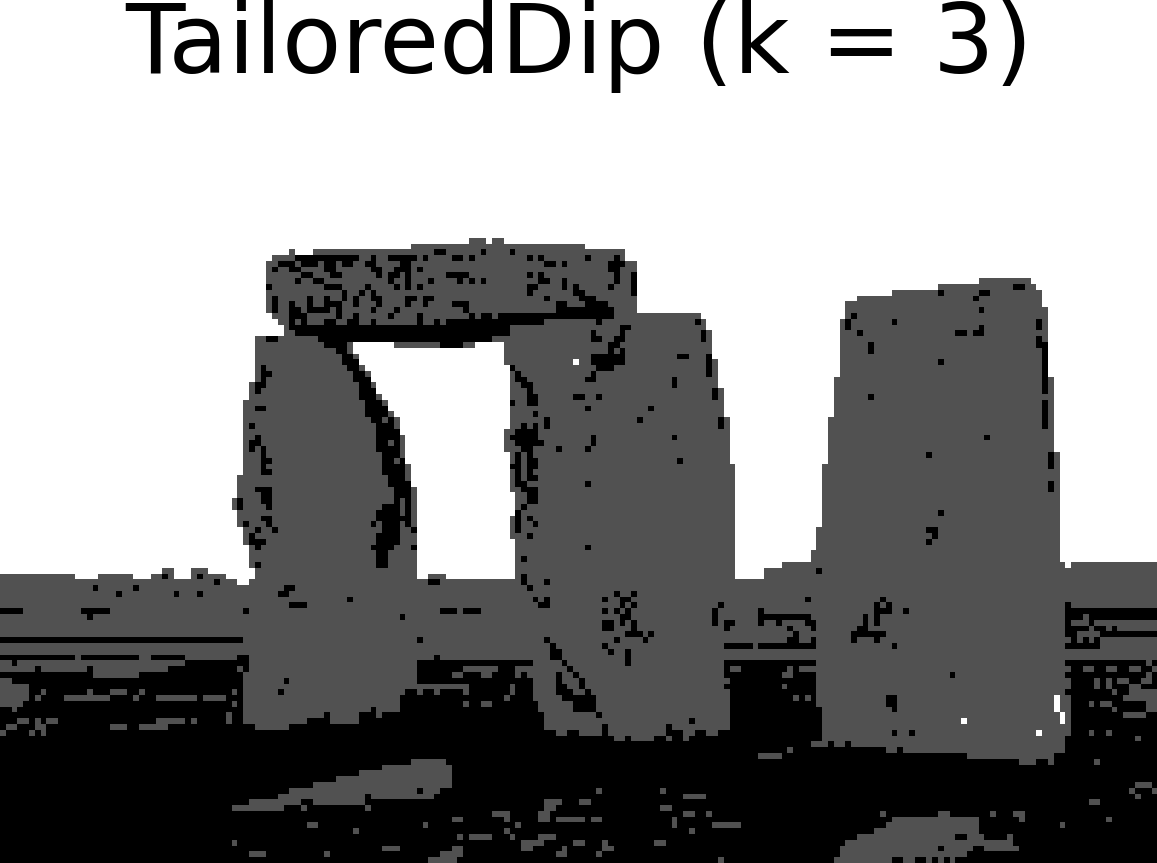}
	\includegraphics[width=0.13\linewidth]{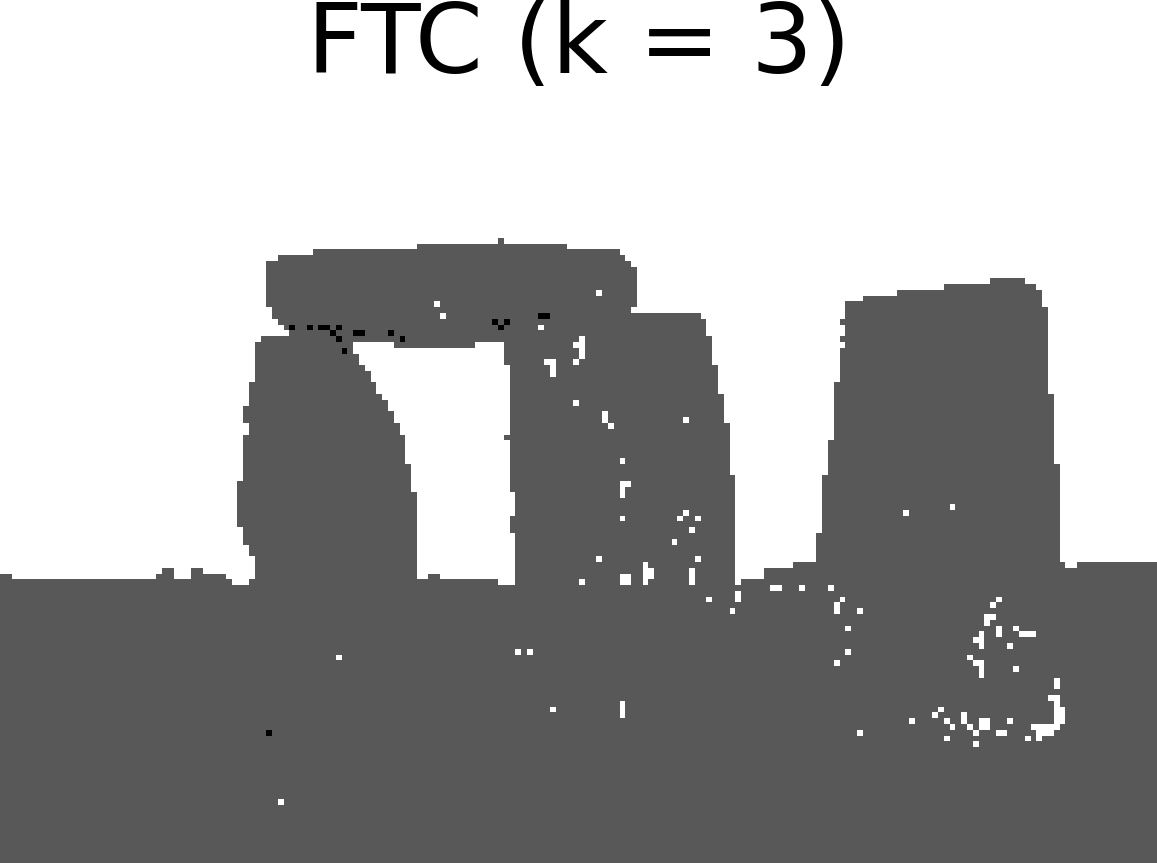}
	\includegraphics[width=0.13\linewidth]{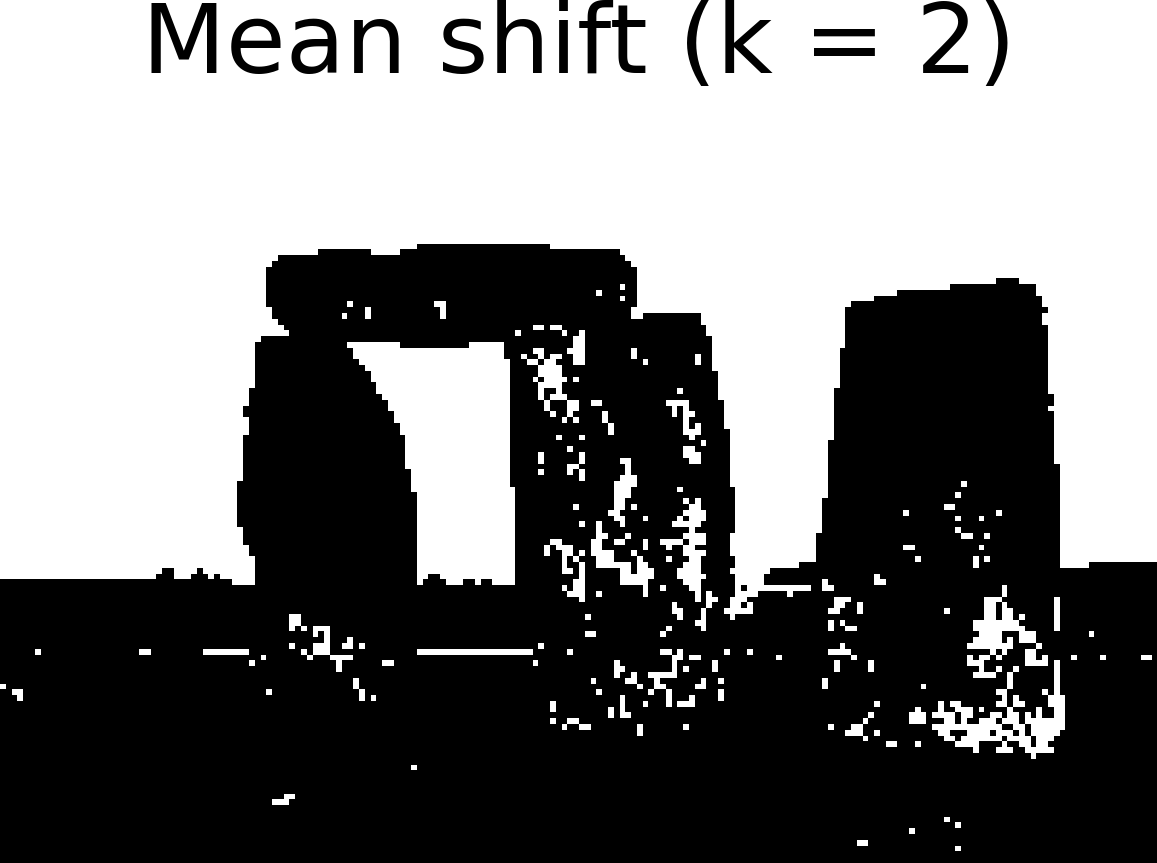}
	\includegraphics[width=0.13\linewidth]{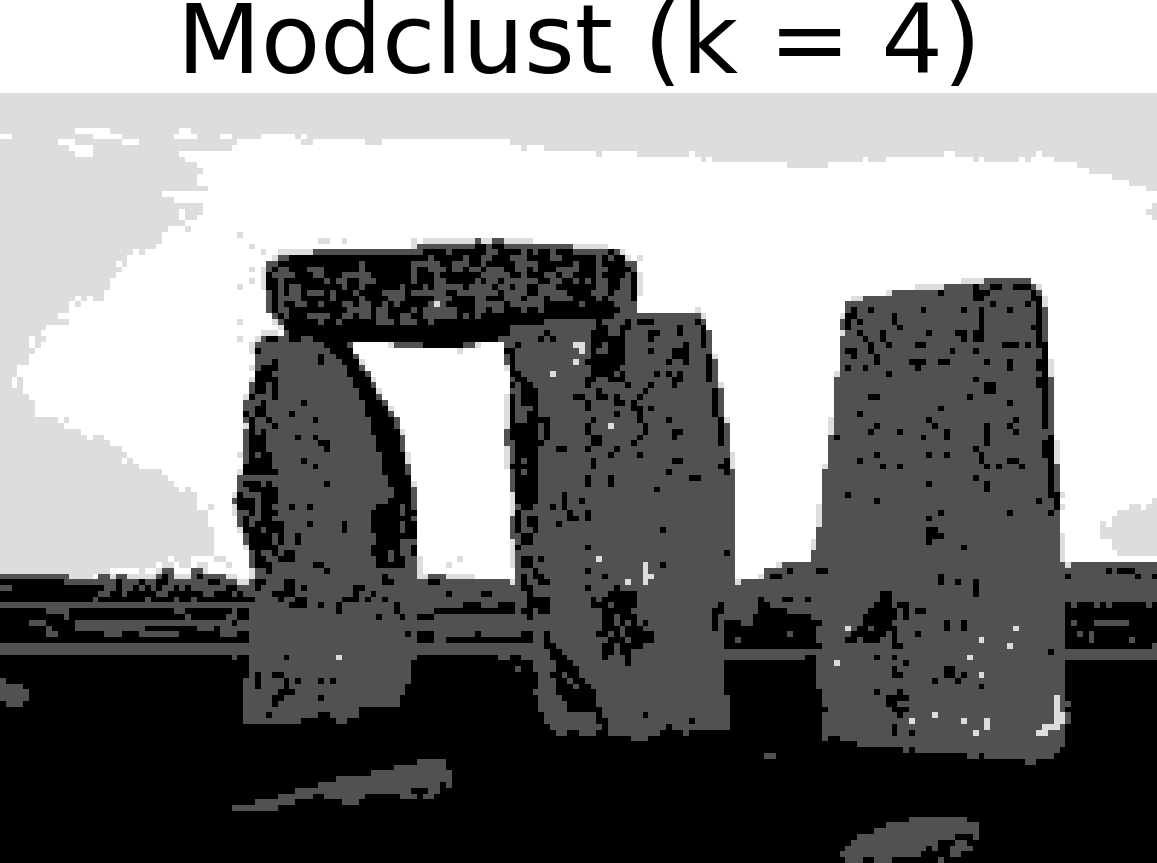}
	\includegraphics[width=0.13\linewidth]{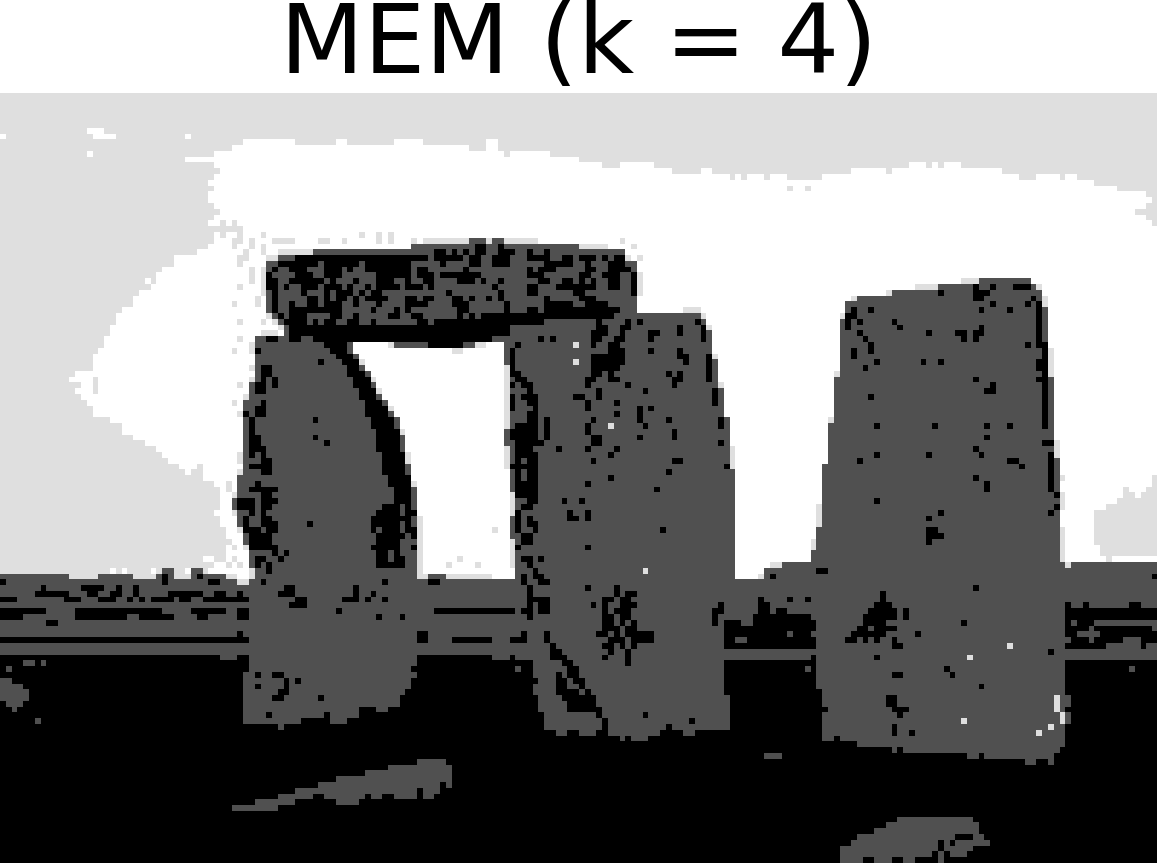}
	\caption*{}
	\label{figure:fig13d}
	\end{subfigure}
	\end{figure}
	\vspace{-1.2cm}
	\begin{figure}[H]\ContinuedFloat
	\centering
	\begin{subfigure}[b]{\linewidth}
		\centering
		\includegraphics[width=0.13\linewidth]{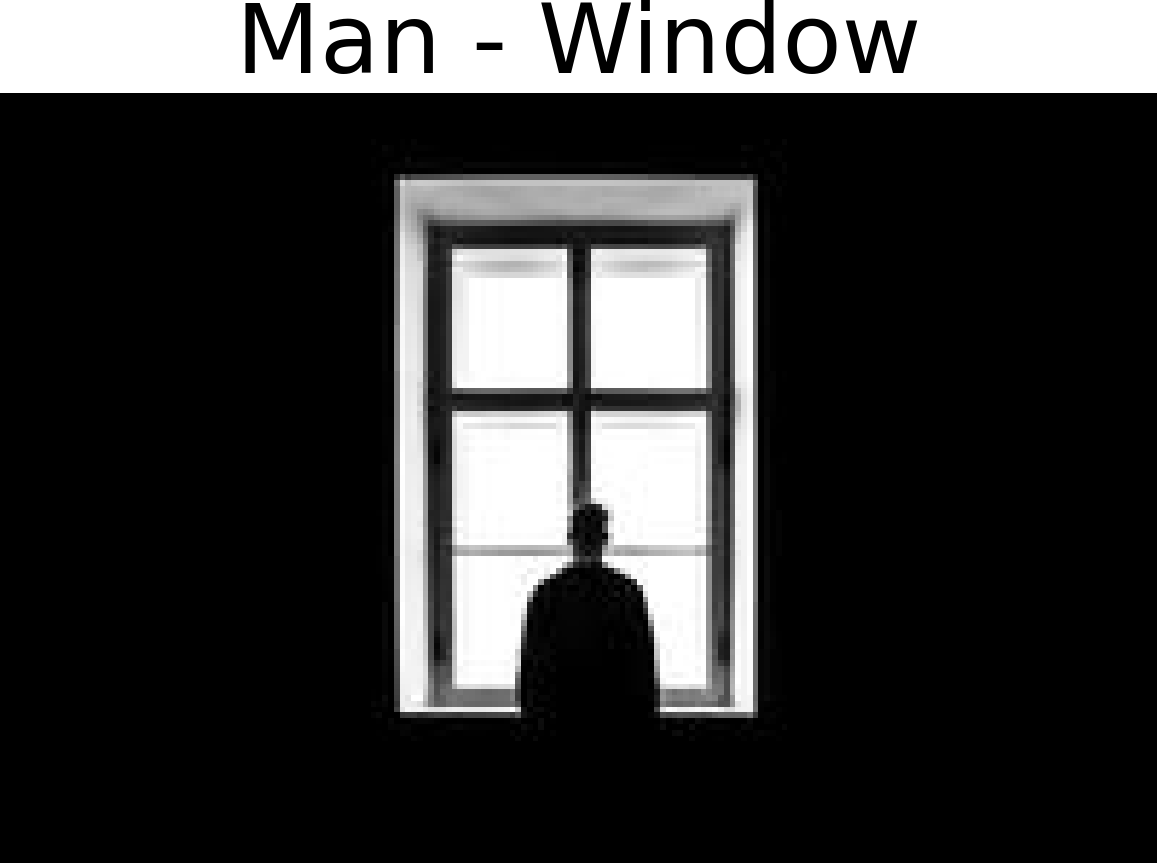}
		\includegraphics[width=0.13\linewidth]{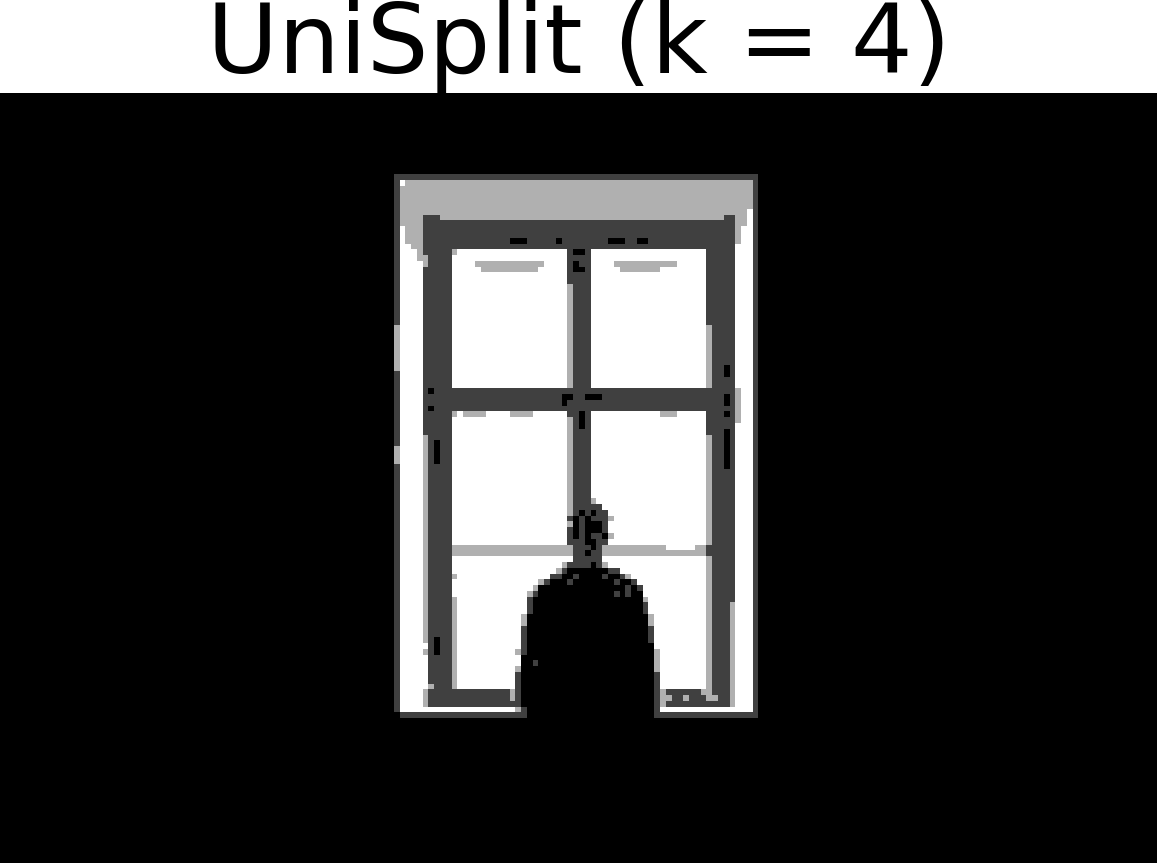}
		\includegraphics[width=0.13\linewidth]{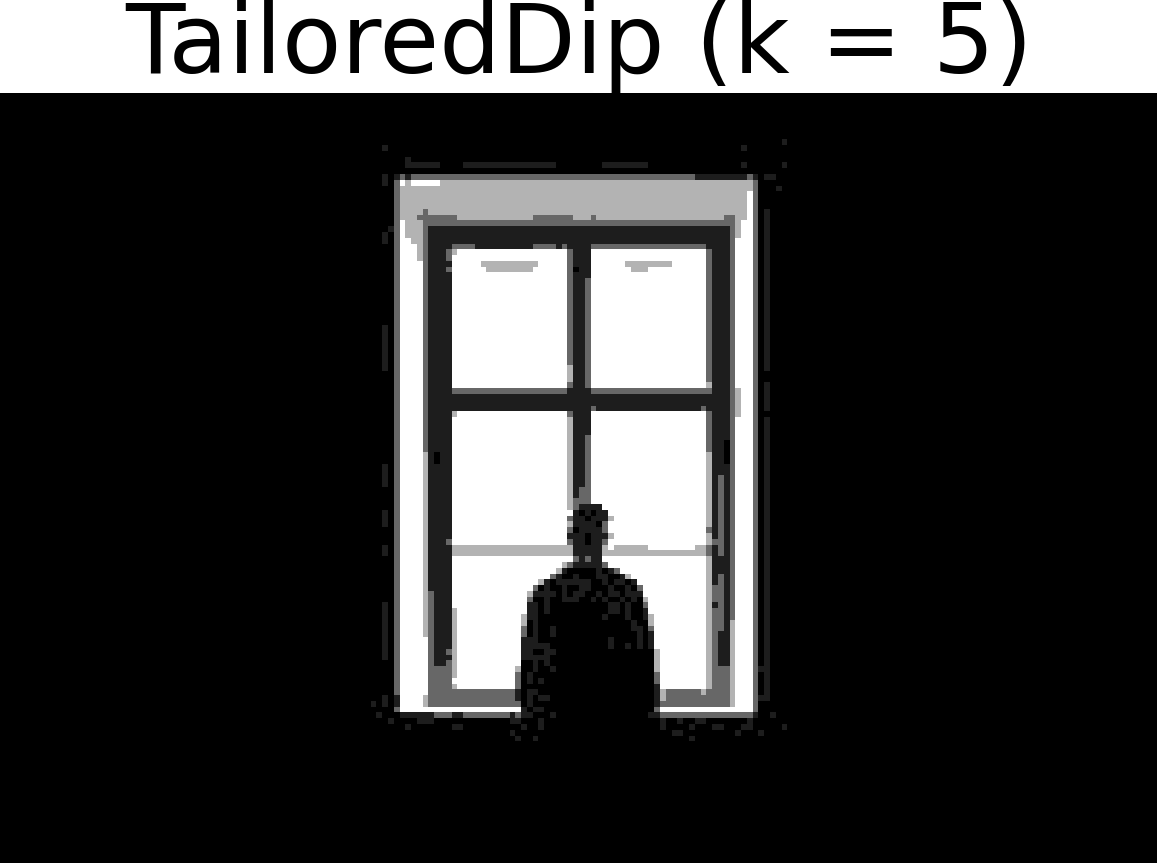}
		\includegraphics[width=0.13\linewidth]{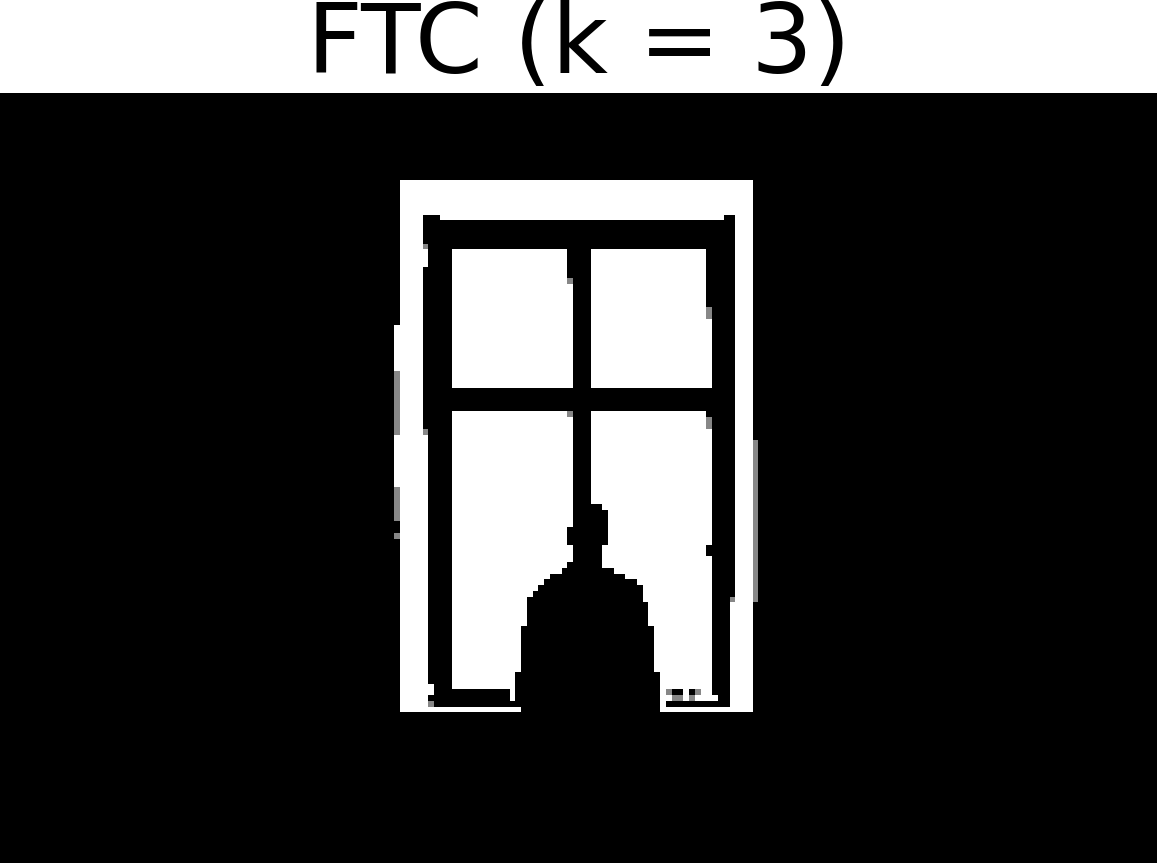}
		\includegraphics[width=0.13\linewidth]{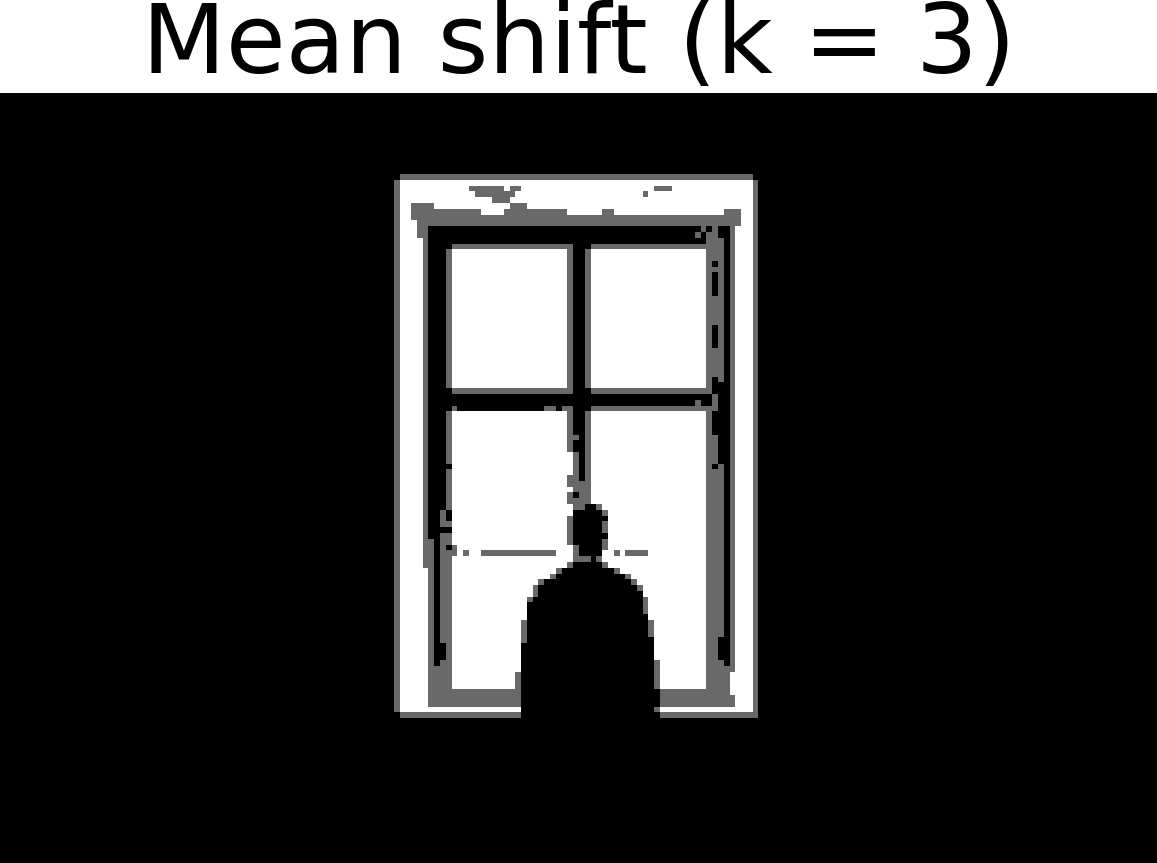}
		\includegraphics[width=0.13\linewidth]{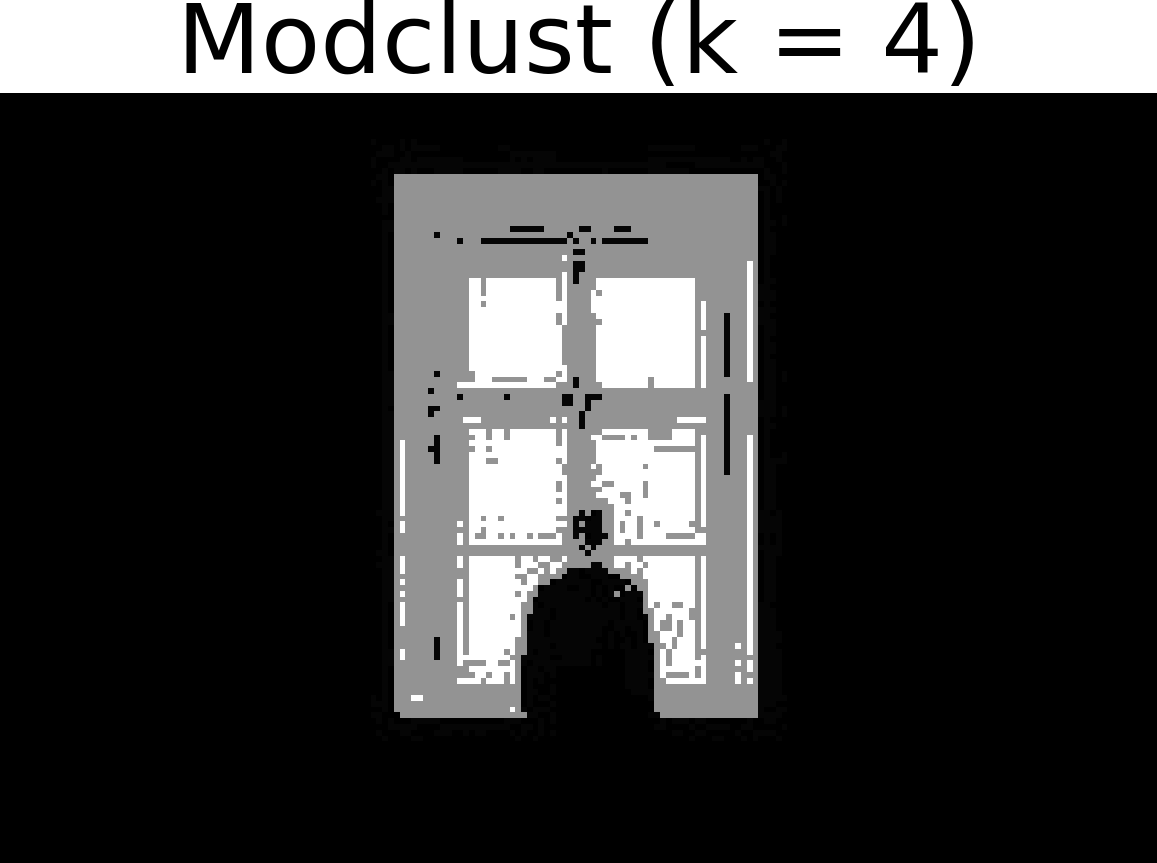}
		\includegraphics[width=0.13\linewidth]{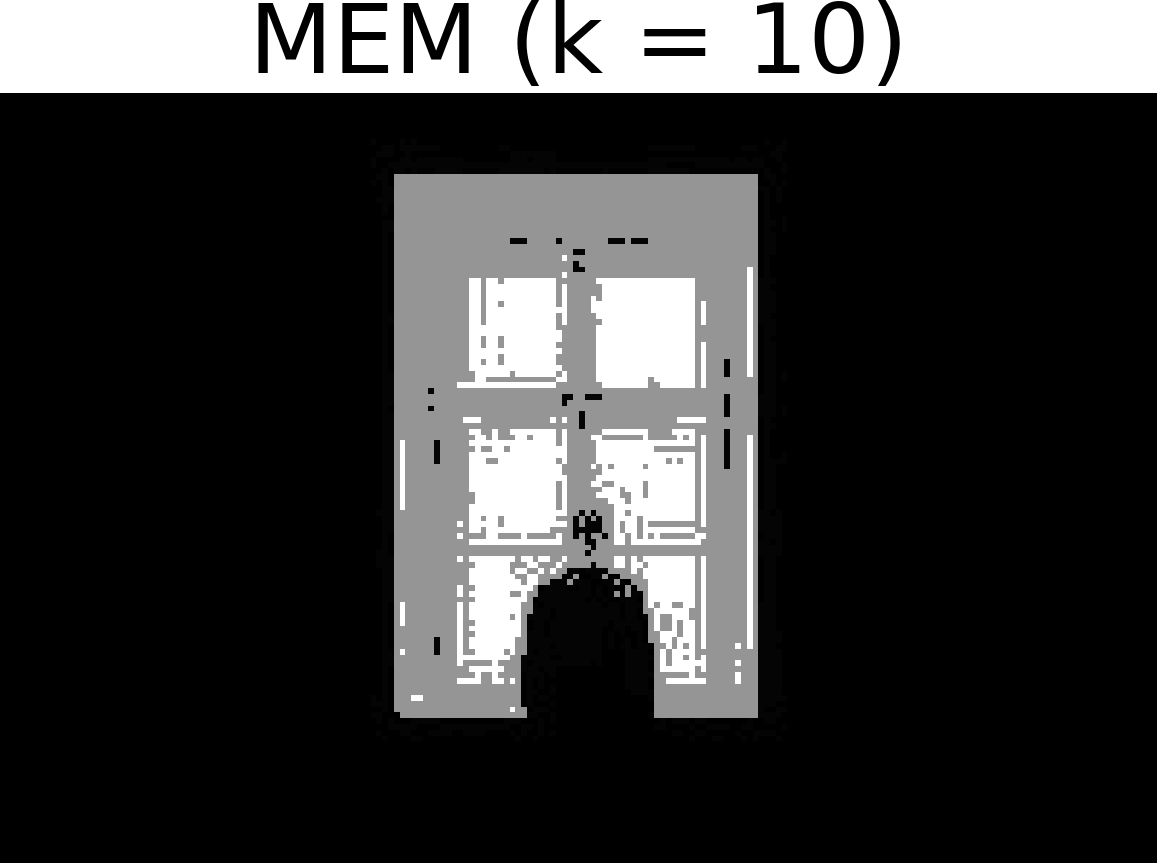}
		\caption*{}
		\label{figure:fig13f}
	\end{subfigure}
\vspace{-1.2cm}
	\caption{Initial images (first column). Segmented images obtained by the compared methods (second - seventh column). For rgb images the ground truth value of colors ($k^\star$) is illustrated, while the estimated number of colors ($k$) is provided for both rgb and grayscale images.}
	\label{figure:fig12}
\end{figure}

In the top four rows of Fig.~\ref{figure:fig12} we present the original images in grayscale (leftmost image in each row) along with the segmented images obtained by each method.
Above the original and segmented images, the ground truth value of colors ($k^\star$) and the obtained value of colors ($k$) by the compared methods are recorded, respectively. For most images, the methods provide similar visual results, accurately detecting the main colors of each image. It can be observed that when additional segments are detected compared to ground truth, these segments correspond to very small regions of the image that are difficult to be visually detected. For instance, UniSplit and TailoredDip detect thin line segments between the three major segments of the France flag. Similarly, in the European flag, mean shift assigns multiple colors to the stars, while MEM produces a noisy segmentation, as indicated in Table~\ref{table:table6}.

We also evaluated the six compared methods on grayscale images utilizing the same parameter values as those used in the previous experiment. To illustrate the segmentation result for each image, we assign to each pixel the average color value of its group, since the number of colors in these images cannot be reliably assessed through visual inspection. Therefore a ground truth partition cannot be specified, thus NMI values cannot be computed. In the bottom three rows of Fig.~\ref{figure:fig12}, the original grayscale images (first column) are presented alongside the segmentation results and the number of segments ($k$) obtained by each method. UniSplit produces results closely resembling the original images in all cases, while the other methods often fall short. Notably, TailoredDip and FTC fail to segment the Moon image accurately, estimating fewer segments, while modclust and MEM generate noisy segmentations in the Man-Window image.

\subsection{UDMM Naive Bayes for Classification}

A machine learning algorithm that requires the statistical model of univariate data is the well-known Naive Bayes classifier \cite{bishop2006pattern}. This method assumes independence among all $d$ features of each example, therefore the per class density of each feature $p(z_i|C_k)$ is estimated by considering the set of values of feature $z_i$ for the examples belonging to class $C_k$. Once the densities $p(z_i|C_k)$ have been determined for all features $z_i$ and classes $C_k$, the posterior probability that an example $z=(z_1, ..., z_d)$ belongs to class $C_k$ is proportional to $P(C_k | z) \propto P(C_k) \prod\limits_{i=1}^{d} p(z_i | C_k)$
where $P(C_k)$ are typically set equal to class frequencies and the example $z$ is assigned to the class with maximum posterior probability.

\begin{figure}[H]
\hspace{-0.5cm}
	\begin{subfigure}[b]{0.35\linewidth}
		\includegraphics[width=\linewidth]{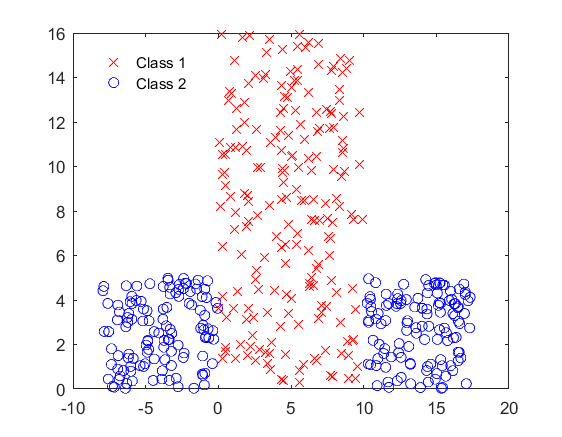}
		\caption{Ground truth solution.}
		\label{figure:fig14a}
	\end{subfigure}
\hspace{-0.3cm}
	\begin{subfigure}[b]{0.35\linewidth}
		\includegraphics[width=\linewidth]{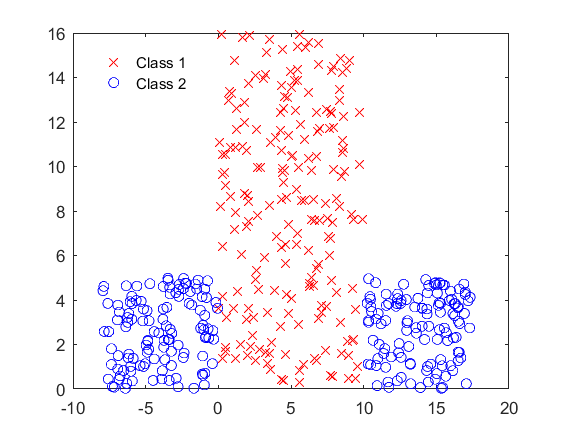}	
		\caption{UDMM-NB solution.}
		\label{figure:fig14b}
	\end{subfigure}
\hspace{-0.3cm}
	\begin{subfigure}[b]{0.35\linewidth}
		\includegraphics[width=\linewidth]{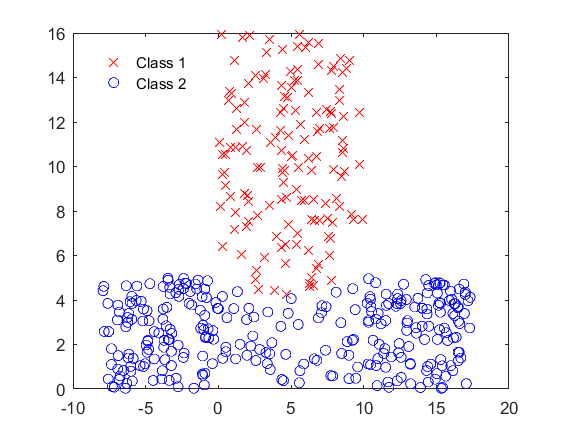}	
		\caption{GNB solution.}
		\label{figure:fig14c}
	\end{subfigure}
	\caption{Data generated by three uniform rectangles assigned to two classes.}
	\label{figure:fig14}
\end{figure}

A widely approach is Gaussian Naive Bayes (GNB), which assumes that $p(z_i | C_k)$ follows a single Gaussian distribution. In this experiment we model each feature density $p(z_i | C_k)$ using a UDMM and we call the resulting method as UDMM-NB. We have considered one synthetic and several real datasets \cite{Dua:2019}. For each dataset, we used 10-fold cross validation to measure the accuracy. 
Table~\ref{table:table7} provides the names and parameters (n: number of samples, d: number of features, K: number of classes) of each dataset in the first four columns, with the average and standard deviation of accuracy values for UDMM-NB and GNB in the fifth and sixth columns. UDMM-NB generally outperforms GNB, except for small datasets like Iris and Prestige, where sample sizes per class are low. A notable example is the synthetic 2-d dataset (Fig.~\ref{figure:fig14}), where UDMM-NB correctly discriminates the two classes, while GNB lacks the flexibility required for correctly modeling the data points (as also indicated in the first row of Table~\ref{table:table7}).

\begin{table}[!t]
	\renewcommand{\arraystretch}{1} 
	\caption{Accuracy results on synthetic and real datasets. Bold numbers indicate the best average performance for each dataset.}
	\label{table:table7}
	\resizebox{\textwidth}{!}{%
		\begin{tabular}{lrrrrr}
			\hline
			& \multicolumn{3}{c}{Parameters} & \multicolumn{2}{c}{Accuracy}                              \\ \hline
			Datasets                        & n          & d       & K       & \multicolumn{1}{c}{UDMM - NB} & \multicolumn{1}{c}{GNB}   \\ \hline
			Synthetic                       & 400        & 2       & 2       & \textbf{0.998 $\pm$ 0.01}     & 0.883 $\pm$ 0.03          \\
			Banknote                        & 1371       & 4       & 2       & \textbf{0.916 $\pm$ 0.02}     & 0.837 $\pm$ 0.04          \\
			Cardiotocography                & 2126       & 21      & 10      & \textbf{0.702 $\pm$ 0.02}     & 0.637 $\pm$ 0.03          \\
			Dermatology                     & 358        & 34      & 6       & 0.891 $\pm$ 0.05              & \textbf{0.893 $\pm$ 0.08} \\
			Glass                           & 214        & 9       & 6       & \textbf{0.560 $\pm$ 0.06}     & 0.458 $\pm$ 0.10          \\
			Image-Segmentation              & 210        & 19      & 7       & \textbf{0.785 $\pm$ 0.09}     & 0.766 $\pm$ 0.11          \\
			Iris                            & 150        & 4       & 3       & 0.946 $\pm$ 0.05              & \textbf{0.960 $\pm$ 0.04} \\
			Page Blocks                     & 5473       & 10      & 5       & \textbf{0.940 $\pm$ 0.01}     & 0.888 $\pm$ 0.02          \\
			Prestige                        & 98         & 5       & 3       & 0.918 $\pm$ 0.06              & \textbf{0.948 $\pm$ 0.06} \\
			Steel Plates Faults             & 1941       & 27      & 7       & \textbf{0.663 $\pm$ 0.02}     & 0.462 $\pm$ 0.02          \\
			Wall Following Robot Navigation & 5456       & 4       & 4       & \textbf{0.972 $\pm$ 0.01}     & 0.891 $\pm$ 0.01          \\
			Wall Following Robot Navigation & 5456       & 24      & 4       & \textbf{0.898 $\pm$ 0.03}     & 0.524 $\pm$ 0.01          \\ \hline
		\end{tabular}%
	}
\end{table}

\subsection{Examples with Noise and Outliers}

\begin{figure}[!t]
\centering
	\begin{subfigure}[b]{0.65\linewidth}
		\includegraphics[width=\linewidth]{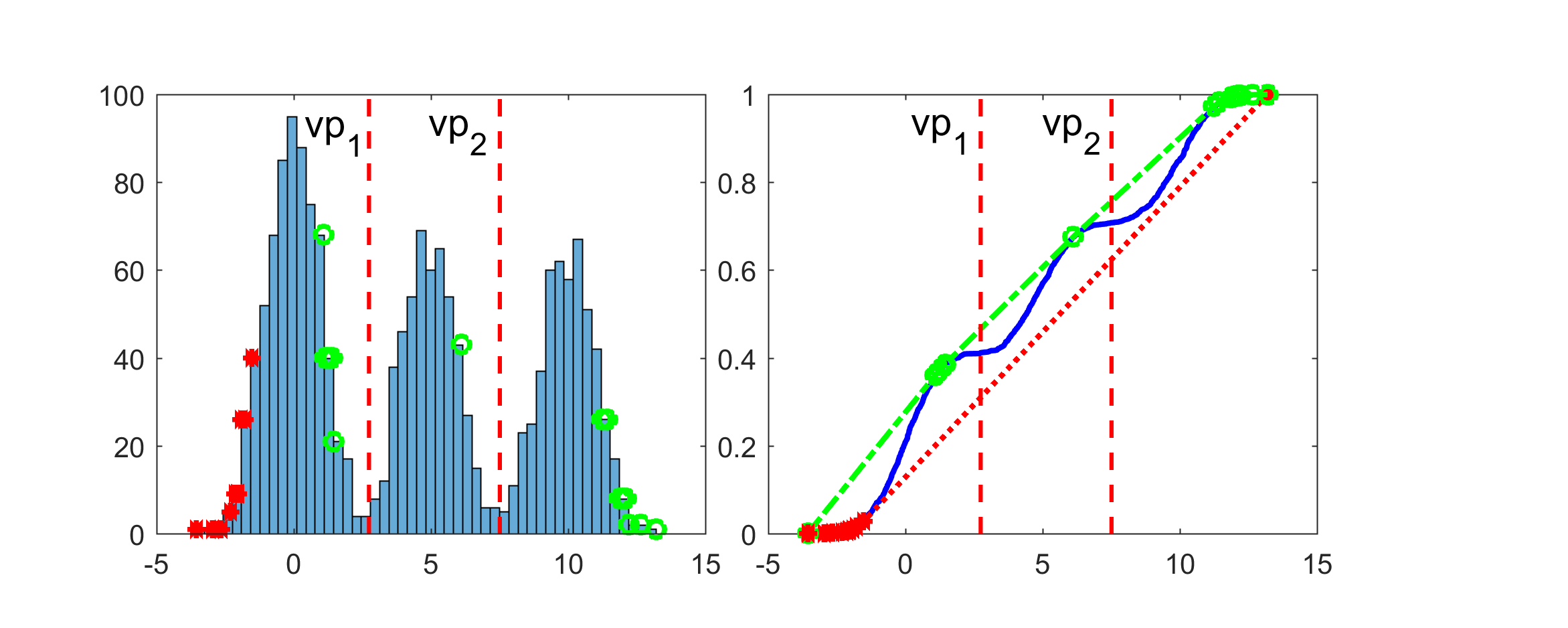}
		\caption{Original trimodal dataset.}
		\label{figure:fig15a}
	\end{subfigure} 
	\begin{subfigure}[b]{0.65\linewidth}
		\includegraphics[width=\linewidth]{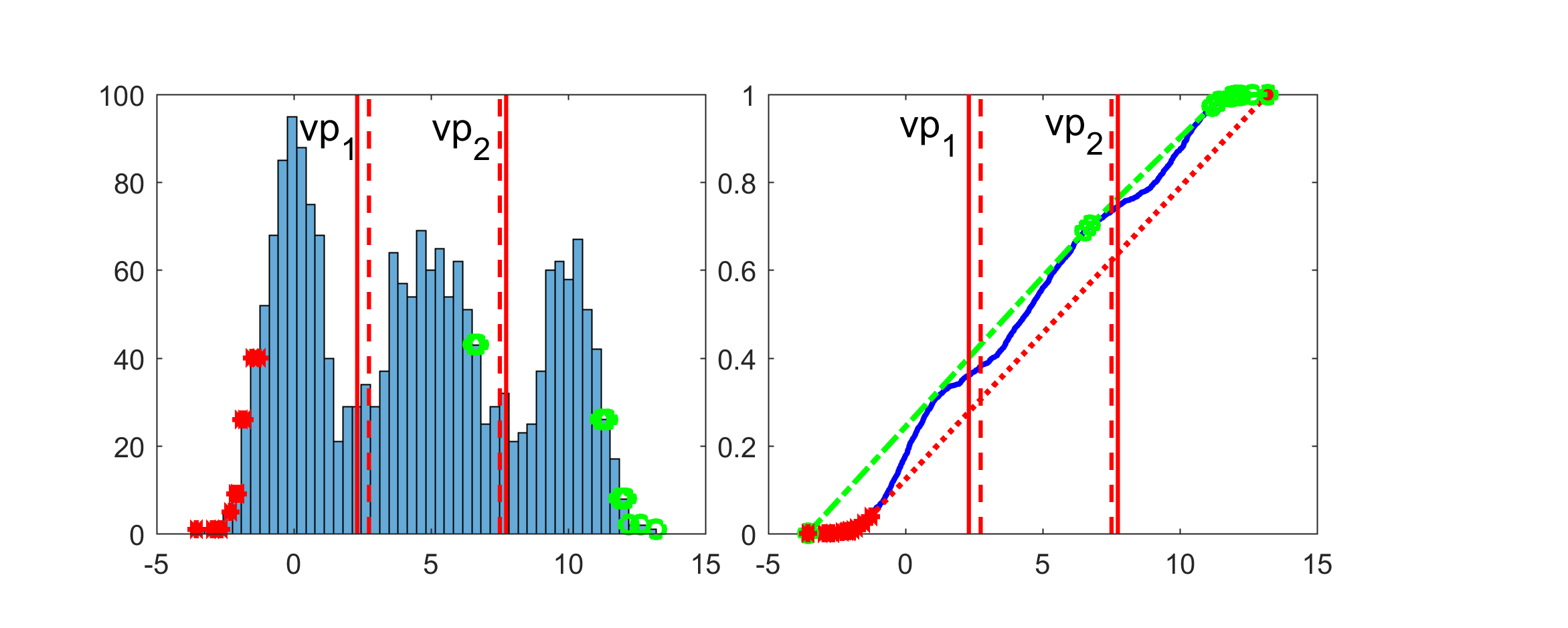}
		\caption{Trimodal dataset with uniform noise added to the valleys.}
		\label{figure:fig15b}
	\end{subfigure}
	\begin{subfigure}[b]{0.65\linewidth}
		\includegraphics[width=\linewidth]{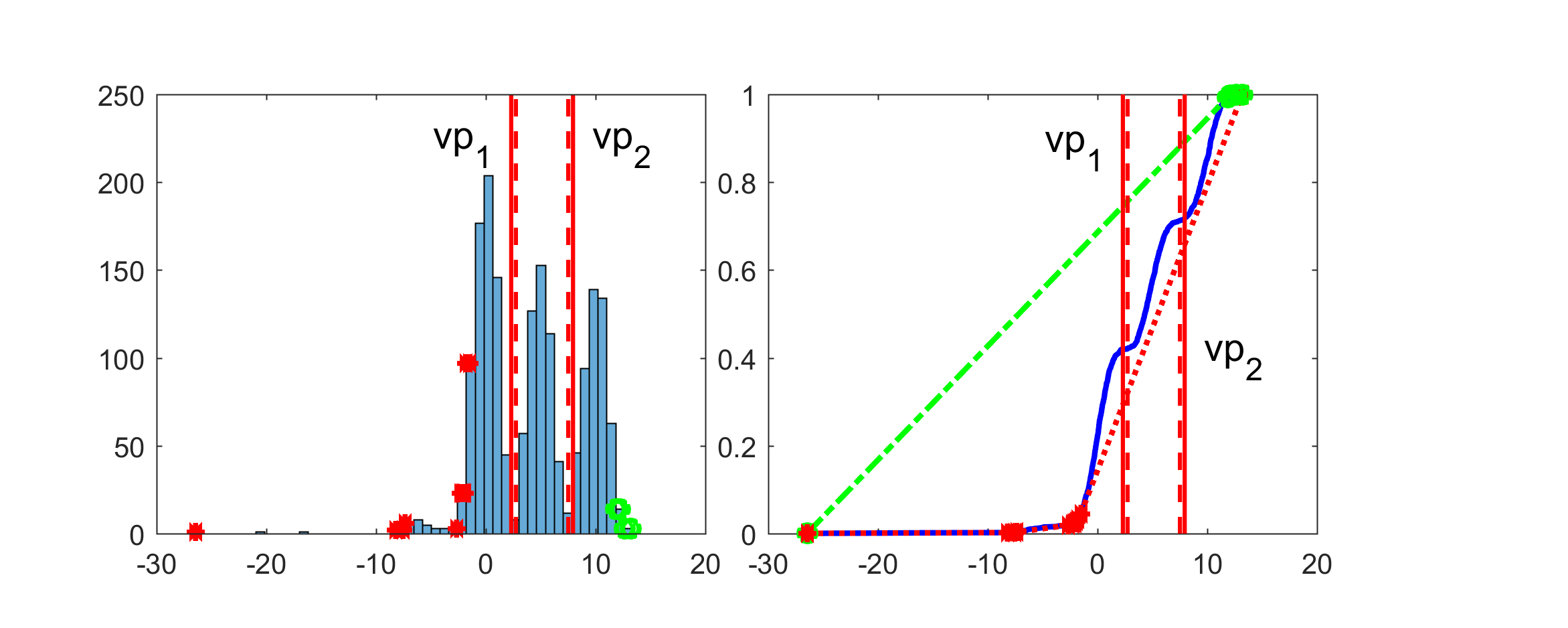}
		\caption{Trimodal dataset with left-side Student's t-distributed noise (outliers).}
		\label{figure:fig15c}
	\end{subfigure}
	\caption{Histogram and ecdf plots of a trimodal dataset before and after adding noise/outliers. The original valley points (dotted vertical lines) are almost identical to the final valley points (solid vertical lines).}
	\label{figure:fig15}
\end{figure}

Noise and outliers can affect the ecdf shape, as well as the gcm/lcm points positions, but as shown in \cite{chasani2022uu}, the unimodality decisions by UU-test remain unaffected. Previous experiments with synthetic datasets containing outliers, such as distributions D6, D8, and D12 (Table~\ref{table:table3}), demonstrated that UDMM outperformed other models in component detection and modeling accuracy, even with extreme values (e.g., in D6 the range is \([-6800, 330]\)). 
We next provide an example that highlights UniSplit's robustness to noise and outliers. For a trimodal dataset (Fig.~\ref{figure:fig15a}), UniSplit identifies two valley points (\(vp_1\), \(vp_2\)) (dotted vertical lines). When uniform noise is added between Gaussians (Fig.~\ref{figure:fig15b}), the ecdf changes, but the valley points (solid vertical lines) remain close to their original locations. Similarly, the addition of outliers (on the left) generated from a Student’s t distribution (Fig.~\ref{figure:fig15c}) shifts gcm/lcm points and modifies the ecdf significantly, however UniSplit detects correctly the number and positions of valley points (solid lines coincide with the original dotted lines), indicating robustness against noise and outliers.

\subsection{Impact of the Statistical Significance Level}

The UniSplit method automatically estimates the number of valleys in univariate multimodal data, leading to the automatic determination of the number of components in the UDMM, unlike other models requiring user-defined hyperparameters. UniSplit requires solely the significance level (\(\alpha\)) of the uniformity test employed in UU-test, which was set to \(\alpha = 0.01\) in all previous experiments.

To examine the influence of \(\alpha\), experiments were repeated with \(\alpha = 0.05\) and \(\alpha = 0.1\) using datasets from Table~\ref{table:table2}. Results showed minimal influence on UDMM's performance or component count. In 9 of 21 datasets, the number of components remained unchanged across all values of \(\alpha\). In 10 datasets, small increases ($0.6\%$–$12.3\%$) were observed as \(\alpha\) increases. For example, in synthetic dataset D4, the average number of components increased from \(k=3\) (when \(\alpha = 0.01\)) to \(k=3.02\) (when \(\alpha = 0.1\)), and in D10, from \(k=4.05\) to \(k=4.55\). In the real datasets stamps and geyser, \(k\) increased more noticeably (from 1 to 3); however they could be considered as borderline cases of unimodality, as evident from histogram inspection.

\section{Conclusions and Future Work} \label{Conclusions}

We have proposed an approach for partitioning and statistical modeling of univariate datasets. The method relies on the notion of unimodality and partitions the dataset into unimodal subsets through a novel approach for determining valley points in the probability density. We have introduced properties of critical points (gcm/lcm points) of the data ecdf that provide indications on the existence of density valleys and further are exploited in the proposed UniSplit algorithm. UniSplit is non-parametric and automatically estimates the number of unimodal subsets. In contrast to other approaches, it requires only a statistical significance threshold as input and no other user specified hyperparameters. After splitting the datasets into unimodal subsets, our approach constructs a Unimodal Mixture Model (UDMM), where each mixture component constitutes a statistical model of the corresponding unimodal subset in the form of a Uniform Mixture Model (UMM). The number of UDMM components is automatically obtained by the proposed UniSplit method, which constitutes a significant advantage over other models (e.g., GMM). In addition UDMM is very flexible and does not assume any specific parametric form for the unimodal mixture components.
Experimental results on various modeling and clustering tasks indicate that UniSplit and UDMM are generally superior to competing methods without requiring any hyperparameter tuning.

A limitation arises when dealing with very small datasets and, specifically, when limited data exist within specific intervals. In such case, valley points might exist without being identified, since the necessary gcm/lcm points might be missing.

Since the proposed approach provides accurate statistical modeling of univariate data, it could be employed in any method or application requiring this type of modeling. Exploitation of the method for partitioning and statistical modeling of multidimensional datasets constitutes an important future research direction. For example, this could be achieved by determining appropriate univariate projections of the data where UniSplit could be employed for data splitting. UniSplit could also be employed for recursively splitting a multidimensional dataset based on the values of a single feature each time. In this way an unsupervised decision tree can be built that provides interpretable clustering solutions in the form of hyperrectangles.

\bibliographystyle{elsarticle-num} 
\bibliography{Arxiv_Manuscript}

\begin{thebibliography}{10}
\expandafter\ifx\csname url\endcsname\relax
  \def\url#1{\texttt{#1}}\fi
\expandafter\ifx\csname urlprefix\endcsname\relax\def\urlprefix{URL }\fi
\expandafter\ifx\csname href\endcsname\relax
  \def\href#1#2{#2} \def\path#1{#1}\fi

\bibitem{bishop2006pattern}
C.~M. Bishop, N.~M. Nasrabadi, Pattern Recognition and Machine Learning,
  Vol.~4, Springer, 2006.

\bibitem{sampaio2024regularization}
R.~A. Sampaio, J.~D. Garcia, M.~Poggi, T.~Vidal, Regularization and
  optimization in model-based clustering, Pattern Recognition 150 (2024)
  110310.

\bibitem{dharmadhikari1988unimodality}
S.~Dharmadhikari, K.~Joag-Dev, Unimodality, convexity, and applications,
  Elsevier, 1988.

\bibitem{kalogeratos2012dip}
A.~Kalogeratos, A.~Likas, Dip-means: an incremental clustering method for
  estimating the number of clusters, in: Advances in Neural Information
  Processing Systems, 2012, pp. 2393--2401.

\bibitem{schelling2020dataset}
B.~Schelling, C.~Plant, Dataset-transformation: improving clustering by
  enhancing the structure with dipscaling and diptransformation, Knowledge and
  Information Systems 62~(2) (2020) 457--484.

\bibitem{chasani2022uu}
P.~Chasani, A.~Likas, The uu-test for statistical modeling of unimodal data,
  Pattern Recognition 122 (2022) 108272.

\bibitem{hartigan1985dip}
J.~A. Hartigan, P.~M. Hartigan, et~al., The dip test of unimodality, The Annals
  of Statistics 13~(1) (1985) 70--84.

\bibitem{siffer2018your}
A.~Siffer, P.-A. Fouque, A.~Termier, C.~Largou{\"e}t, Are your data gathered?,
  in: Proceedings of the 24th ACM SIGKDD International Conference on Knowledge
  Discovery \& Data Mining, 2018, pp. 2210--2218.

\bibitem{fukunaga1975estimation}
K.~Fukunaga, L.~Hostetler, The estimation of the gradient of a density
  function, with applications in pattern recognition, IEEE Transactions on
  Information Theory 21~(1) (1975) 32--40.

\bibitem{cheng1995mean}
Y.~Cheng, Mean shift, mode seeking, and clustering, IEEE Transactions on
  Pattern Analysis and Machine Intelligence 17~(8) (1995) 790--799.

\bibitem{scrucca2019transformation}
L.~Scrucca, A transformation-based approach to gaussian mixture density
  estimation for bounded data, Biometrical Journal 61~(4) (2019) 873--888.

\bibitem{li2006two}
J.~Li, H.~Zha, Two-way poisson mixture models for simultaneous document
  classification and word clustering, Computational Statistics \& Data Analysis
  50~(1) (2006) 163--180.

\bibitem{banerjee2005clustering}
A.~Banerjee, I.~S. Dhillon, J.~Ghosh, S.~Sra, G.~Ridgeway, Clustering on the
  unit hypersphere using von mises-fisher distributions., Journal of Machine
  Learning Research 6~(9) (2005).

\bibitem{chacon2019mixture}
J.~E. Chac{\'o}n, Mixture model modal clustering, Advances in Data Analysis and
  Classification 13 (2019) 379--404.

\bibitem{hartigan1975clustering}
J.~Hartigan, Clustering algorithms, John Wiley \& Sons, New York, 1975.

\bibitem{chacon2020modal}
J.~E. Chac{\'o}n, The modal age of statistics, International Statistical Review
  88~(1) (2020) 122--141.

\bibitem{menardi2016review}
G.~Menardi, A review on modal clustering, International Statistical Review
  84~(3) (2016) 413--433.

\bibitem{myhre2018robust}
J.~N. Myhre, K.~{\O}. Mikalsen, S.~L{\o}kse, R.~Jenssen, Robust clustering
  using a knn mode seeking ensemble, Pattern Recognition 76 (2018) 491--505.

\bibitem{sheikh2007mode}
Y.~A. Sheikh, E.~A. Khan, T.~Kanade, Mode-seeking by medoidshifts, in: 2007
  IEEE 11th International Conference on Computer Vision, IEEE, 2007, pp. 1--8.

\bibitem{rodriguez2014clustering}
A.~Rodriguez, A.~Laio, Clustering by fast search and find of density peaks,
  Science 344~(6191) (2014) 1492--1496.

\bibitem{rasool2023overcoming}
Z.~Rasool, S.~Aryal, M.~R. Bouadjenek, R.~Dazeley, Overcoming weaknesses of
  density peak clustering using a data-dependent similarity measure, Pattern
  Recognition 137 (2023) 109287.

\bibitem{li2007nonparametric}
J.~Li, S.~Ray, B.~G. Lindsay, A nonparametric statistical approach to
  clustering via mode identification., Journal of Machine Learning Research
  8~(8) (2007).

\bibitem{scrucca2021fast}
L.~Scrucca, A fast and efficient modal em algorithm for gaussian mixtures,
  Statistical Analysis and Data Mining: The ASA Data Science Journal 14~(4)
  (2021) 305--314.

\bibitem{silverman1981using}
B.~W. Silverman, Using kernel density estimates to investigate multimodality,
  Journal of the Royal Statistical Society: Series B (Methodological) 43~(1)
  (1981) 97--99.

\bibitem{maurus2016skinny}
S.~Maurus, C.~Plant, Skinny-dip: clustering in a sea of noise, in: Proceedings
  of the 22nd ACM SIGKDD International Conference on Knowledge Discovery and
  Data Mining, 2016, pp. 1055--1064.

\bibitem{bauer2023extension}
L.~G. Bauer, C.~Leiber, C.~B{\"o}hm, C.~Plant, Extension of the dip-test
  repertoire-efficient and differentiable p-value calculation for clustering,
  in: Proceedings of the 2023 SIAM International Conference on Data Mining
  (SDM), SIAM, 2023, pp. 109--117.

\bibitem{delon2006nonparametric}
J.~Delon, A.~Desolneux, J.-L. Lisani, A.~B. Petro, A nonparametric approach for
  histogram segmentation, IEEE Transactions on Image Processing 16~(1) (2006)
  253--261.

\bibitem{dodge2008kolmogorov}
Y.~Dodge, Kolmogorov--smirnov test, The Concise Encyclopedia of Statistics
  (2008) 283--287.

\bibitem{scrucca2016mclust}
L.~Scrucca, M.~Fop, T.~B. Murphy, A.~E. Raftery, mclust 5: clustering,
  classification and density estimation using gaussian finite mixture models,
  The R journal 8~(1) (2016) 289.

\bibitem{scrucca2023model}
L.~Scrucca, C.~Fraley, T.~B. Murphy, A.~E. Raftery, Model-based clustering,
  classification, and density estimation using mclust in R, Chapman and
  Hall/CRC, 2023.

\bibitem{JSSv097i09}
J.~Ameijeiras-Alonso, R.~M. Crujeiras, A.~Rodriguez-Casal,
  \href{https://www.jstatsoft.org/index.php/jss/article/view/v097i09}{multimode:
  An r package for mode assessment}, Journal of Statistical Software 97~(9)
  (2021) 1–32.
\newblock \href {https://doi.org/10.18637/jss.v097.i09}
  {\path{doi:10.18637/jss.v097.i09}}.
\newline\urlprefix\url{https://www.jstatsoft.org/index.php/jss/article/view/v097i09}

\bibitem{rousseeuw1987silhouettes}
P.~J. Rousseeuw, Silhouettes: a graphical aid to the interpretation and
  validation of cluster analysis, Journal of Computational and Applied
  Mathematics 20 (1987) 53--65.

\bibitem{scott2015multivariate}
D.~W. Scott, Multivariate density estimation: theory, practice, and
  visualization, John Wiley \& Sons, 2015.

\bibitem{silverman1986density}
B.~Silverman, Density estimation for statistics and data analysis, Monographs
  on Statistics and Applied Probability (1986).

\bibitem{leiber2023benchmarking}
C.~Leiber, L.~Miklautz, C.~Plant, C.~B{\"o}hm, Benchmarking deep clustering
  algorithms with clustpy, in: 2023 IEEE International Conference on Data
  Mining Workshops (ICDMW), IEEE, 2023, pp. 625--632.

\bibitem{Dua:2019}
D.~Dua, C.~Graff, \href{http://archive.ics.uci.edu/ml}{{UCI} machine learning
  repository} (2017).
\newline\urlprefix\url{http://archive.ics.uci.edu/ml}

\end{thebibliography}

\end{document}